\documentclass{article}

\PassOptionsToPackage{numbers, compress}{natbib}

\usepackage[preprint]{neurips_2021}

\usepackage[utf8]{inputenc} 
\usepackage[T1]{fontenc}    
\usepackage{url}            
\usepackage{booktabs}       
\usepackage{amsfonts}       
\usepackage{nicefrac}       
\usepackage{microtype}      
\usepackage{xcolor}         
\usepackage{graphicx}
\usepackage{siunitx,etoolbox}
\robustify\bfseries
\usepackage{booktabs}
\usepackage{multirow}
\def\rot{\rotatebox}
\usepackage{adjustbox}
\usepackage{subfigure}
\usepackage[inline]{enumitem}
\usepackage{hyperref}  
\usepackage{rotating}

\title{PDFNet: Pointwise Dense Flow Network for Urban-Scene Segmentation}

\author{%
  Venkata Satya Sai Ajay Daliparthi \\
  Blekinge Insitute of Technology\\
  Karlskrona, Sweden \\
  \texttt{veda18@student.bth.se} \\
}

\begin{document}

\maketitle

\begin{abstract}
In recent years, using a deep convolutional neural network (CNN) as a feature encoder (or backbone) is the most commonly observed architectural pattern in several computer vision methods, and semantic segmentation is no exception. The two major drawbacks of this architectural pattern are: (i) the networks often fail to capture small classes such as wall, fence, pole, traffic light, traffic sign, and bicycle, which are crucial for autonomous vehicles to make accurate decisions. (ii) due to the arbitrarily increasing depth, the networks require massive labeled data and additional regularization techniques to converge and to prevent the risk of over-fitting, respectively. While regularization techniques come at minimal cost, the collection of labeled data is an expensive and laborious process. In this work, we address these two drawbacks by proposing a novel lightweight architecture named point-wise dense flow network (PDFNet). In PDFNet, we employ dense, residual, and multiple shortcut connections to allow a smooth gradient flow to all parts of the network. The extensive experiments on Cityscapes and CamVid benchmarks demonstrate that our method significantly outperforms baselines in capturing small classes and in few-data regimes. Moreover, our method achieves considerable performance in classifying out-of-the training distribution samples, evaluated on Cityscapes to KITTI dataset.
\end{abstract}

\section{Introduction}
\label{1}
Semantic segmentation is a fundamental computer vision task in fields such as autonomous driving and robotic navigation. The pioneering work \textit{fully convolutional network} (FCN) \cite{long2015fully} illustrated that the image classification networks can be adopted for semantic segmentation. Several works further improved the FCN \cite{long2015fully} architecture and have proven to be successful in diverse segmentation benchmarks.\newline
However, due to the repeated convolutional and pooling operations, the final layers of the deep CNNs cannot fully capture the contextual information regarding various objects in the input image. Earlier works such as FCN \cite{long2015fully} introduced skip-connections from different stages of the network to recover the lost information. The encoder-decoder networks such as DeconvNet \cite{noh2015learning}, SegNet \cite{badrinarayanan2017segnet}, and U-Net \cite{ronneberger2015u} learned the up-sampling process through a decoder network. Recent works \cite{chen2018encoder,zhao2017pyramid} replaced the decoder network with a simple feature map interpolation method and mainly focused on exploiting contextual information through multi-scale context \cite{8578486, zhao2017pyramid, 7913730,ChenPSA17,chen2018encoder}, relational context \cite{abs-1809-00916,10.1007/978-3-030-58539-6_11,Zhang_2018_CVPR,8578911,9010415,8954430}, and boundary context modules \cite{Bertasius_2015_ICCV,CB2016Semantic,Ding_2019_ICCV,MarinHVCTYB19,takikawa2019gated,yuan2020segfix}.\newline
Therefore, the current semantic segmentation networks pipeline can be viewed as: (1) the backbone network, (2) context encoding modules, and (3) feature map interpolation. \newline
The backbone network is used to extract features from the input image and context modules are applied on top of those extracted features. Finally, the feature map is interpolated to match the input image resolution.\newline
In 2015, ResNets \cite{he2016deep} arbitrarily increased the depth of the networks up to 150+ layers and achieved state-of-the-art performance on several vision benchmarks. Since then, the deeper networks such as ResNet \cite{he2016deep}, DenseNet \cite{huang2017densely}, and Inception \cite{szegedy2017inception} became the de-facto backbone choice for several vision methods, and semantic segmentation is no exception.
The two main drawbacks of using deeper backbone networks are:\newline
(i) On urban driving scene benchmarks, the networks often fail to capture the small classes such as wall, fence, pole, traffic light, traffic sign, and bicycle. Accurately classifying the small classes is crucial for autonomous vehicles to better understand the surroundings and to make accurate decisions. Several existing methods \cite{Bertasius_2015_ICCV,CB2016Semantic,Ding_2019_ICCV,MarinHVCTYB19,takikawa2019gated,yuan2020segfix} exploited boundary information to refine output. However, they mainly focus on overall edge errors rather than performance on small classes.\newline
(ii) Due to the arbitrarily increasing depth, the networks often require a massive amount of labeled data and additional regularization techniques to converge and to prevent the risk of over-fitting, respectively. While regularization techniques come at minimal cost, the collection of labeled data is an expensive and time-consuming process. Due to this reason, the U-Net \cite{ronneberger2015u} architecture is still exploited in several medical image segmentation methods \cite{siddique2020u}.\newline
In this work, we address the formerly mentioned drawbacks by designing a novel lightweight architecture named \textit{point-wise dense flow network} (PDFNet). In PDFNet, we combine both the dense and residual connections to allow a smooth flow of gradient to all parts of the network.
Furthermore, we attempt to investigate the architectural bottleneck limiting the performance on few data samples and small classes by proposing ``the strided convolution hypothesis."\newline
We conduct extensive experiments on Cityscapes and CamVid benchmarks to evaluate our method and our hypothesis.
The empirical results show that our PDFNet significantly outperforms baselines on, (i) accurately labeling small classes that appear in the urban driving scenario, and (ii) dealing with few data samples.\newline
Moreover, the empirical results illustrate that our strided convolution hypothesis might be valid in few-data regimes. In other words, replacing the strided Conv with regular Conv in the first layer of the network might improve the performance on few data samples.\newline
Additionally, we show that our networks achieve considerable out-of-training distribution performance from Cityscapes to KITTI benchmarks.

\section{Related work}
\label{2}
\subsection{Encoder-decoder methods} 
The encoder-decoder networks such as DeconvNet \cite{noh2015learning}, SegNet \cite{badrinarayanan2017segnet}, U-Net \cite{ronneberger2015u}, RefineNet \cite{Lin:2017:RefineNet,lin2019refinenet}, and FC-DenseNet \cite{fc-densenet} employs an encoder module that encodes the semantic information by reducing spatial resolution. The spatial information lost during encoding process is generally recovered through a decoder module. The networks often uses skip-connections and pooling indices to allow information exchange between the encoder and decoder.\newline
Several medical image segmentation methods such as DU-Net \cite{DU-Net}, MDU-Net \cite{MD-UNet}, SDN \cite{SDN}, HyperDense-Net \cite{HyperDense-Net}, Hybrid-DenseUNet \cite{Hybrid-DenseUNet}, Cascaded 3D Dense-UNet \cite{Cascaded-3D-Dense-UNet}, CUNet \cite{CUNet}, FDU-Net \cite{FDU-Net}, and Ladder-style DenseNets \cite{Ladder-style} employed dense connections \cite{huang2017densely} in encoder-decoder based architectures. \newline
However, all the formerly mentioned encoder-decoder methods learn the up-sampling process through a decoder module that is similar to the encoder module. In contrast, the decoder module in our method only employs 1x1 Convs and feature map interpolations.\newline
The full-resolution residual network \cite{pohlen2017full} maintains an additional full-resolution stream that consists of information at the full-scale resolution. It replaces skip-connections by exchanging information from each unit in the down-sampling and up-sampling process to the full-resolution stream. In our method, we do not employ any dual-stream path and did not combine multi-scale information until the final layer.\newline
The HRNet \cite{SunZJCXLMWLW19} maintains high-resolution representations throughout the network and combines information from the parallel layers. The stem in HRNet employs two 3x3 convolutional layers with stride two that reduces the size of the resulting feature map. Therefore, it only maintains representations that are four times lower than the input image resolution. On the other hand, our method maintains representations at the full-scale resolution.

\subsection{Context encoding methods}
\noindent The multi-scale context modules, such as  SPP \cite{zhao2017pyramid}, ASPP \cite{7913730,ChenPSA17,chen2018encoder}, and DenseASPP \cite{8578486} employs multiple parallel convolutional layers with different receptive fields to capture the  multi-scale information.\newline
The relational context modules such as DANet \cite{fu2019dual}, OCNet \cite{abs-1809-00916}, OCR \cite{10.1007/978-3-030-58539-6_11}, EncNet \cite{Zhang_2018_CVPR}, Non-local \cite{8578911}, ACFNet \cite{9010415}, and CoCurNet \cite{8954430} uses self-attention \cite{vaswani2017attention} approaches to compute similarities between each pixel and weights them accordingly.\newline
In contrast, our method does not employ any of the formerly mentioned context encoding modules. On the other hand, similar to FPN \cite{lin2017feature}, we use multi-scale features within the network to capture the contextual information.

\subsection{Boundary refinement methods} 
\noindent The traditional methods \cite{7913730,crfRNN} employed DenseCRFs \cite{denseCRF} to refine the segmentation output. Recent methods such as \cite{Bertasius_2015_ICCV,CB2016Semantic,Ding_2019_ICCV,MarinHVCTYB19}, Gated-SCNN \cite{takikawa2019gated}, and SegFix \cite{yuan2020segfix} exploited boundary information through introducing refinement modules and additional streams.\newline
The formerly mentioned boundary refinement methods only focus on overall edge error rates. In contrast, our method focuses on accurately labeling small classes.
\subsection{Few-shot semantic segmentation (FSS)}
\noindent The FSS methods employs techniques such as meta-learning (knowledge distillation) \cite{xu2016deep,dong2018few,RakellySDEL18,2009-06680,TianWQWSG20} and metric-learning (similarity learning) \cite{BMVC2017_167,9108530,Wang_2019_ICCV,Zhang_2019_CVPR,WangZHYCZ20,WangZHYCZ20,Zhang_2019_ICCV,YangLLJY20} to learn from few-data samples. They involve multi-stage training and also deals with novel classes during testing. \newline
Our method does not employ any formerly mentioned techniques to deal with few data samples and is limited to classes seen during the training. Therefore, our method is more closely related to supervised learning methods rather than few-shot learning methods.
\section{Method}
\label{3}
\subsection{Pointwise dense flow network (PDFNet)}
\label{3.1}

\begin{figure}[t!]
\begin{center}
   \includegraphics[width=1\linewidth]{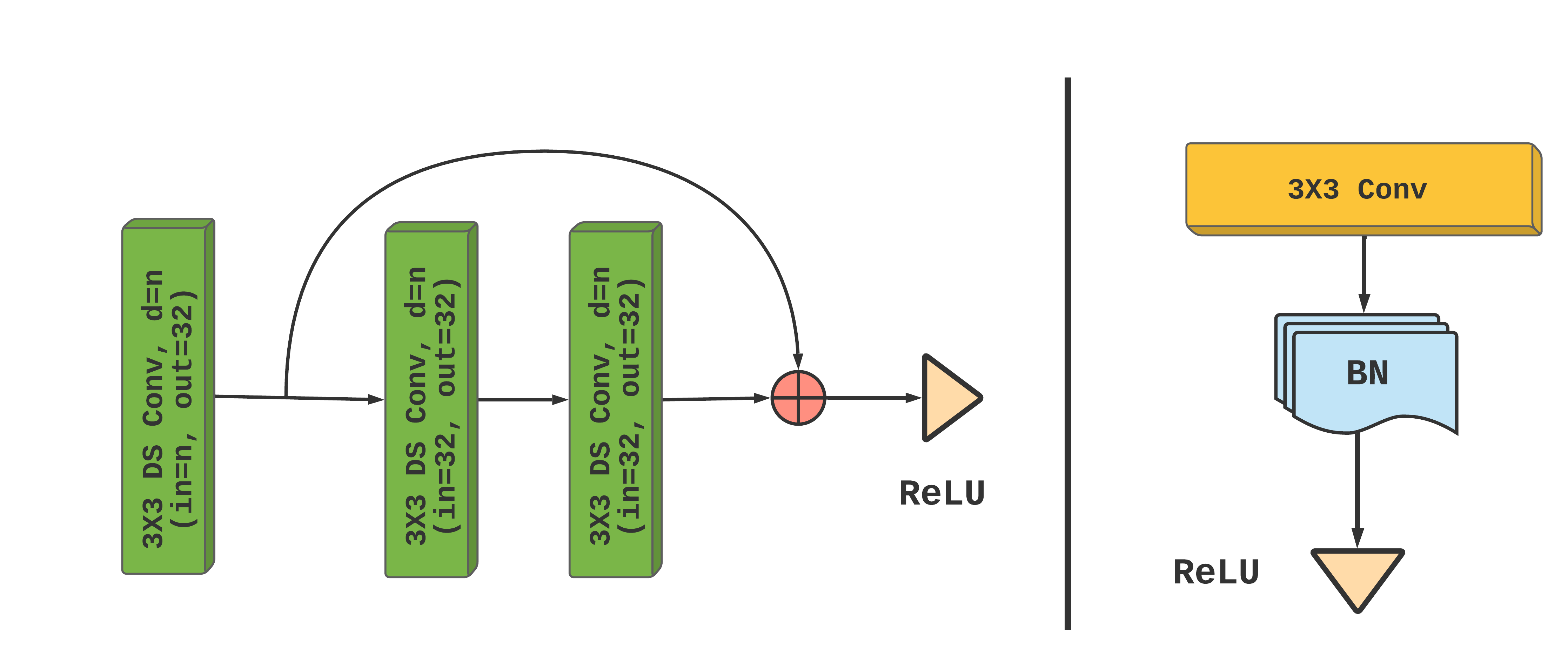}
\end{center}
   \caption{The glance module}
\label{fig:glance}
\end{figure}

\noindent The PDFNet architecture consists of three basic building blocks. They are (i) the glance module, (ii) the $1$x$1$ Conv layer, and (ii) the average pooling layer. \newline
The glance module consists of three $3$x$3$ dilated depth-wise separable Convs with the same dilation rate and a residual connection in between, as shown in Figure \ref{fig:glance}.\newline
The purpose of the glance module is to explore new features and it is only employed in the encoder module. It accepts any arbitrary number of input filters and always returns $32$ filters as output.\newline
The networks such as VGG \cite{simonyan2014very} ($3$x$3$ Convs), Inception \cite{7298594} ($3$x$3$, $5$x$5$, $7$x$7$ Convs $\approx$ $3$x$3$ Convs with dilation rate = $1$, $2$, $3$), ResNet \cite{he2016deep} (residual connection), and Xception \cite{8099678} (depth-wise separable Convs \cite{sifre2014rigid}) inspire the design choices of the glance module.\newline
For a given feature map, we employ $1$x$1$ Conv layers \cite{lin2013network} and average pooling layers to reduce filter dimensions and spatial resolution, respectively. \newline
Except for the last $1$x$1$ Conv layer that returns final output, every Conv layer is in the PDFNet is followed by a batch normalization \cite{ioffe2015batch} and a ReLU \cite{nair2010rectified} activation layer. \newline
\begin{figure}[t!]
\begin{center}
   \includegraphics[width=1\linewidth]{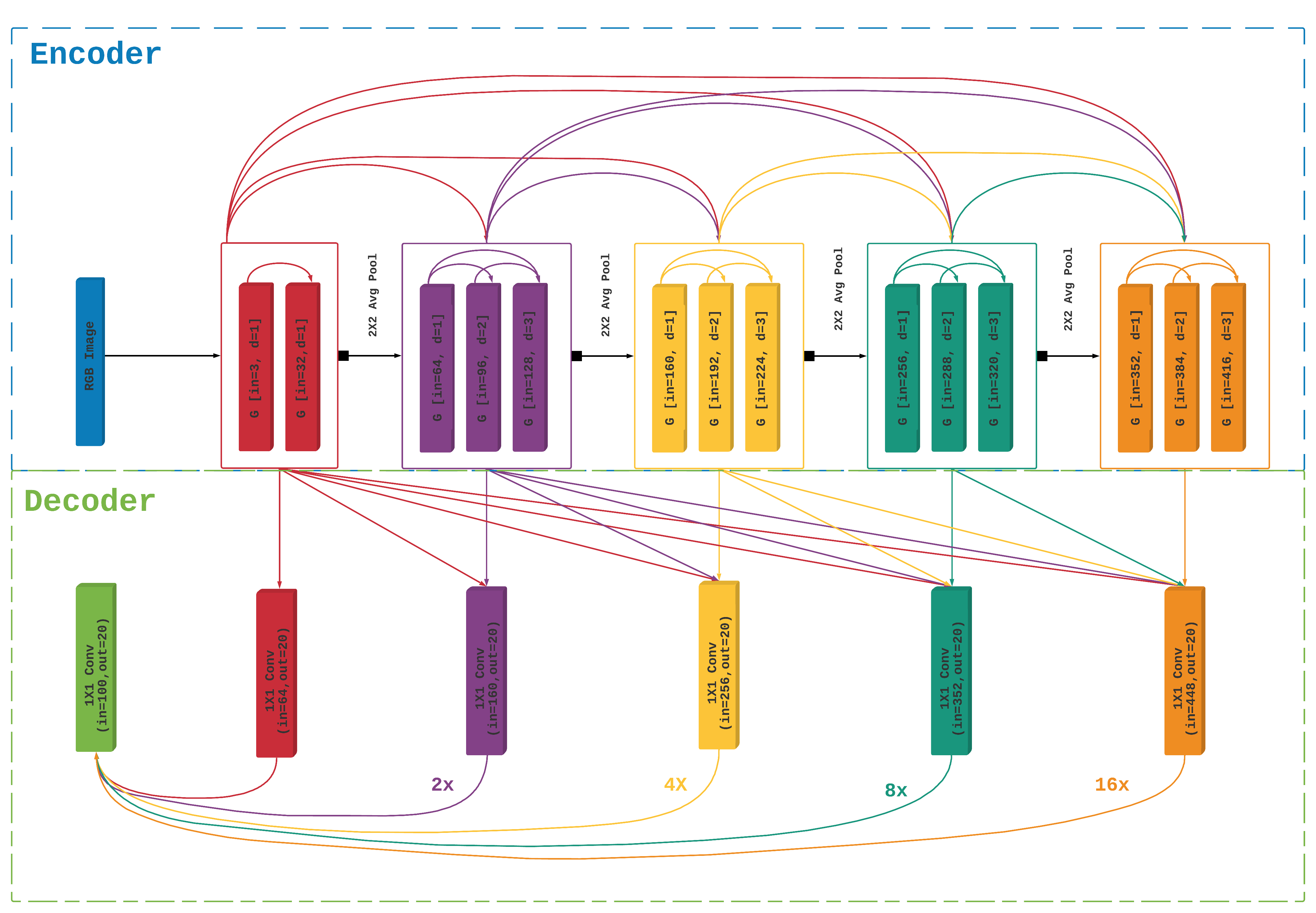}
\end{center}
   \caption{The PDFNet3 architecture (best viewed in color)}
\label{fig:PDFNet3}
\end{figure}
In Figure \ref{fig:PDFNet3}, we present the PDFNet3 architecture that consists of an encoder and decoder.\newline
At the encoder, every layer receives its preceding layers output as input through dense connections \cite{huang2017densely}\footnote{The encoder connections in the PDFNet are similar to the full dense connectivity pattern introduced in CondenseNet \cite{CondenseNet}. However, the basic building block of \cite{CondenseNet} consists of a learnable group $1$x$1$ Conv layer (to prune the incoming filters), a shuffling layer, and multiple $3$x$3$ group Conv layers, which is entirely different from our glance module.}.
At the first stage, we pass the RGB input image through the first glance module \footnote{For the first Conv layer in the first glance module, we replace the depth-wise separable Conv with a regular Conv layer.} and pass the resulting output (with $32$ filters) through the second glance module. Then, we concatenate the resulting outputs and name it $S_{1} \Rightarrow (32 + 32 = 64) $. Here, we pass $S_{1}$ through an average pooling layer to reduce the spatial resolution by half.  Then, we pass the pooled $S_{1}$ through the second stage of the network that consists of three glance modules with different dilation rates $(d=1,2,3)$. The second stage of the network returns $S_{2}$ with $160$ filters ( $64$ from stage one + $96$ from stage two). The same process repeats for another three stages and at every stage the networks stores the resulting outputs ($S_{1} (64)$, $S_{2} (160)$, $S_{3} (256)$, $S_{4} (352)$, and $S_{5} (448)$) .\newline
At the decoder, we pass encoder outputs through an individual $1$x$1$ Conv layer that reduces the number of filters to $20$ ($S_{1} (20)$, $S_{2} (20)$, $S_{3} (20)$, $S_{4} (20)$, and $S_{5} (20)$)\footnote{This is closely related to FPN \cite{lin2017feature}, which also uses multi-scale information within the network. However, FPN \cite{lin2017feature} omits the first Conv layer features and learns $256$ filters from the last three stages, which are further refined with multiple $3$x$3$ Conv layers. On the other hand, our method learns only $20$ filters from each stage and does not employ additional $3$x$3$ Conv layers to refine them, which allows us to operate on the full-scale resolution.}. Then, we bi-linearly interpolate the outputs from the last four stages ($S_{2}$, $S_{3}$, $S_{4}$, and $S_{5}$) to match with the resolution of the first stage ($S_{1}$).
After that, we concatenate the five-stage outputs ($20 * 5 = 100$) and pass them through a $1$x$1$ Conv layer, which takes in $100$ filters and returns $20$ filters, that is the final output of the network.\newline
In PDFNet, every layer receives gradient through successive layers (see the shortcut connections in Figure \ref{fig:PDFNet3}). For instance, the output of first glance module $S_{1}$  contributes to every stage in the decoder.\newline
For any PDFNet variant, the depth of the first two stages remains unchanged. However, for PDFNet6, PDFNet9, and PDFNet12, the number of glance modules in the last three stages are increased to six, nine, and twelve, respectively.

\subsection{The strided convolution hypothesis}
\label{3.2}
\noindent In recent years, the most commonly observed architectural pattern in deep CNNs such as ResNet \cite{he2016deep}, and DenseNet \cite{huang2017densely} is to apply a Conv layer with stride two on the input image, which reduces the size of the resulting feature map by half. In the existing literature, this is also referred to as \textit{`` the stem module''}. \newline
To the best of our knowledge, the stride in the stem module is due to the computational complexity of operating on the full-scale resolution.\newline 
While adopting deep CNNs for the high-level tasks such as semantic segmentation, many methods \cite{7913730,ChenPSA17,chen2018encoder}\footnote{Some methods replaced the first $7$x$7$ Conv layer in the ResNet \cite{he2016deep} with a $3$x$3$ Conv layer.} often replaced the Conv layers in the last stage with dilated Convs, to increase the receptive field while maintaining the spatial resolution.
However, the stride in the stem module remains unchanged in almost every method.\newline
Here, we observe that, while the input and output resolutions are equal in semantic segmentation, many methods are not capturing representations at the full-scale resolution.\newline
Therefore, we hypothesize that \textit{``the stride two in the first Conv layer might be the bottleneck limiting the performance on small classes and few data samples'', i.e, the strided convolution hypothesis}  \newline
To validate our hypothesis, we design four PDFNet variants named PDFNet3-2S, PDFNet6-2S, PDFNet9-2S, and PDFNet12-2S, by removing the first stage and employing stride two in the first Conv layer in the networks PDFNet3, PDFNet6, PDFNet9, and PDFNet12, respectively.\newline
On the other hand, the earlier networks such as DeconvNet \cite{noh2015learning}, SegNet \cite{badrinarayanan2017segnet}, and U-Net \cite{ronneberger2015u} use regular Conv in the first layer by following VGG \cite{simonyan2014very}, and captures representations at the full-scale resolution.

\section{Experiments}
\label{4}
\subsection{Experimental setup}
\label{4.1}
\noindent Framework : PyTorch $1.8$ \cite{NEURIPS2019_9015}\newline
GPU: 1 X NVIDIA Tesla P-$100$ ($16$ GB VRAM)\newline
Epochs : $180$ \newline
Batch size : $2$\newline
Criterion : Pixel-wise cross-entropy loss \newline
Learning rate scheduler : ReduceLROnPlateau (decrease factor = $0.5$ and patience = $20$ epochs) with an initial learning rate of $1e-06$. \newline
Optimizer : Stochastic gradient descent \cite{robbins1951stochastic} with Nesterov momentum \footnote{For all the baselines, we set momentum value to $0.9$ by following \cite{chen2018encoder,huang2017densely,he2016deep,mobileNetV2,TanL19,RegNet,ronneberger2015u,SunZJCXLMWLW19}. In contrast, we set momentum value to $0.7$ for all the variants of PDFNet. 
In our preliminary experiments, we observe that the training of PDFNet is unstable with $0.9$ momentum. We hypothesize that this phenomenon is due to the small size of PDFNet compared to baseline networks.}  \cite{nesterov1983method}. \newline
Random seed : We set random seed $42$ as argument in the function \textit{torch.utils.data.RandomSplit} to ensure that the data splits are reproducible.\newline
Pre-processing: We normalize all the images with mean and standard deviation values of \cite{deng2009imagenet} and did not use any data augmentation techniques.\newline
Baselines : We use the open-source implementations for networks DeepLabV3+ (ResNet-101) \cite{Resnet-101}, DeepLabV3 (DenseNet-161) \cite{Densenet-161}, HRNet-V2 \cite{HRNet-V2},  and U-Net \cite{U-Net}. We import DeeplabV3+ with encoder networks such as ResNet \cite{he2016deep}, MobileNet-V2 \cite{mobileNetV2}, ResNext \cite{Resnext}, EfficientNet \cite{TanL19}, and RegNet \cite{RegNet} from the segmentation models library \cite{Yakubovskiy:2019}\footnote{In all the experiments, we use ImageNet \cite{deng2009imagenet} pre-trained weights for DeepLabV3+ \cite{chen2018encoder} with backbone networks ResNet-101 \cite{he2016deep}, DenseNet-161 \cite{huang2017densely}, HRNet-V2 \cite{SunZJCXLMWLW19}, EfficientNet-b6 \cite{TanL19}, and RegNetY-80 \cite{RegNet}.  Because in the existing literature, it is mentioned that these methods used an ImageNet \cite{deng2009imagenet} pre-trained network as a feature encoder and reported results with pre-trained weights only.}.

\subsection{Experiments on Cityscapes}
\label{4.2}
\subsubsection{Dataset}
The Cityscapes \cite{cordts2016cityscapes} pixel-level labeling dataset consists of $5,000$ high-resolution images finely annotated into $35$ classes. Which are further divided into $2,975/500/1,525$ images for training, validation, and testing, respectively. We convert $35$ classes into $20$ classes (including background) and resize the images from $1024$x$2048$ to $512$x$1024$. 
\subsubsection{Baseline Experiments}
\label{4.2.2}
To evaluate the performance of our method on different classes that appear in an urban driving scene, we conduct baseline experiments on the entire Cityscapes dataset. We select DeepLabV3+ (ResNet-101) \cite{he2016deep}, DeepLabV3 (DenseNet-161) \cite{huang2017densely}, HRNet-V2 \cite{SunZJCXLMWLW19}, and U-Net \cite{ronneberger2015u} as the baselines, along with four PDFNet and PDFNet-2S variants.\newline
In Table \ref{table-1}, we present mean class-wise IoU results of the networks evaluated on the val set. From Table \ref{table-1}, we observe and hypothesize the following: \newline
(i) Every PDFNet variant outperforms all other baselines networks, and PDFNet12 achieves top performance ($54.9$). \newline
(ii) Within the baseline networks, U-Net \cite{ronneberger2015u} has shown top performance ($49.3$), followed by HRNet-V2 \cite{SunZJCXLMWLW19} ($48.0$).\newline
(iii) Comparing ResNet-101 \cite{he2016deep} ($42.8$), DenseNet-161 \cite{huang2017densely} ($41.5$), and HRNet-V2 \cite{SunZJCXLMWLW19} ($48.0$). The HRNet-V2 \cite{SunZJCXLMWLW19} outperformed other networks in classes such as pole, traffic sign, person, rider, bus, and bicycle. These performance gains of HRNet-V2 \cite{SunZJCXLMWLW19} are might be due to the maintained high-resolution representations.\newline
(iv) Comparing HRNet-V2 \cite{SunXLW19} ($48.0$) and U-Net \cite{ronneberger2015u} ($49.3$). The HRNet-V2 \cite{SunXLW19} employs two strided Convs in the stem module, which results in a feature map that is four times smaller than the input image. On the other hand, U-Net \cite{ronneberger2015u} captures $64$ filters at the full-scale resolution before reducing the spatial resolution. The U-Net \cite{ronneberger2015u} outperforms HRNet-V2 \cite{SunZJCXLMWLW19} in classes (such as wall, fence, traffic light, traffic sign, person, motorcycle, and bicycle), and HRNet-V2 \cite{SunZJCXLMWLW19} outperforms U-Net \cite{ronneberger2015u} in classes (such as terrain, truck, bus, and train). We attribute the former behavior of U-Net \cite{ronneberger2015u} to the captured full-scale representations, and the latter behavior of HRNet-V2 \cite{SunZJCXLMWLW19} to the depth on low-resolution feature maps.\newline
(v) Comparing U-Net \cite{ronneberger2015u} ($49.3$) and PDFNet variants [PDFNet12 ($54.9$), PDFNet9 ($53.0$), PDFNet6 ($54.6$), and PDFNet3 ($52.0$)]. Even though, both the networks capture full-scale representations before reducing the spatial resolution. The PDFNet variants outperform U-Net \cite{ronneberger2015u} in classes such as wall, fence, pole, traffic light, traffic sign, terrain, rider, truck, bus, and train. We attribute this behavior of PDF-Net to the dense connections, simple decoder, and network design choices.\newline 
(vi) Comparing PDFNet3 ($52.0$) and PDFNet-2S variants [PDFNet12-2S ($52.9$), PDFNet9-2S ($53.7$), PDFNet6-2S ($49.6$), and PDFNet3-2S ($48.4$)]. The PDFNet3 outperformed all the PDFNet-2S variants in classes such as pole, traffic light , and traffic sign.\newline
(vii) Despite the fact that PDFNet-2S variants do not capture representations at full-scale resolution, their performance on small classes is almost similar to U-Net \cite{ronneberger2015u}. \newline
From the above observations, we found no clear evidence to support our strided convolution hypothesis (\ref{3.2}), and found that PDFNet variants significantly outperforms baselines on small classes.
\begin{table*}[t!]
\sisetup{detect-weight=true,detect-inline-weight=math}
\begin{center}
\begin{adjustbox}{width=1\textwidth}
\begin{tabular}{lSSSSSSSSSSSSSSSSSSSS}
 \toprule
  {Method} & \rot{90}{road} & \rot{90}{sidewalk} &\rot{90}{building}  & \rot{90}{wall} &\rot{90}{fence}  & \rot{90}{pole} &\rot{90}{traffic light}  & \rot{90}{traffic sign} &\rot{90}{vegetation}  & \rot{90}{terrain} &\rot{90}{sky}  & \rot{90}{person} &\rot{90}{rider}  & \rot{90}{car} &\rot{90}{truck}  & \rot{90}{bus} &\rot{90}{train}  & \rot{90}{motorcycle} &\rot{90}{bicycle}  & \rot{90}{Average}      \\
  \midrule
ResNet101 \cite{he2016deep} &  95.0 & 66.1 & 81.9& 15.0 & 13.5 & 26.7 & 20.7 & 29.5 & 86.7 & 55.4 & 89.3 & 48.5 & 6.3 & 85.5 & 6.8 & 26.1 & 19.0 & 9.8 & 32.0 & 42.8\\
DenseNet161 \cite{huang2017densely} & 94.8 & 64.5 & 81.3& 20.1 & 13.0 & 15.8 & 15.6 & 28.7 & 84.6 & \bfseries 58.7 & 86.1 & 44.1 & 0.6 & 84.7 &  17.0 & 19.7 & 23.1 & 4.3 & 31.4 & 41.5\\
HRNet-V2 \cite{SunZJCXLMWLW19}&  94.9 & 68.6 & 84.2& 24.0 & 24.5 & 39.0 & 23.2 & 42.3 & 86.9 & 51.5 & 90.2 & 55.6 & 15.3 & 86.1 & 19.9 & 36.1 & 21.2 & 2.2 & 46.1 & 48.0\\
U-Net \cite{ronneberger2015u} &  94.9 & 69.4 & 85.3 & 27.3 & 28.7 & 41.0 & 32.2 & 49.0 & 88.6 & 46.3 & 90.4 & 59.1 & 14.5 & 86.5 & 12.4 & 28.4 & 15.5 & 10.9 & 55.6 & 49.3\\
PDFNet3& 94.7 & 71.3 & 85.9 & 31.8 & 37.9 & 44.1 & 36.4 & 52.1 & 87.9 & 49.6 & 91.4 & 55.1 & 13.5 & 86.0 &  25.4 & 43.9 & 20.0 & 9.4 & 52.4 & 52.0\\
PDFNet6& \bfseries 95.5 & 72.3 & 86.0 & 34.4 & 39.6 & 43.4 & 37.1 & 51.9 & 88.3 & 52.0 & 91.4 & 58.6 & 23.3 & 87.1 &   35.4 & 47.8 &  25.4 & 11.2 & 56.8 & 54.6\\
PDFNet9&  94.8 & \bfseries 72.9 & 85.8 & 34.0 & 39.7 & 41.7 &\bfseries 39.1 & \bfseries 55.7 & 88.0 & 51.5 & 91.4 & 55.8 & 16.1 & 86.9 & 23.1 & 40.6 & 20.9 & \bfseries 14.9 & 53.8 & 53.0\\
PDFNet12& 94.6 & 72.1 & \bfseries 86.5& \bfseries 36.2 & \bfseries 39.8 & \bfseries 44.8 & 36.0 &  54.7 & \bfseries 88.8 & 53.1 & \bfseries 91.7 & \bfseries 59.6 &  24.8 & \bfseries 88.0 & 35.0 & \bfseries 52.5 & 18.4 & 9.6 & \bfseries 57.7 & \bfseries 54.9\\
\midrule
PDFNet3-2S& 95.0 & 69.4 & 84.4 & 25.4 & 35.0 & 38.8 & 23.5 & 42.8 & 87.1 & 48.5 & 90.0 & 50.8 & 1.9 & 83.9 & 19.0 & 39.6 & 24.9 & 8.5 & 50.3 & 48.4\\
PDFNet6-2S& 94.7 & 69.2 & 85.0 & 33.9 & 33.1 & 38.7 & 28.5 & 47.9 & 87.5 & 51.0 & 91.1 & 54.4 & 14.4 & 85.3 &  21.0 & 33.6 & 13.8 & 8.2 & 51.6 & 49.6\\
PDFNet9-2S& 95.1 & 71.2 & 85.4 & 30.9 & 32.6 & 41.1 & 31.0 & 47.9 & 87.5 & 53.3 & 90.9 & 58.2 & \bfseries 25.5 & 86.6 &  \bfseries 37.5 & 50.2 & \bfseries 31.5 & 12.1 & 52.2 & 53.7\\
PDFNet12-2S& 95.2 & 72.3 & 85.6 & \bfseries 36.2 & 32.9 & 42.2 & 32.5 & 49.1 & 87.8 & 52.6 & 90.8 & 57.6 & 20.4 & 86.4 &  33.2 & 45.5 & 19.0 & 14.1 & 52.3 & 52.9\\
\bottomrule
\end{tabular}
\end{adjustbox}
\end{center}
\caption{Class-wise results of the Cityscapes baseline experiments evaluated on the validation set}
\label{table-1}
\end{table*}

\subsubsection{Data ablation study}
\label{4.2.3}
To evaluate the performance of our method on few data samples, we conduct a data ablation study on the Cityscapes dataset. The size of the baseline networks ResNet-101 \cite{he2016deep} ($59.3$M), DenseNet-161 \cite{huang2017densely} ($43.2$M), HRNet-V2 \cite{SunZJCXLMWLW19} ($65.9$M), and U-Net \cite{ronneberger2015u} ($31.0$M) is huge compared to PDFNet variants (PDFNet12 ($1.2$M), PDFNet9 ($758$K), PDFNet6 ($405$K), and PDFNet3 ($164$K)). \newline
The less size of networks might strongly benefit while training on few data samples. Hence, we also add the DeeplabV3+ \cite{chen2018encoder} with several light-weight encoder networks such as ResNet-18 \cite{he2016deep}, MobileNet-V2 \cite{mobileNetV2}, EfficientNet-b1 \cite{TanL19}, and RegNetY-08 \cite{RegNet} in this data ablation study along with PDFNet-2S variants. \newline
We train each network on five different subsets of training data $T_{1487}$, $T_{743}$, $T_{371}$, $T_{185}$, and $T_{92}$ (the number in suffix represents the training samples in each set) by using the same val set. In Table \ref{Table2}, we present mean IoU results of the networks evaluated on validation set, parameters, and GFLOPs \cite{GFLOPS} (calculated with an input resolution of $1$x$512$x$1024$x$3$). From Table \ref{Table2}, we observe that:\newline
(i) Every PDFNet and PDFNet-2S variant outperforms the baseline networks by a considerable margin \footnote{Here, one might argue that the network size of the PDFNet variants is very less compared to any baseline network. We provide additional experiments in the supplementary material, to show that the performance gains of PDFNet on few data samples are due to the proposed architecture and not due to the less size of the network.} and PDFNet12 achieves the top average IoU score ($40.0$). \newline
(ii) Every PDFNet-2S variant under-performed while compared to every PDFNet variant. \newline
(iii) The PDFNet3-2S requires fewer parameters, and GFLOPs than other networks.\newline
(iv) Even though the size of U-Net ($31.0$M) \cite{ronneberger2015u} is larger than the lightweight encoder networks (ResNet-18 ($12.3$M) \cite{he2016deep}, MobileNet-V2 ($4.4$M) \cite{mobileNetV2}, EfficientNet-b1 ($7.4$M) \cite{TanL19}, and RegNetY-08 ($7.0$M) \cite{RegNet}. The U-Net  \cite{ronneberger2015u} still outperforms other baseline networks in the average IOU score. \newline
Here, we observe that the performance of the networks trained on few data samples correlates more with the selected architecture rather than network size.\newline
From the above, we found some evidence (ii) to support our strided convolution hypothesis (\ref{3.2}) in this few data regime.\newline
Additionally, In Table \ref{table-1}, even though the M.IoU score difference between HRNet-V2 \cite{SunZJCXLMWLW19} and U-Net \cite{ronneberger2015u} is only $1.3$, in Table \ref{Table2}  its is $11.3$ ($T_{Avg}$). This hints that performance of HRNet-V2 \cite{SunZJCXLMWLW19} might highly depend upon the training data.
\begin{table*}[t!]
\sisetup{detect-weight=true,detect-inline-weight=math}
\begin{center}
\begin{adjustbox}{width=1\textwidth}
  \begin{tabular}{lSSSSSSSS}
    \toprule
     {Backbone} &
     {$T_{1487}$} & {$T_{743}$} & {$T_{371}$} & {$T_{185}$} & {$T_{92}$} & {\textbf{$T_{avg}$}} & {Param(M)} &{GFLOPS}  \\
      \midrule
    ResNet-18  \cite{he2016deep} & 42.6 & 35.6 & 27.9 & 22.4 & 21.0 & 29.9 & 12.3 & 36.8 \\
    MobileNet-V2 \cite{mobileNetV2} & 38.5 & 32.2 & 30.6 & 22.5 & 19.2 & 28.6 &  4.4& 12.3 \\
    EfficientNet-b1 \cite{TanL19} & 37.8 & 32.5 & 26.9 & 24.6 & 19.8 & 28.3 & 7.4 &  4.6 \\
    RegNetY-08 \cite{RegNet} &  28.5& 31.9 & 29.4 & 27.4 & 22.1 & 27.9  & 7.0 & 17.2 \\
    ResNet-101 \cite{he2016deep} & 29.3 & 28.8 & 28.6 & 21.6 & 19.4 & 25.5 & 59.3 & 177.8 \\
    DenseNet-161 \cite{huang2017densely} & 33.3 & 30.1 & 26.0 & 24.9 & 20.8 & 27.0 & 43.2 & 129.4 \\
    HRNet-V2 \cite{SunZJCXLMWLW19} & 27.8 & 18.8 & 23.3 & 18.3 & 15.4 & 20.7 & 65.9 & 187.8 \\
    U-Net\cite{ronneberger2015u} & 42.8 & 34.2 & 30.2 & 27.8 & 25.0 & 32.0 & 31.0 & 387.1 \\
    PDFNet3 &  47.1 & 42.9 & 39.9 & 34.7 & 29.7 &  38.9  &  0.2  & 8.0 \\
    PDFNet6 & 49.6 & 43.2 & 39.1 & \bfseries 35.8 & \bfseries 30.5 &   39.6  & 0.4 & 10.3 \\
    PDFNet9 &  48.8 & 44.9 & \bfseries 40.0 &  34.1 & 29.3 & 39.4  & 0.8 & 13.5 \\
    PDFNet12 &  \bfseries 50.8 &  \bfseries 46.6 & 38.5 &  33.8 & 30.3 &  \bfseries 40.0  & 1.2 & 17.5 \\
\midrule
    PDFNet3-2S &  44.7 &  40.0 & 34.0 &  33.4 & 29.9 &  36.4  & \bfseries 0.1 & \bfseries 2.7 \\
    PDFNet6-2S &  46.9 &  40.5 & 38.2 &  32.6 & 24.3 &  36.5  & 0.3 &  4.5 \\
    PDFNet9-2S &  47.3 &  43.7 & 35.9 &  32.9 & 27.6 &  37.5  & 0.7 & 7.5 \\
    PDFNet12-2S &  46.0 &  40.1 & 36.9 &  31.4 & 28.0 &  36.5  & 1.1 & 11.1 \\
    \bottomrule 
  \end{tabular}
\end{adjustbox}
\end{center}
\caption{Cityscapes data ablation experiments evaluated on the validation set}
 \label{Table2}
\end{table*}

\subsection{Experiments on CamVid}
\label{4.3}
\subsubsection{Dataset}
The CamVid dataset \cite{BrostowSFC:ECCV08} for semantic segmentation consists of $700$ images divided into three sets $367$ training, $101$ validation, and $233$ testing. By following \cite{badrinarayanan2017segnet,segnet}, we use $12$ classes (including background) and resize the images from $720$x$960$ to $368$x$480$.
\subsubsection{Baseline experiments}
\label{4.3.2}
We conduct baseline experiments on CamVid dataset using three different sets, divided according to the networks size.\newline
Set-1 consists of DeeplabV3+ \cite{chen2018encoder}  with encoders networks, Resnet-18 \cite{he2016deep}, EfficientNet-b1 \cite{TanL19}, RegNetY-08 \cite{RegNet}, MobileNet-V2 \cite{mobileNetV2}, PDFNet3, PDFNet3-2S, PDFNet6, and PDFNet6-2S.\newline
Set-2 consists of DeeplabV3+ \cite{chen2018encoder}  with encoder networks, Resnet-50 \cite{he2016deep}, EfficientNet-b4 \cite{TanL19}, RegNetY-40 \cite{RegNet}, ResNext-50 \cite{Resnext}, U-Net, PDFNet9, and PDFNet9-2S.\newline
Set-3 consists of DeeplabV3+ \cite{chen2018encoder} with encoder networks, Resnet-101 \cite{he2016deep}, EfficientNet-b6 \cite{TanL19}, RegNetY-80 \cite{RegNet}, DeepLabV3 (DenseNet-161 \cite{huang2017densely}), HRNet-V2, PDFNet12, and PDFNet12-2s.

\begin{table*}[t!]
 \sisetup{detect-weight=true,detect-inline-weight=math}
\begin{center}
 \begin{adjustbox}{width=1\textwidth}
  \begin{tabular}{lSSSSSSSSSS}
    \toprule
    \multirow{2}{*}{Backbone} &
      \multicolumn{2}{c}{$T_{367}$} &
      \multicolumn{2}{c}{$T_{183}$} &
      \multicolumn{2}{c}{$T_{91}$} &
      \multicolumn{2}{c}{$T_{avg}$} \\
      \cmidrule(r){2-3}
      \cmidrule(r){4-5}
      \cmidrule(r){6-7}
      \cmidrule(r){8-9}
       & {Val} & {Test}  & {Val} & {Test}& {Val} & {Test}& {Val} & {Test} &{Param(M)} & {GFLOPS}  \\
      \midrule
     ResNet-18 \cite{he2016deep} & 83.3 & 64.9 & 79.7 &  63.7 & 70.0 & 56.6 & 77.7 &61.7  & 12.3 & 12.4 \\
     EfficientNet-b1 \cite{TanL19}& \bfseries 84.4 &  \bfseries 68.4 & 75.0 & 61.3 & 77.0 & 58.8 &  78.8& 62.8 & 7.4 &  1.5 \\
     RegNetY-08 \cite{RegNet}& 80.4 & 64.3 & 77.7 & 61.4 & 70.9 & 57.8 & 76.3 &61.2 & 7.0 &5.8  \\
     MobileNet-V2 \cite{mobileNetV2}& 80.8 & 63.9 & 77.3 & 56.1 & 64.4 & 57.0 & 74.2 &59.0  &4.4 & 4.1 \\
     PDFNet3 & 83.3 &  67.5 & \bfseries 82.6  & 63.0  &  \bfseries 80.7  & \bfseries 64.5 &\bfseries  82.2 &  65.0  & 0.2 & 2.3\\
     PDFNet3-2S & 83.1 &  67.4 & 81.8  & 61.7  &  78.6  & 62.0 &   81.2 &  63.7  & \bfseries 0.1 & \bfseries 0.8\\
     PDFNet6 & 84.0 &  68.0 &  82.2  & \bfseries 66.8  &  80.2  &  62.5 &  82.1 & \bfseries 65.8  & 0.4 & 3.0 \\
     PDFNet6-2S & 82.2 &  65.9 &  80.5  & 64.0  &  77.9  &  59.9 &  80.2 & 63.3  & 0.3 & 1.4 \\
     \midrule
     ResNet-50 \cite{he2016deep}& 78.6 & 61.6 &79.6  &60.3  & 78.3  &55.9 &78.8  & 59.3  & 26.7 & 25.0\\
     EfficientNet-b4 \cite{TanL19}&82.7  &64.1  &77.7  &62.2  &75.6  &  60.5 & 78.7 & 62.3 &18.6 & \bfseries 1.7 \\
     RegNetY-40 \cite{RegNet}& 80.8 & 63.8 & 76.4 & 61.0 & 74.9 & 59.2 & 77.4 &61.3 &21.5 & 18.8   \\
     ResNext-50 \cite{Resnext}& 80.1 & 62.6 & 81.0 &63.9   & 77.9  &  59.9 &79.7& 62.1 & 26.2 & 25.0  \\
     U-Net \cite{ronneberger2015u} & 83.0 & \bfseries 69.5 & 78.0 &62.8  & 76.8  & 61.6  &79.3& 64.6 & 31.0 & 130.0  \\
     PDFNet9 & \bfseries 83.8  &  67.4 & \bfseries 83.5 &  \bfseries 66.0  & \bfseries 82.0  & \bfseries 64.7 & \bfseries 83.1 & \bfseries 66.0 &  0.7 & 4.1 \\
     PDFNet9-2S & 82.5  &  66.0 & 80.9 &  64.4  & 78.1  & 61.7 & 80.5 & 64.0 & \bfseries 0.6 & 2.4 \\
     \midrule
     ResNet-101 \cite{he2016deep}& 81.6 & 63.8 &75.6 & 56.4  & 70.1 & 55.7  & 75.8 &58.6 & 59.3 & 59.9   \\
     EfficientNet-b6 \cite{TanL19}& 80.6 &65.0  & 80.3 & 57.8 &  77.4 & 60.4 & 79.4 & 61.0 & 42.0 & \bfseries 1.9  \\
     RegNetY-80 \cite{RegNet}& 78.5 & 62.0 & 78.2 &63.8  & 66.2 & 53.8 &  74.3&  59.9 & 40.3 & 34.4 \\
     DenseNet-161 \cite{huang2017densely}&77.8  & 58.6 & 75.5 & 57.7 & 73.0 &53.8  & 75.4 &56.7 & 43.2 & 43.6    \\
     HRNet-V2 \cite{SunZJCXLMWLW19} & 81.1  & 63.6 & 79.1 & 62.9 & 72.9 & 55.0  & 77.7 & 60.5 & 65.9 & 63.5    \\
     PDFNet12 & \bfseries 84.3 & \bfseries 68.0 & \bfseries 82.0 & \bfseries  67.0 &  \bfseries 80.9  & \bfseries 60.9 & \bfseries 82.4  & \bfseries 65.3 &  1.2 & 5.4   \\
     PDFNet12-2S & 83.1  & 65.5 & 80.9 & 62.4 & 79.5 & 60.2  & 81.2 & 62.7 & \bfseries 1.1 & 3.6    \\
    \bottomrule
  \end{tabular}
  \end{adjustbox}
  \end{center}
   \caption{Camvid baseline experiments evaluated on the validation and the test sets}
  \label{table-3}
\end{table*}

\noindent We train each network on three subsets of the training set ($T_{367}$, $T_{183}$, and $T_{91}$) by using the same validation set and test set. 
In Table \ref{table-3}, we present M.IoU score on validation set, test test, parameters, and GFLOPS \cite{GFLOPS} (calculated with an input resolution of $1$x$368$x$480$x$3$).\newline
From Table \ref{table-3}, we observe that: \newline
(i) In all three sets, PDFNet variants outperforms other networks in  $T_{183}$, $T_{91}$ and  $T_{Avg}$, and require fewer parameters and GFLOPs.\newline
(ii) With in the networks trained on subset  $T_{367}$, EfficientNet-b1 \cite{TanL19} in Set-1, U-Net \cite{ronneberger2015u} in Set-2, and PDFNet12 in Set-3 achieves top performance.\newline
(iii) In all three sets, PDFNet variants significantly outperforms PDFNet-2S variants in  $T_{183}$, $T_{91}$ and  $T_{Avg}$. \newline
From the above observations, we found a similar pattern as Table \ref{Table2} that supports our strided convolution hypothesis (\ref{3.2}) in the less data regimes. 
\subsection{Generalization on KITTI}

\subsubsection{Dataset}
The KITTI semantic segmentation dataset \cite{Alhaija2018IJCV} consists of 400 images divided into 200 training and 200 testing sets.

\subsubsection{Cityscapes to KITTI}

\begin{table}[t!]
 \sisetup{detect-weight=true,detect-inline-weight=math}
\begin{center}
 \begin{adjustbox}{width=1\textwidth}
  \begin{tabular}{lSSSSSSSSSSSS}
    \toprule
    \multirow{2}{*}{Backbone} &
      \multicolumn{2}{c}{$T_{1487}$} &
      \multicolumn{2}{c}{$T_{743}$} &
      \multicolumn{2}{c}{$T_{371}$} &
      \multicolumn{2}{c}{$T_{185}$}&
      \multicolumn{2}{c}{$T_{92}$}& 
      \multicolumn{2}{c}{$T_{avg}$}\\
      \cmidrule(r){2-3}
      \cmidrule(r){4-5}
      \cmidrule(r){6-7}
      \cmidrule(r){8-9}
      \cmidrule(r){10-11}
      \cmidrule(r){12-13}
      & {Class} & {Cat}  &  {Class} & {Cat} &  {Class} & {Cat}&  {Class} & {Cat} &   {Class} & {Cat} &   {Class} & {Cat} \\
      \midrule
     ResNet-18 \cite{he2016deep} & 15.1 & 30.8 & 12.0 &  27.0 & 10.5 & 25.2 & 7.9 & 20.4  & 9.8 & 24.7 &11.1 & 25.6 \\
     MobileNet-V2 \cite{mobileNetV2}& \bfseries 21.6 & \bfseries 42.6 & 17.2 & \bfseries 38.5 & 15.8 & \bfseries 37.6 & 13.2 & 30.9  & 13.6 & 32.9 & 16.3 & 36.6 \\
     EfficientNet-b1 \cite{TanL19}&  19.5 &  42.2 & 15.2 & 36.5 & 14.2 & 34.2 &  8.1 & 20.9& 6.1 & 15.4 & 12.6& 29.8\\
     RegNetY-08 \cite{RegNet}& 10.8 & 26.6 & 12.9 & 30.4 & 12.6 & 29.8 & 11.5 & 27.6 & 10.1 & 24.0 & 11.6 & 27.7   \\
    ResNet-101 \cite{he2016deep} & 12.0 & 26.4 & 11.6 & 28.3 & 8.8 & 21.3 & 8.8 & 22.3 & 7.5& 19.0 &9.7   & 23.5\\
    DenseNet-161 \cite{huang2017densely} & 10.5 & 24.0 & 8.0 & 19.8 & 9.7 & 23.3 & 11.2 & 27.0& 8.4 & 21.7 & 9.6 & 23.2\\
     HRNet-V2 \cite{SunZJCXLMWLW19} & 6.9 & 16.8 & 8.1 & 19.9 & 7.9 & 20.6 & 9.3 & 24.0 & 11.2& 29.5 & 8.7   & 22.2\\
    U-Net \cite{ronneberger2015u}  & 14.1 & 31.8 & 12.8 & 30.5 & 9.3 & 23.2 & 9.0 & 22.6 & 7.9 & 20.6 & 10.6 & 25.7\\
     PDFNet3 & 17.2 &  36.8 &  13.0  & 28.3  &  16.0  & 34.4 & 12.3 &  28.4  & 15.0  & 33.0 & 14.7 & 32.2 \\
     PDFNet6 & 18.7 & 38.2 & 15.2  & 32.5  & 15.2  & 34.4 & 14.4  & 30.8 & 16.4 & 35.7 & 16.0 & 34.3\\
     PDFNet9 & 20.9 &  39.2 &  \bfseries 17.3  & 35.6  &  15.5  & 33.4 & 14.4 & 32.2  & 14.6 & 32.9 & 16.5 &34.7 \\
     PDFNet12 & 18.3 &  37.3 &  17.1  & 35.5  & \bfseries 17.8  & 36.7 & \bfseries 16.0 & \bfseries  36.7  & \bfseries 17.4 & \bfseries 40.0 & \bfseries 17.3 & \bfseries 37.2\\
     \midrule
     PDFNet3-2S & 14.3 &  29.5 &  12.1  & 27.4  &  13.3  & 30.5 & 12.7 &  29.4  & 14.3  & 32.2 & 13.4 & 29.8 \\
     PDFNet6-2S & 19.4 &  36.7 &  14.8  & 33.2  &  14.5  & 32.5 & 13.9 &  30.5  & 13.8  & 32.6 & 15.8 & 33.1 \\
     PDFNet9-2S & 16.7 &  35.1 &  15.4  & 34.6  &  16.9  & 37.4 & 14.5 &  32.9  & 14.1  & 33.1 & 15.5 & 34.6 \\
     PDFNet12-2S & 18.6 &  39.1 &  16.3  & 34.6  &  16.9  & 36.5 & 14.6 &  34.7  & 13.4  & 32.3 & 16.0 & 35.4 \\
  \bottomrule
  \end{tabular}
  \end{adjustbox}
  \end{center}
   \caption{Cityscapes data ablation experiments evaluated on the KITTI training set}
  \label{table-4}
\end{table}
Since the data format and metrics of KITTI dataset are consistent with Cityscapes \cite{Alhaija2018IJCV}, we use it to evaluate the out-of-training distribution performance of the networks trained on Cityscapes dataset. \newline
In Table \ref{table-4}, we provide mean class and category IoU scores of the networks from  Cityscapes data ablation study evaluated on KIITTI training set. From Table \ref{table-4}, we observe that:\newline
(i) The PDFNet12 outperforms other networks in average class and category IoU score.\newline
(ii) The PDFNet-2S variants performed almost similar to PDFNet variants.\newline
(iii) Within the baseline networks, the MobileNet-V2 \cite{mobileNetV2} shown considerable generalization performance.\newline
Here, we hypothesize that the generalization performance of PDFNet variants is due to the captured generalized features than dataset-specific features. 
\section{Conclusion}
\noindent (i) In this work, we introduced a novel lightweight framework for urban scene segmentation named PDFNet to deal with small classes and few data samples. \newline
The extensive experiments on Cityscapes and CamVid benchmarks demonstrate the effectiveness of our method compared to baselines. Moreover, our method achieves considerable generalization performance in labeling out-of-training distribution samples.\newline
(ii) In this work, we found that replacing the strided Conv with regular Conv in the first layer (or the stem module) of the networks might help in dealing with few data samples. \newline
In other words, we observe that our strided convolution hypothesis (\ref{3.2}) is valid in the few data regimes.\newline
The (i) and (ii) are the main contributions of this work.\newline
Improving the performance on small classes, few data samples and real-time applicability on several hardware platforms will be studied in future work.
\bibliographystyle{unsrt} 
\bibliography{main.bib}

\newpage
\appendix
\section{Appendix}
\subsection{Extended data ablation study}
\begin{figure}[h!]
\begin{center}
   \includegraphics[width=1\linewidth]{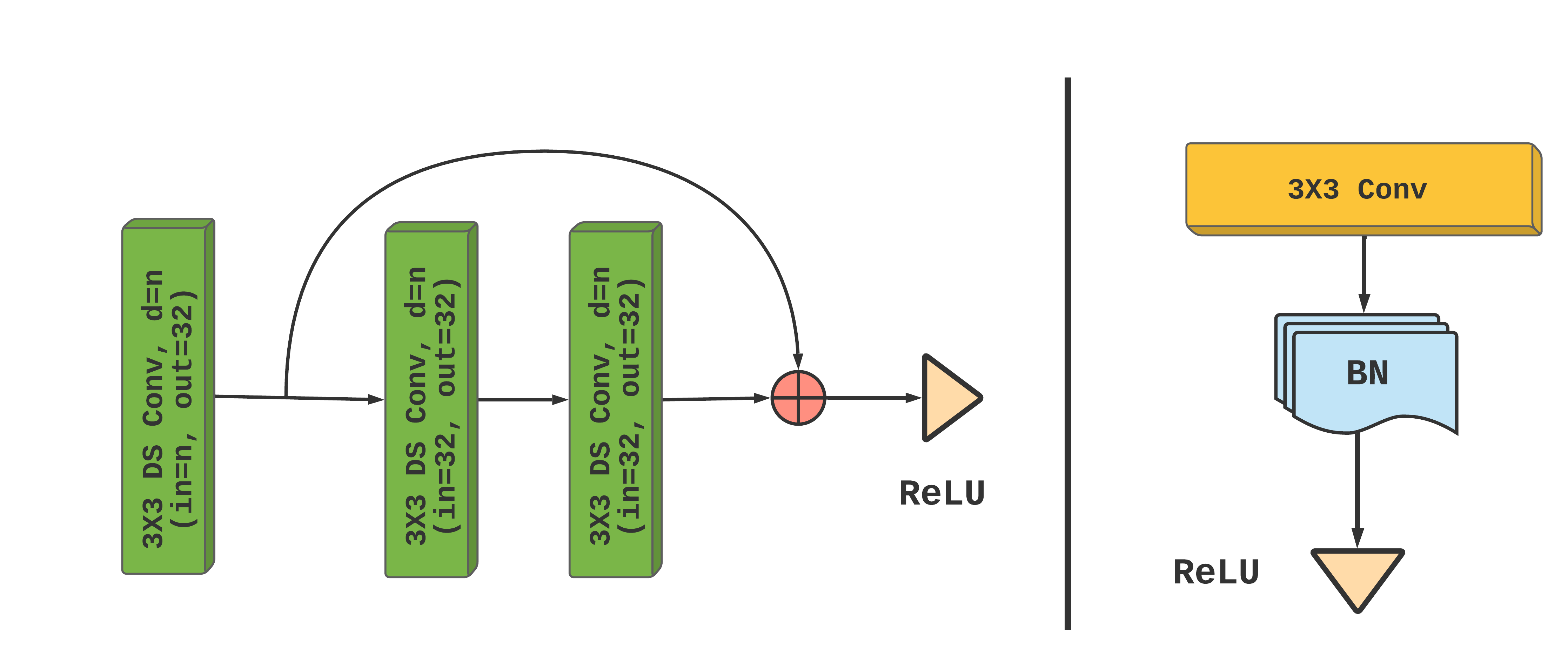}
\end{center}
   \caption{The glance module of PDFNet}
\label{PDF-glance}
\end{figure}
\begin{figure}[h!]
\begin{center}
   \includegraphics[width=1\linewidth]{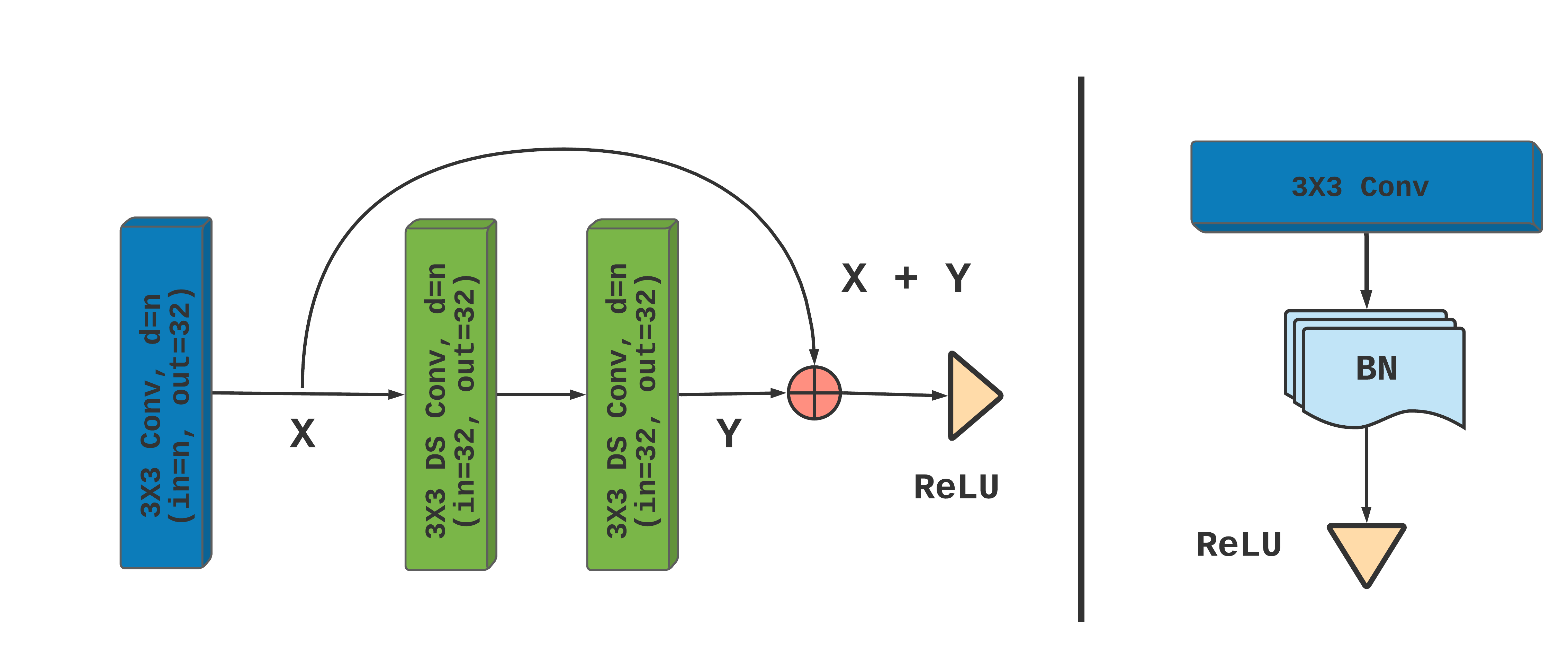}
\end{center}
   \caption{The glance module of DFNet}
\label{DF-glance}
\end{figure}
\noindent The PDFNet variants (PDFNet12 ($1.2$M), PDFNet9 ($758$K), PDFNet6 ($405$K), and PDFNet3 ($164$K)) consists of fewer parameters than any baseline network.
Here, one might argue that the network size of the PDFNet variants is very less compared to any baseline network.\newline
Hence, we conduct additional experiments to verify whether the performance gains of PDFNet on few data samples are due to the proposed architecture or not.\newline
\noindent In Figure \ref{PDF-glance}, we present the glance module of PDFNet. We modified the PDFNet-glance module by replacing the first Conv layer with a regular Conv layer, as shown in Figure \ref{DF-glance}.\newline
We design DFNet variants (DFNet12 ($7.6$M), DFNet9 ($4.6$M), DFNet6 ($2.3$M), and DFNet3 ($0.9$M)) by replacing the original glance module with modified glance module in PDFNet variants.\newline
We train each network on five different subsets of training data $T_{1487}$, $T_{743}$, $T_{371}$, $T_{185}$, and $T_{92}$ and provide the results in Table \ref{dblation} (including results from the main paper).\newline
From Table \ref{dblation}, we observe that:\newline
(i) The DFNet variants outperformed all the baseline networks.\newline
(ii) The PDFNet variants outperforms DFNet variants while requiring fewer parameters and GFLOPs.\newline
From the above, we verify that the performance gains of PDFNet on few data samples are due to the proposed architecture and not due to the less size of the network.\newline
Moreover, we provide the class-wise IoU results of the data ablation experiments in Table \ref{a-table-2}.
\begin{table*}[ht]
\sisetup{detect-weight=true,detect-inline-weight=math}
\begin{center}
\begin{adjustbox}{width=1\textwidth}
  \begin{tabular}{lSSSSSSSS}
    \toprule
     {Backbone} &
     {$T_{1487}$} & {$T_{743}$} & {$T_{371}$} & {$T_{185}$} & {$T_{92}$} & {\textbf{$T_{avg}$}} & {Param(M)} &{GFLOPS}  \\
      \midrule
    ResNet-18  & 42.6 & 35.6 & 27.9 & 22.4 & 21.0 & 29.9 & 12.3 & 36.8 \\
    MobileNet-V2  & 38.5 & 32.2 & 30.6 & 22.5 & 19.2 & 28.6 &  4.4& 12.3 \\
    EfficientNet-b1 & 37.8 & 32.5 & 26.9 & 24.6 & 19.8 & 28.3 & 7.4 &  4.6 \\
    RegNetY-08  &  28.5& 31.9 & 29.4 & 27.4 & 22.1 & 27.9  & 7.0 & 17.2 \\
    ResNet-101  & 29.3 & 28.8 & 28.6 & 21.6 & 19.4 & 25.5 & 59.3 & 177.8 \\
    DenseNet-161  & 33.3 & 30.1 & 26.0 & 24.9 & 20.8 & 27.0 & 43.2 & 129.4 \\
    HRNet-V2  & 27.8 & 18.8 & 23.3 & 18.3 & 15.4 & 20.7 & 65.9 & 187.8 \\
    U-Net & 42.8 & 34.2 & 30.2 & 27.8 & 25.0 & 32.0 & 31.0 & 387.1 \\
    PDFNet3 &  47.1 & 42.9 & 39.9 & 34.7 & 29.7 &  38.9  &  0.2  & 8.0 \\
    PDFNet6 & 49.6 & 43.2 & 39.1 & \bfseries 35.8 & \bfseries 30.5 &   39.6  & 0.4 & 10.3 \\
    PDFNet9 &  48.8 & 44.9 & \bfseries 40.0 &  34.1 & 29.3 & 39.4  & 0.8 & 13.5 \\
    PDFNet12 &   50.8 &  \bfseries 46.6 & 38.5 &  33.8 & 30.3 &  \bfseries 40.0  & 1.2 & 17.5 \\
\midrule
    PDFNet3-2S &  44.7 &  40.0 & 34.0 &  33.4 & 29.9 &  36.4  & \bfseries 0.1 & \bfseries 2.7 \\
    PDFNet6-2S &  46.9 &  40.5 & 38.2 &  32.6 & 24.3 &  36.5  & 0.3 &  4.5 \\
    PDFNet9-2S &  47.3 &  43.7 & 35.9 &  32.9 & 27.6 &  37.5  & 0.7 & 7.5 \\
    PDFNet12-2S &  46.0 &  40.1 & 36.9 &  31.4 & 28.0 &  36.5  & 1.1 & 11.1 \\
\midrule
    DFNet3 &  48.8 & 44.4 & 35.9 & 32.5 & 28.5 &  38.0  &  0.9  & 28.1 \\
    DFNet6 & 50.9 & \bfseries 46.6 & 37.1 & 34.0 & 28.6 &  39.4  & 2.3 & 42.1 \\
    DFNet9 & \bfseries 52.2 & 43.0 & 36.9 &  34.1 & 28.7 & 39.0  & 4.6 & 61.7 \\
    DFNet12 &  49.4 &  44.4 & 38.2 &  33.5 &  29.2 &  38.9  & 7.6 & 86.9 \\

    \bottomrule 
  \end{tabular}
\end{adjustbox}
\end{center}
\caption{Cityscapes additional data ablation experiments evaluated on the validation set}
 \label{dblation}
\end{table*}
\begin{table*}
\sisetup{detect-weight=true,detect-inline-weight=math}
\begin{center}
\begin{adjustbox}{width=1\textwidth}
\begin{tabular}{SlSSSSSSSSSSSSSSSSSSSS}
 \toprule
  {Subset}& {Method} & \rot{90}{road} & \rot{90}{sidewalk} &\rot{90}{building}  & \rot{90}{wall} &\rot{90}{fence}  & \rot{90}{pole} &\rot{90}{traffic light}  & \rot{90}{traffic sign} &\rot{90}{vegetation}  & \rot{90}{terrain} &\rot{90}{sky}  & \rot{90}{person} &\rot{90}{rider}  & \rot{90}{car} &\rot{90}{truck}  & \rot{90}{bus} &\rot{90}{train}  & \rot{90}{motorcycle} &\rot{90}{bicycle}  & \rot{90}{Average}      \\
  \midrule
   &ResNet18& 93.7 & 64.4 & 81.5& 14.5 & 13.8 & 27.8 & 17.8 & 26.3 & 85.0 &46.2 & 88.8 &  46.7 &  7.4 & 81.3 & 23.8 & 34.5 & 10.1 & 5.0 & 39.8 & 42.6\\
  &MobileNetV2& 93.9 & 64.6 & 81.8& 15.8 & 16.2 & 24.0 & 1.0 &17.6 & 84.4 & 39.9 & 88.6 & 39.2 & 0.0 & 82.5 & 13.0 & 26.8 & 9.0 & 0.0 & 32.6 & 38.5\\
   &EfficientNetb1& 93.3 & 64.9 & 81.7& 0.3 & 15.0 & 26.2 & 2.5 & 23.8 & 83.7 & 41.4 & 88.6 & 42.3 & 0.0 & 81.0 & 0.0 & 25.7 &  17.0 &0.0 &30.7 & 37.8\\
   &RegNetY08&  94.1 & 57.7 & 77.6& 0.0 & 0.3 & 0.0 & 0.0 & 0.0 & 82.0 & 41.4 & 88.4 & 25.7 & 0.0 & 75.0 &0.0 & 0.0 & 0.0 & 0.0 & 0.0 & 28.5\\
    & ResNet101-1& 93.0 & 54.0 & 75.9& 12.9 & 0.9 & 3.4 & 0.0 & 0.2 & 81.7 & 38.8 & 86.9 & 30.9 & 0.0 & 73.7 & 2.7 & 0.0 & 0.0 & 0.0 & 1.0 & 29.3\\
   &DenseNet161-1& 93.1 & 57.7 & 78.9& 11.2 & 6.6 & 11.1 & 0.0 & 12.4 & 81.5 & 38.3 & 84.9 & 35.0 & 0.0 & 76.7 & 13.4 & 0.3 & 2.3 & 0.1 & 28.3 & 33.3\\
   &HRNet-V2& 92.3 & 52.3 & 76.9 &  0.7 & 0.5 & 0.0 & 0.0 & 0.0 & 81.8 & 34.5 & 82.9 & 9.4 & 0.0 & 72.3 & 0.0 & 5.3 & 0.0 & 0.0 & 18.6 & 27.8\\
 $T_{1487}$  &U-Net& 94.0 & 65.6 & 83.2 &  13.9 & 20.5 & 34.3 & 21.7 & 43.9 & 87.4 & 43.1 & 89.5 & 49.9 & 0.0 & 84.2 & 12.1 & 9.9 & 12.9 & 0.0 & 47.9 & 42.8\\
   &PDFNet3& 93.0 & 68.2 & 84.7 &  25.2 & \bfseries 33.6 & 39.2 & 25.4 & 44.6 & 87.1 & 48.7 & \bfseries 90.4 & 49.6 & 6.1 & 84.5 & 15.2 & 31.8 & 13.1 & 5.3 & 48.6 & 47.1 \\
   &PDFNet6& 93.6 & 69.6 & 84.7 &  25.9 & 32.1 & 40.1 & 29.2 & 47.2 & 87.1 & 49.2 & 90.3 &  54.2 & \bfseries 18.2 & 84.8 & 20.5 & 37.6 & 18.6 & 5.9 & \bfseries 53.0 & 49.6 \\
   &PDFNet9& 93.7 & 70.5 & \bfseries 85.1 &  23.2 & 31.8 & 39.4 & \bfseries 30.2 & 44.9 & 87.1 & 47.3 & 90.3 & 52.6 & 14.1 &  85.4 & 16.0 & 39.2 & \bfseries 23.5 & 2.8 & 50.2 & 48.8 \\
   &PDFNet12& \bfseries 95.2 & \bfseries 71.0 &  84.9 &   31.9 & 32.2 & \bfseries 41.6 & 28.9 & \bfseries 47.8 & \bfseries 87.6 & \bfseries 51.4 & 90.2 & \bfseries 55.4 & 15.1 & \bfseries 85.9 & \bfseries 26.0 & \bfseries 43.1 & 18.1 & \bfseries 8.4 & 50.2 & \bfseries 50.8 \\
   &PDFNet3-2S& 94.5 & 67.0 & 83.3 &  15.8 & 25.5 & 35.0 & 22.7 & 39.7 & 86.1 & 44.2 & 89.7 & 48.2 & 4.9 & 82.6 & 15.0 & 35.4 & 8.3 & 6.4 & 46.0 & 44.7\\
   &PDFNet6-2S& 93.5 & 68.3 & 83.2 &  20.7 & 33.0 & 38.0 & 24.5 & 40.4 & 85.8 & 48.1 & 88.6 & 51.1 & 10.9 & 83.5 & 19.7 & 29.2 & 19.3 & 5.7 & 47.7 & 46.9 \\
   &PDFNet9-2S& 94.2 & 68.4 &  83.9 & \bfseries 33.3 & 30.8 & 37.8 & 16.9 & 38.8 & 86.5 & 48.6 & 89.8 & 49.9 & 6.8 & 84.1 & 20.8 & 31.7 & 19.9 & 7.4 & 48.6 & 47.3\\
   &PDFNet12-2s& 94.2 & 66.0 & 84.0 &  21.6 & 27.9 & 32.9 & 21.1 & 40.7 & 86.5 & 49.1 & 89.1 & 48.6 & 4.6 & 83.8 & 18.9 & 38.9 & 16.1 & 4.9 & 44.1 & 46.0 \\

    \midrule
   &ResNet18& 92.8 & 56.3 & 78.0&  15.3 & 8.0 & 15.4 & 4.9 & 18.9 & 82.3 & 42.8 & 85.6 & 35.4 & 0.1 & 75.2 &  13.1 & 13.9 & 1.9 & 0.0 & 36.6 & 35.6\\
   &MobileNetV2& 92.7 & 57.3 & 77.8& 6.1 & 7.8 & 0.7 & 0.1 & 11.3 & 81.5 & 39.2 & 85.2 & 30.8 & 0.1 & 75.9 & 3.5 &  22.9 & 2.9 & 0.0 & 15.4 & 32.2\\
   &EfficientNetb1& 93.5 & 60.5 & 77.1&4.1 & 3.9 & 9.1 & 0.0 & 14.0 & 81.8 & 39.6 & 84.6 & 22.8 &  1.6 & 75.1 & 9.8 & 18.6 & 0.0 & 0.0 & 20.8 & 32.5\\
   &RegNetY08& 93.9 & 58.7 & 78.8& 3.4 & 9.1 & 0.0 & 0.0 & 17.5 & 83.1 &  45.2 & 87.1 & 32.0 & 0.0 & 76.8 & 2.3 & 0.3 & 0.0 & 0.0 & 17.1 & 31.9\\
  &ResNet101& 90.5 & 44.6 & 72.2& 9.2 & 3.3 & 5.2 & 0.0 & 12.5 & 79.8 & 36.2 & 79.8 & 25.6 & 0.0 & 65.3 & 0.0 & 0.0 & 0.0 & 0.0 & 0.0 & 28.8\\
   &DenseNet161& 91.1 & 50.8 & 74.7& 13.9 & 3.3 & 4.4 & 1.1 & 12.1 & 78.4 & 32.1 & 80.8 & 28.6 & 0.0 & 69.6 & 2.2 & 1.4 & 0.2 & 2.3 & 25.7 &30.1\\
   &HRNet-V2& 86.0 & 23.6 & 62.0 &  0.0 & 0.0 & 0.0 & 0.0 & 0.0 & 60.3 & 10.9 & 78.1 & 0.0 & 0.0 & 37.3 & 0.0 & 0.0 & 0.0 & 0.0 & 0.0 & 18.8\\
  $T_{743}$   &U-Net& 93.6 & 62.1 & 81.3&  5.2 & 11.6 & 19.3 & 0.4 & 28.1 & 86.0 & 42.3 & 87.8 & 36.5 & 0.0 & 79.6 & 9.6 & 0.5 & 0.0 & 0.0 & 5.5 & 34.2\\
   &PDFNet3& 93.8 & 63.7 & 81.6 &  17.6 & 21.1 & 30.5 & 12.3 & 31.3 & 84.2 & 42.4 &  88.2 & 43.6 & 1.0 & 79.4 & 10.4 & 19.3 & 3.4 & 1.0 & 36.1 & 40.0 \\
   &PDFNet6& \bfseries 94.0 & 65.1 & 83.2 &  17.2 & 25.9 & 34.9 & 17.5 & 38.9 & 85.9 & 44.2 & 90.0 & 46.6 & 1.2 & 81.2 & 17.1 & 24.9 & 9.6 & 0.3 & 43.7 & 43.2 \\
   &PDFNet9& 93.9 & 66.9 & 83.4 &  19.5 & 25.3 & 36.3 & 23.8 & 40.4 & 85.6 & 43.9 & 88.9 & 48.7 & 2.1 & 82.8 &  17.1 & 34.6 & \bfseries 10.3 &  2.4 & 47.5 & 44.9 \\
   &PDFNet12& 93.4 & \bfseries 68.9 & \bfseries  84.1 &  \bfseries 30.0 & \bfseries 33.0 & \bfseries 38.3 & \bfseries 26.0 & \bfseries 46.9 & \bfseries 86.5 & \bfseries 49.5 & \bfseries 89.2 & \bfseries 50.2 & \bfseries 14.2 &\bfseries  83.9 & 15.4 & \bfseries 27.5 & 2.8 & 1.7 & \bfseries 47.8 & \bfseries 46.6\\
   &PDFNet3-2S& 90.2 & 49.5 & 74.4 &  5.6 & 10.9 & 15.3 & 0.0 & 8.2 & 79.0 & 34.6 & 82.1 & 22.6 & 0.0 & 68.6 & 1.1 & 0.9 & 4.2 & 0.0 & 17.5 & 29.7\\
   &PDFNet6-2S& 93.5 & 63.4 & 81.7 &  18.6 & 17.2 & 31.2 & 13.5 & 30.0 & 85.0 & 42.5 & 88.2 & 42.6 &  0.9 & 80.5 & 8.0 &  27.2 & 4.6 & 1.5 & 39.5 &  40.5 \\
   &PDFNet9-2S& 94.0 & 65.9 &  82.6 &  27.7 & 22.4 & 34.8 & 18.0 & 35.4 & 85.5 & 45.1 & 88.9 & 47.2 & 7.7 & 82.8 & \bfseries 18.6 & 23.4 & 3.2 &\bfseries  4.4 & 43.3 & 43.7\\
    &PDFNet12-2s& 93.5  & 62.0 & 81.7 &  14.1 & 22.3 & 32.8 & 12.7 & 29.8 & 83.9 & 38.9 & 87.5 & 43.3 & 1.9 & 80.4 & 8.9 & 21.6 & 6.4 & 0.5 & 39.0 & 40.1 \\
   
   \midrule
  &ResNet18& 88.4 & 46.5 & 73.5& 2.6 & 1.1 & 4.0 & 0.0 & 5.3 & 78.4 & 36.4 & 82.9 & 27.9 & 0.0 & 68.5 & 0.0 & 0.0 & 0.0 &  1.4 & 13.1 & 27.9\\
   &MobileNetV2& 91.4 & 50.2 & 75.2& 10.1 &  7.3 & 5.6 & 0.3 & 8.5 & 81.0 & 34.6 & 83.0 & 27.8 & 0.0 & 72.7 &  \bfseries 16.0 &  1.4 & 0.0 & 0.0 & 16.6 & 30.6\\
   &EfficientNetb1& 89.8 & 46.4 & 72.4& 5.3 & 4.9 & 7.6 & 0.3 & 6.5 & 75.7 & 32.7 & 77.3 & 20.7 & 0.0 & 65.1 & 0.0 & 0.0 & 0.0 & 0.0 & 6.8 & 26.9\\
   &RegNetY08&  91.6 & 50.5 & 76.3 &  10.3 & 5.8 & 0.0 & 0.3 & 10.3 & 81.2 &  38.3 &  84.5 & 26.2 & 0.0 & 70.6 & 8.1 & 0.0 & 0.0 & 0.0 & 3.8 & 29.4\\
   &ResNet101& 89.2 & 45.8 & 73.6 & 9.8 & 2.7 & 3.1 & 0.0 & 8.0 & 79.9 & 34.8 & 81.6 & 26.9 & 0.0 & 65.3 & 8.9 & 0.0 & 0.0 & 0.0 & 13.8 & 28.6\\
   &DenseNet161& 89.0 & 42.4 & 71.9& 8.1 & 2.1 & 0.0 & 0.0 & 0.3 & 75.7 & 33.8 & 77.4 & 18.6 & 0.0 & 64.3 & 3.3 & 0.2 & 0.0 & 0.0 & 6.2 & 26.0\\
   &HRNet-V2& 87.4 & 35.3 & 68.7 &  0.1 & 0.0 & 0.0 & 0.0 & 0.0 & 74.7 & 33.9 & 76.4 & 8.9 & 0.0 & 56.2 & 0.0 & 0.0 & 0.0 & 0.0 & 0.3 & 23.3 \\
  $T_{371}$   &U-Net& 92.3 & 57.4 & 76.9 &  4.7 & 7.4 & 2.7 & 1.2 & 14.1 & 83.4 & 36.6 & 85.8 & 28.5 & 0.0 & 75.8  & 3.3 & 0.0 & 1.9 & 0.0 & 1.4 & 30.2\\
   &PDFNet3& 92.5 & 54.0 & 78.1 &  9.5 & 15.8 & 21.1 & 0.0 & 17.0 & 82.0 & 38.2 & 86.3 & 36.6 & 0.0 & 74.9 & 0.8 & 8.3 & 2.7 & 0.0 & 28.5 & 34.0 \\
   &PDFNet6& 93.2 & \bfseries 61.2 & 80.7 &  11.9 & 17.9 & 28.4 & 3.7 & 26.9 & \bfseries 84.6 &\bfseries  40.8 & 87.4 & 39.7 & 0.0 & 78.5 & 15.6 & \bfseries 30.3 & 6.9 & 0.2 & 35.7 & 39.1 \\
   &PDFNet9& 93.3 & \bfseries61.2 &  \bfseries 81.0 &  15.1 & \bfseries 19.3 & 29.3 & 8.0 & 30.0 & 83.9 & 39.8 & 87.8 & 41.3 & \bfseries 6.3 & \bfseries 79.0 & 11.5 & 26.3 & \bfseries 8.9 & 0.3 & \bfseries 37.0 & \bfseries 40.0\\
   &PDFNet12& 92.6 & 61.1 &  \bfseries 81.0 & \bfseries 24.8 & 14.9 & \bfseries 30.2 & \bfseries 9.5 & \bfseries 33.2 & 84.0 & 36.2 &\bfseries 88.0 &\bfseries  42.0 & 0.5 & 78.7 & 14.1 & 6.7 & 0.0 & \bfseries 3.3 & 30.6 & 38.5 \\
   &PDFNet3-2S& 90.2 & 49.5 & 74.4 &  5.6 & 10.9 & 15.3 & 0.0 & 8.2 & 79.0 & 34.6 & 82.1 & 22.6 & 0.0 & 68.6 & 1.1 & 0.9 & 4.2 & 0.0 & 17.5 & 29.7\\
   &PDFNet6-2S& 92.8 & 60.4 & 79.6 &  8.7 & 17.2 & 27.6 & 5.5 & 29.2 & 83.0 & 40.4 & 87.4 & 39.9 &  0.4 & 77.9 & 8.5 &  24.2 & 7.4 & 0.1 & 36.1 &  38.2 \\
   &PDFNet9-2S& 92.6 & 60.3 &  79.1 &  18.0 & 13.7 & 24.8 & 4.8 & 16.1 & 82.8 & 38.2 & 87.3 & 36.2 & 0.0 & 76.5 & 12.2 & 10.6 & 0.0 & 0.0 & 29.0 & 35.9\\
   &PDFNet12-2s& \bfseries 93.5 & 58.2 & 79.0 &  13.3 & 14.6 & 28.6 & 5.4 & 25.4 & 82.5 & 39.7 & 86.1 & 38.8 & 1.3 & 77.8 &  2.8 & 17.7 & 3.7 & 0.0 & 31.8 & 36.9 \\
   \midrule
   &ResNet18& 85.0 & 32.3 & 70.4& 1.6 & 0.0 & 0.0 & 0.0 & 0.0 & 75.9 & 32.0 & 75.2 & 0.0 & 0.0 & 53.0 & 0.0 & 0.0 & 0.0 & 0.0 & 0.0 & 22.4\\
  &MobileNetV2& 85.9 & 37.3 & 68.8& 0.0 & 0.0 & 0.0 & 0.0 & 0.0 & 74.1 & 34.6 & 71.3 & 0.0 & 0.0 & 54.6 & 0.0 & 0.0 & 0.0 & 0.0 & 0.0 & 22.5\\
   &EfficientNetb1& 85.2 & 38.8 & 71.0& 4.1 & 1.5 & 0.0 & 0.0 & 0.2 & 78.6 & 31.8 & 80.0 & 17.9 & 0.0 & 58.1 &0.0 & 0.0 & 0.0 & 0.0 & 0.0 & 24.6\\
   &RegNetY08& 89.4 & 47.2 & 74.3&  6.5 & 2.5 & 0.0 & 0.2 & 8.2 & 78.9 &  35.7 &  82.6 & 22.4 & 0.0 &  66.9 & 3.1 & 0.0 & 0.0 & 0.0 & 2.2 & 27.4\\
     &ResNet101& 85.4 & 29.7 & 66.9& 0.0 & 0.0 & 0.0 & 0.0 & 0.0 & 73.4 & 31.3 & 75.0 & 0.0 & 0.0 & 48.5 & 0.0 & 0.0 & 0.0 & 0.0 & 0.0 & 21.6\\
   &DenseNet161& 87.8 & 40.2 & 72.2& 3.2 & 2.4 & 0.0 & 0.0 & 0.0 & 76.9 & 28.3 & 78.6 & 15.2 & 0.0 & 61.5 & 2.3 & 0.0 & 0.0 & 0.0 & 5.1 & 24.9\\
   &HRNet-V2& 86.4 & 19.3 & 63.8 &  0.0 & 0.0 & 0.0 & 0.0 & 0.0 & 67.9 & 0.0 & 70.6 & 0.0 & 0.0 & 38.8 & 0.0 & 0.0 & 0.0 & 0.0 & 0.0 & 18.3\\
 $T_{185}$  &U-Net& 90.2 & 50.1 & 74.0 &  7.5 & 0.6 & 0.0 & 0.0 & 8.1 & 81.1 & 33.1 & 82.7 & 13.2 & 0.0 & 68.4 & 0.0 & 0.0 & 0.6 &  0.2 & 18.4 & 27.8\\
   &PDFNet3& \bfseries92.1 & 55.9 & 77.8 &  11.3 & 14.0 & 24.5 & 2.7 & 22.2 & 82.6 & \bfseries 37.0 & 85.1 & 34.0 & 0.7 & 71.9 & 1.3 & 15.3 &  4.7 & 0.1 & 26.0 & 34.7 \\
   &PDFNet6& 91.4 & \bfseries 56.8 & 77.7 &  8.7 & \bfseries 14.8 & \bfseries 28.2 & \bfseries 7.1 & 23.3 & 82.3 & 35.8 & 84.1 & \bfseries 34.6 & \bfseries 4.2 &  73.8 & 3.7 & \bfseries 18.0 & 3.7 & 0.1 & \bfseries 31.9 & \bfseries35.8 \\
   &PDFNet9& 91.9 & 55.2 &  77.9 &  7.9 & 7.6 & 26.8 & 3.1 & 22.1 & 81.3 & \bfseries 37.0 & \bfseries 86.0 & 32.4 & 0.0 &  73.8 & \bfseries 6.5 & 15.1 & 2.6 & 0.0 & 21.6 & 34.1\\
   &PDFNet12& 91.0 & 54.5 &  \bfseries 78.0 & \bfseries  15.5 & 9.7 & 25.8 & 6.3 & \bfseries 23.8 &\bfseries 82.8 & 36.1 & 84.3 & \bfseries 34.6 & 0.0 & 73.4 & 4.8 & 1.0 & 0.0 & 0.0 & 21.3 & 33.8 \\
   &PDFNet3-2S& 91.8 & 51.2 & 76.1 &  8.9 & 7.2 & 22.9 & 1.6 & 22.4 & 81.7 & 36.9 & 84.3 & 31.7 & 2.8 & 71.4 & 3.0 & 7.3 & 0.5 & \bfseries 1.2 & 30.8 & 33.4\\
   &PDFNet6-2S& 91.7 & 54.2 & 76.7 &  10.4 & 7.3 & 21.6 & 0.1 & 15.5 & 79.7 & 34.0 & 85.4 & 29.2 &  0.0 & 72.6 & 3.6 &  8.3 & \bfseries 5.7 & 0.0 &  22.5 &  32.6 \\
   &PDFNet9-2S& 91.1 & 54.0 &  76.2 &  12.2 & 8.6 & 24.2 & 2.0 & 18.8 & 80.9 & 35.7 & 83.1 & 33.2 & 0.1 & \bfseries 74.0 & 3.5 & 2.8 & 0.0 & \bfseries 1.2 & 24.3 & 32.9 \\
   &PDFNet12-2s&  91.0 & 51.4 & 75.6 &  9.5 &  3.1 & 20.4 & 0.0 & 15.3 & 80.1 & 36.5 & 82.5 & 27.0 & 0.0 & 72.5 & 1.4 & 0.8 & 1.4 & 0.0 & 28.8 & 31.4 \\
   
   \midrule
   &ResNet18& 85.7 & 26.4 & 66.4& 0.0 & 0.0 & 0.0 & 0.0 & 0.0 & 74.2 & 26.9 & 72.9 & 0.0 & 0.0 & 47.4 & 0.0 & 0.0 & 0.0 & 0.0 & 0.0 & 21.0\\
   &MobileNetV2& 82.1 & 20.4 & 63.7& 0.0 & 0.0 & 0.0 & 0.0 & 0.0 & 69.2 & 25.5 & 67.7 & 0.0 & 0.0 & 36.6 & 0.0 & 0.0 & 0.0 & 0.0 & 0.0 & 19.2\\
   &EfficientNetb1& 84.7 & 26.6 & 59.8& 1.6 & 0.0 & 0.0 & 0.0 & 0.1 & 67.2 & 29.9 & 58.6 &2.4 & 0.0 & 46.1 & 0.0 & 0.0 & 0.0 & 0.0 & 0.0 & 19.8\\
   &RegNetY08& 86.7 & 33.4 & 68.5 & 0.0 & 0.0 & 0.0 & 0.0 & 0.0 & 74.8 & 27.9 & 76.3 & 0.0 & 0.0 & 52.8 & 0.0 & 0.0 & 0.0 & 0.0 & 0.0 & 22.1\\
  &ResNet101& 81.4 & 20.7 & 62.2 & 0.0 & 0.0 & 0.0 & 0.0 & 0.0 & 69.4 & 28.2 & 68.6 & 0.0 & 0.0 & 39.9 & 0.0 & 0.0 & 0.0 & 0.0 & 0.0 & 19.4\\
  &DenseNet161& 84.0 & 24.3 & 67.5 & 0.0 & 0.3 & 0.0 & 0.0 & 0.0 & 72.7 & 23.0 & 74.6 & 0.0 & 0.0 & 49.3 & 0.0 & 0.0 & 0.0 & 0.0 & 0.0 & 20.8\\
   &HRNet-V2& 81.0 & 3.2 & 49.5 &  0.0 & 0.0 & 0.0 & 0.0 & 0.0 & 52.0 & 0.0 & 67.2 & 0.0 & 0.0 & 38.9 &  0.0 & 0.0 & 0.0 & 0.0 & 0.0& 15.4\\
  $T_{92}$  &U-Net& 87.6 & 45.8 & 71.3 &  2.2  & 0.0 & 0.0 & 0.0 & 5.8 & 79.2 & 31.1 & 79.8 & 14.2 & 0.0 & 58.4 & 0.0 & 0.0 & 0.3 & 0.0 & 0.0 & 25.0 \\
   &PDFNet3& 90.2 & 49.5 & 74.4 &  5.6 & 10.9 & 15.3 & 0.0 & 8.2 & 79.0 & 34.6 & 82.1 & 22.6 & 0.0 & 68.6 & 1.1 & 0.9 & \bfseries 4.2 & 0.0 & 17.5 & 29.7\\
   &PDFNet6& 90.7 & \bfseries 50.9 & 75.3 &  6.3 & 7.2 & \bfseries 22.3 & 0.3 & \bfseries 14.6 & \bfseries 79.9 & 32.9 & 82.9 & 20.8 & \bfseries 0.7 & 67.8 & 0.4 &  1.9 & 2.6 & 0.0 &  21.7 & \bfseries 30.5 \\
   &PDFNet9& 90.1 & 50.8 &  74.7 &  6.7 & 7.7 & 14.9 & 0.0 & 12.2 & 79.4 & \bfseries 36.0 & 81.4 & 17.5 & 0.0 & 66.7 & 0.3 & 0.5 & 2.5 & 0.0 & 16.1 & 29.3\\
   &PDFNet12& \bfseries 90.8 & 48.7 &  \bfseries 75.4 & \bfseries 9.8 & \bfseries 13.6 & 19.5 & \bfseries 0.7 & 8.4  & 79.8 & 32.9 & \bfseries 84.1 & \bfseries 30.1 & 0.0 & \bfseries 69.0 & \bfseries 1.7 & 0.0 & 0.0 & 0.4 & 10.6 & 30.3\\
   &PDFNet3-2S& 90.5 & 45.8 & 72.1 &  6.9 & 3.3 & 16.0 & 0.4 & 9.6 & 78.6 & 35.7 & 81.4 & 29.5 & 0.5 & 67.8 & 0.9 & \bfseries 4.8 & 0.2 & \bfseries 0.7 & \bfseries 24.4 & 29.9\\
   &PDFNet6-2S& 89.4 & 45.9 & 71.3 &  3.6 & 1.7 & 0.0 & 0.0 & 0.0 & 76.8 & 28.8 & 78.1 & 4.0 & 0.0 & 59.4 & 0.1 &  0.0 & 1.6 & 0.0 & 1.0 & 24.3 \\
   &PDFNet9-2S& 88.9 & 46.5 &  74.6 &  5.8 & 8.0 & 9.3 & 0.2 & 1.2 & 79.3 &  31.1 & 81.2 & 26.4 & 0.0 & 68.3 & 0.7 & 0.0 & 0.0 & 0.0 & 2.4 & 27.6\\
   &PDFNet12-2s& 90.4 & 44.8 & 72.0 &  7.7 & 1.4 & 17.3 & 0.0 & 1.9 & 77.4 & 33.5 & 80.1 & 21.0 & 0.0 & 65.5 & 0.4 & 3.9 & 0.6 & 0.0 & 14.7 & 28.0 \\
   \bottomrule
\end{tabular}
\end{adjustbox}
\end{center}
 \caption{Class-wise results of the Cityscapes data ablation experiments evaluated on val set}
 \label{a-table-2}
\end{table*}
\subsubsection{Training plots}
\begin{figure*}[ht]
\centering     
\subfigure{\includegraphics[width=65mm]{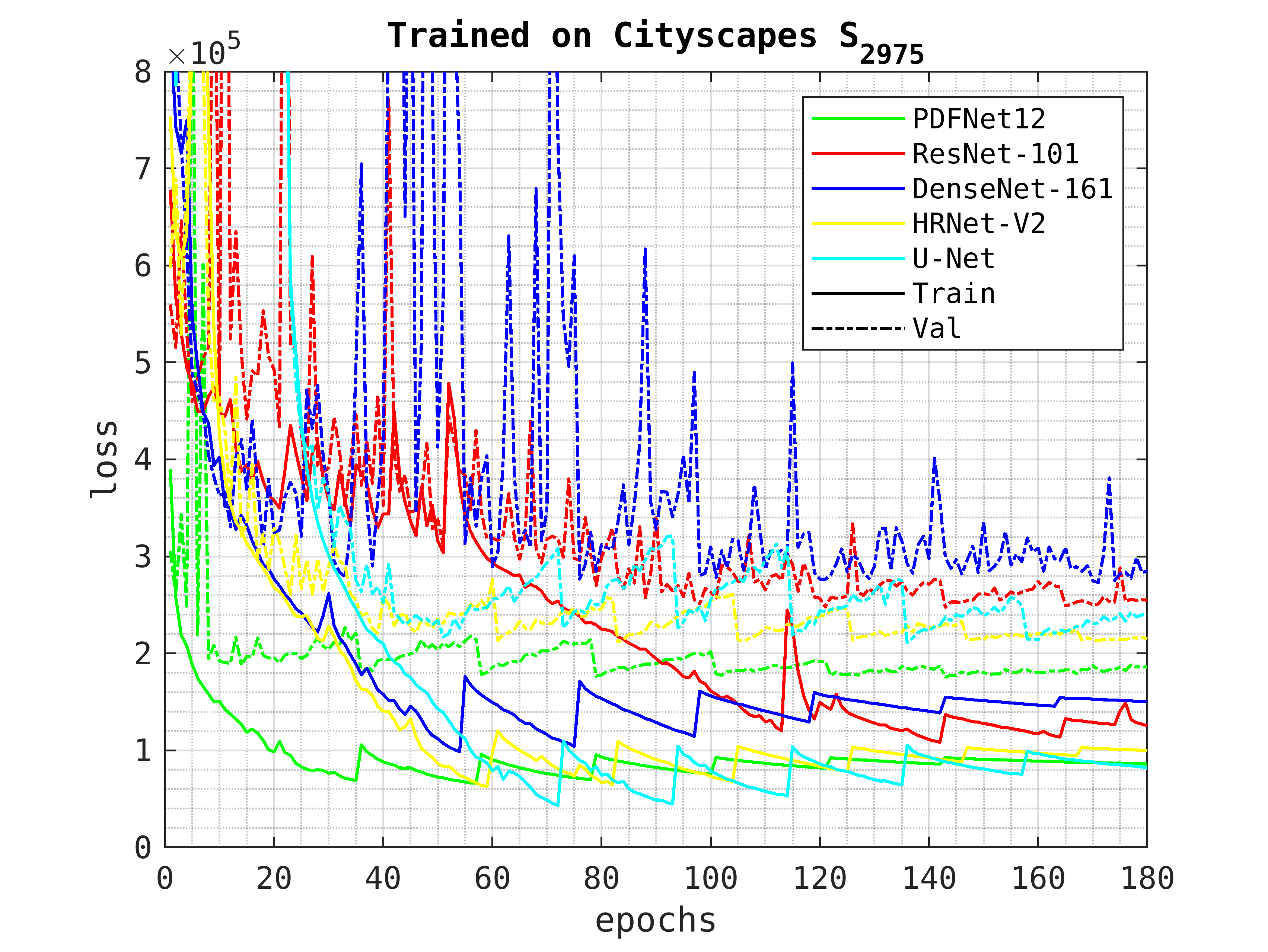}}
\subfigure{\includegraphics[width=65mm]{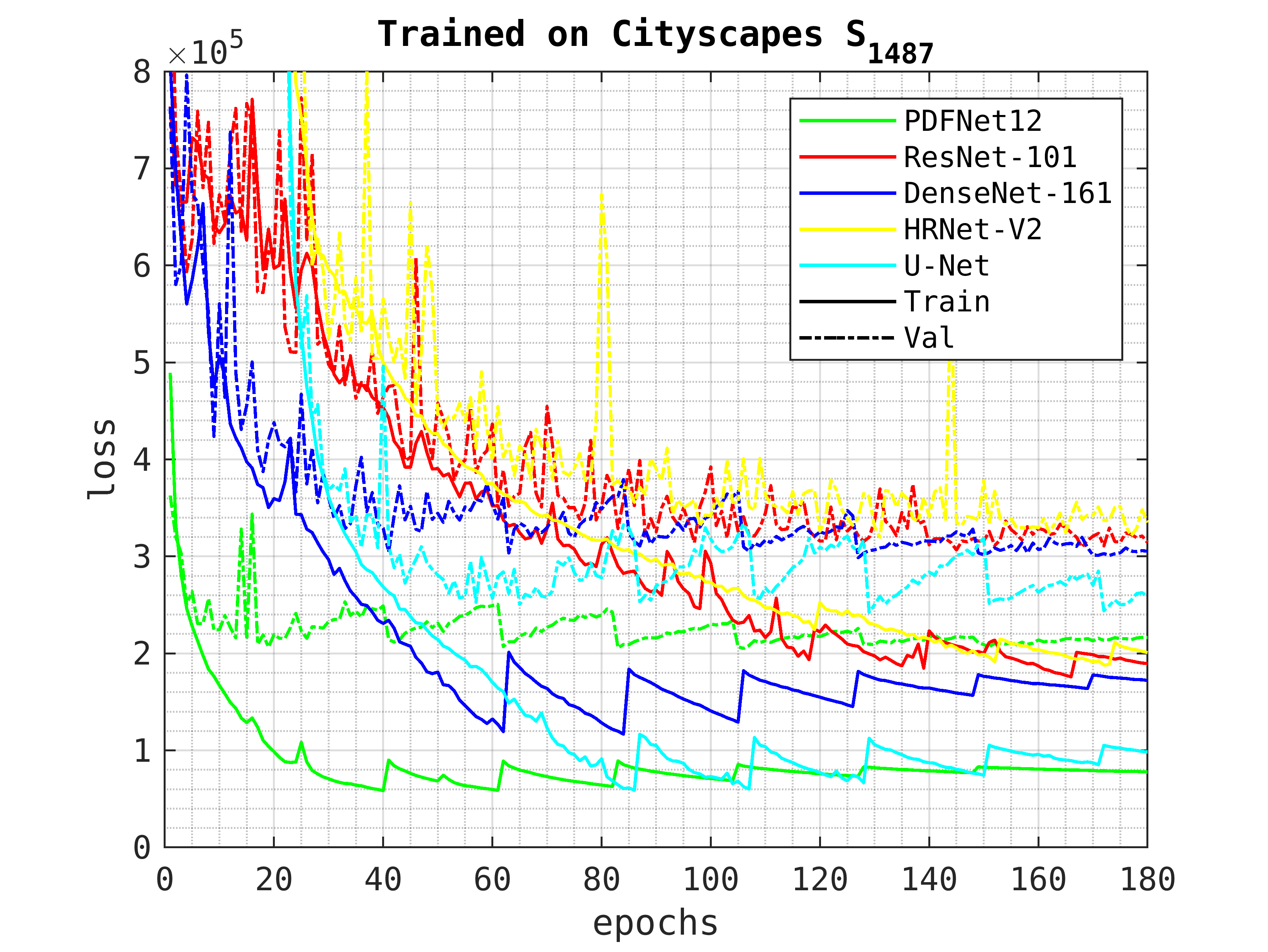}}
\subfigure{\includegraphics[width=65mm]{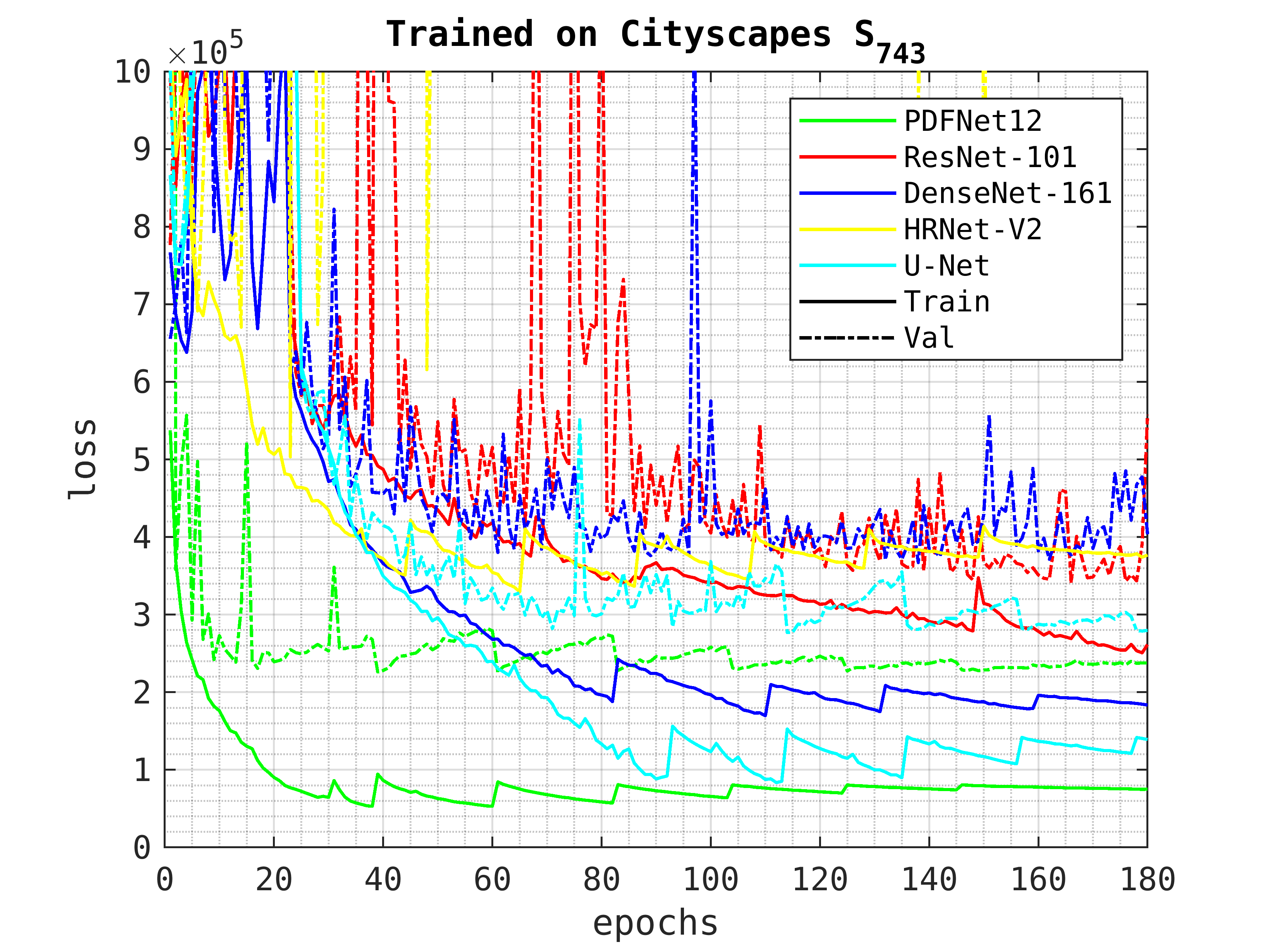}}
\subfigure{\includegraphics[width=65mm]{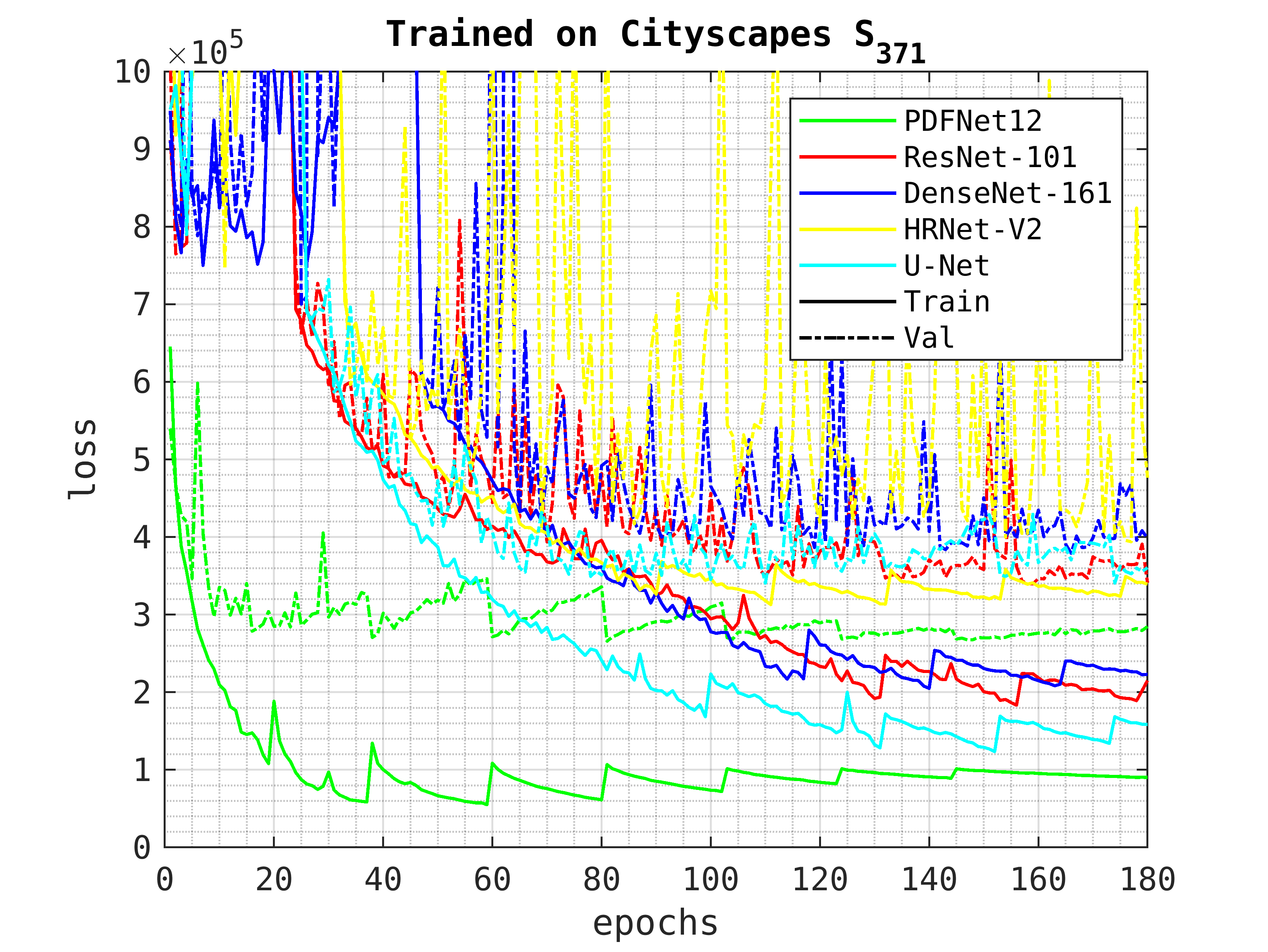}}
\subfigure{\includegraphics[width=65mm]{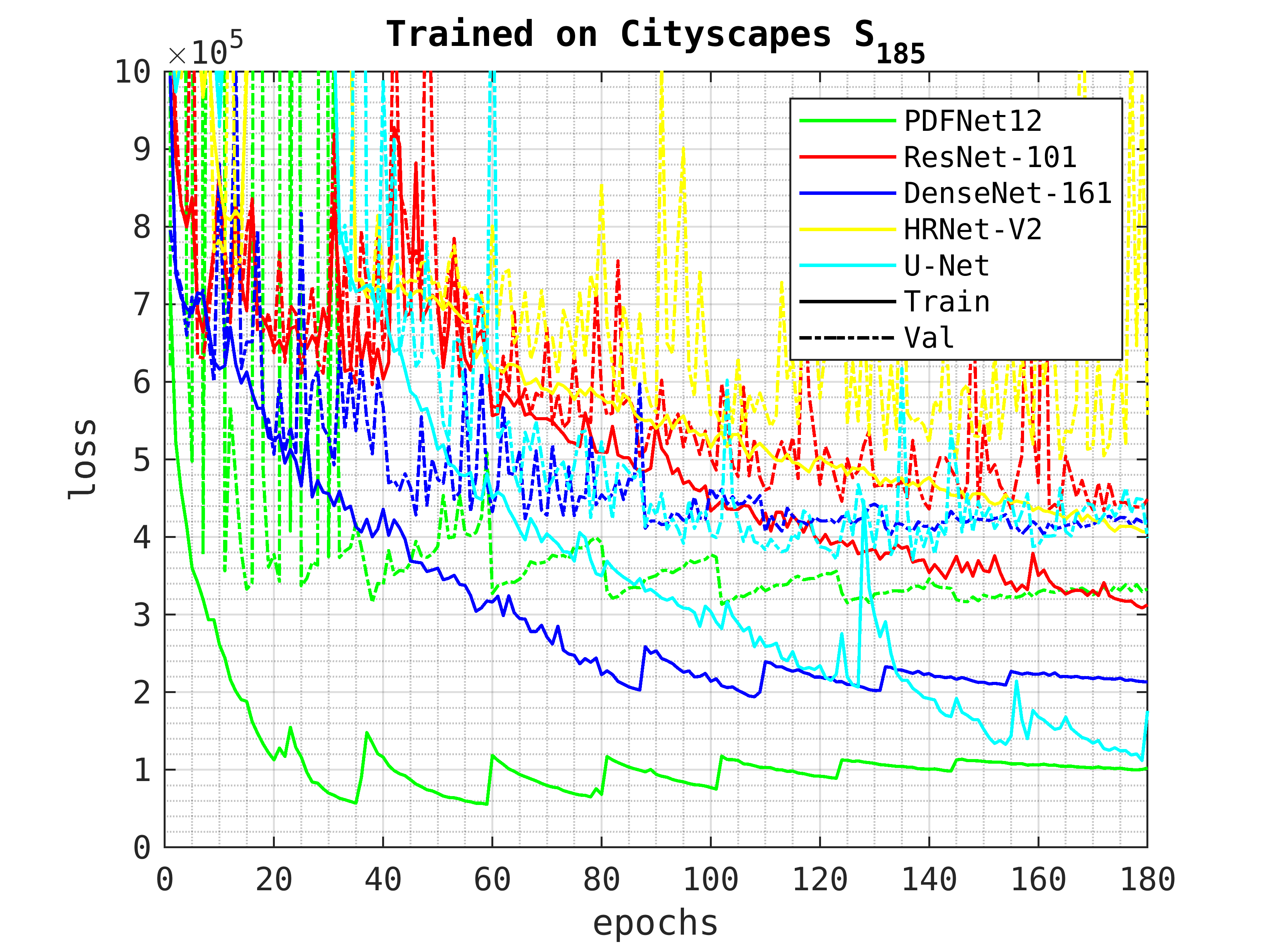}}
\subfigure{\includegraphics[width=65mm]{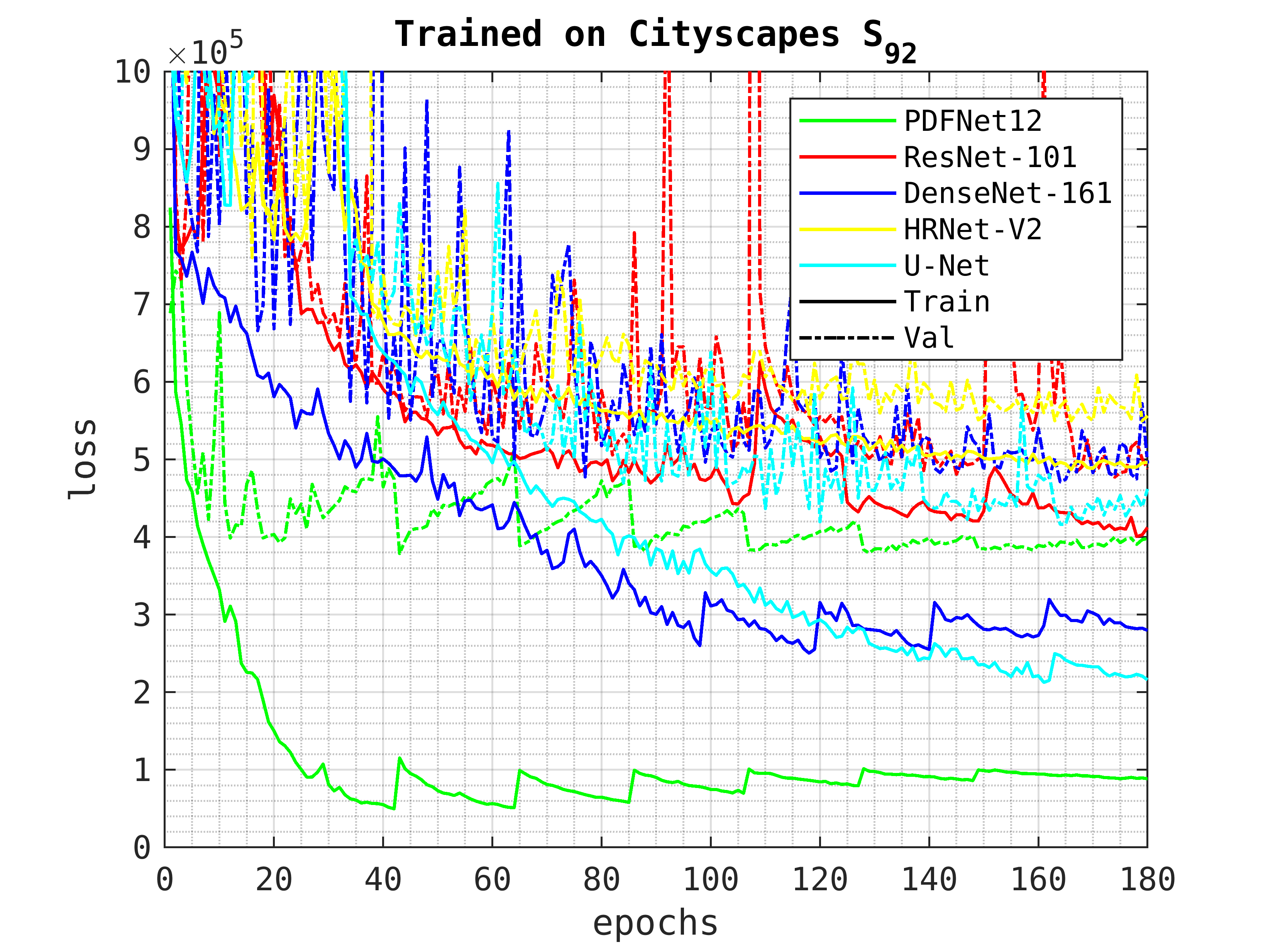}}
\caption{Cityscapes Set-1 training plots}
\label{C1}
\end{figure*}

\begin{figure*}[ht]
\centering     
\subfigure{\includegraphics[width=65mm]{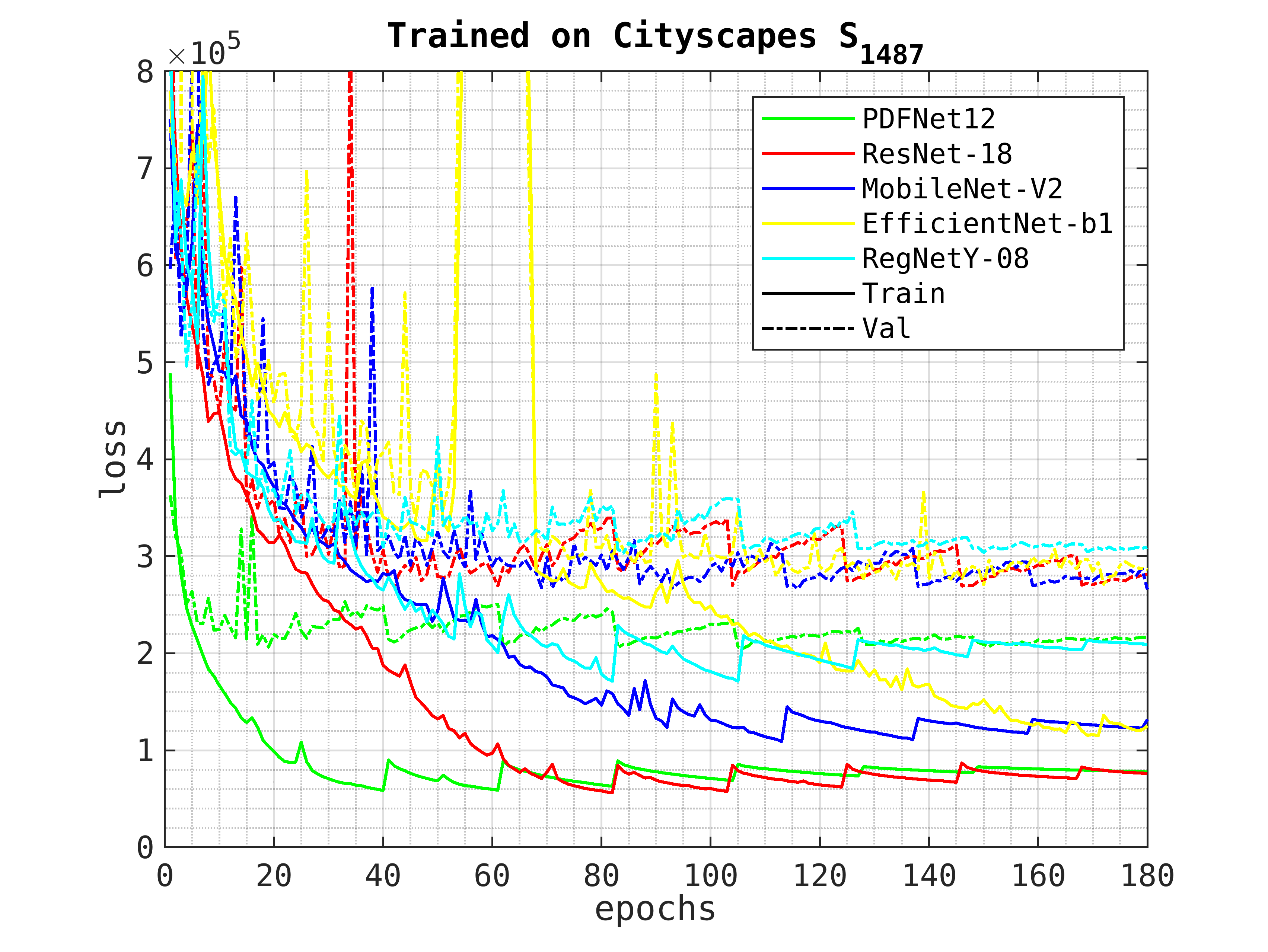}}
\subfigure{\includegraphics[width=65mm]{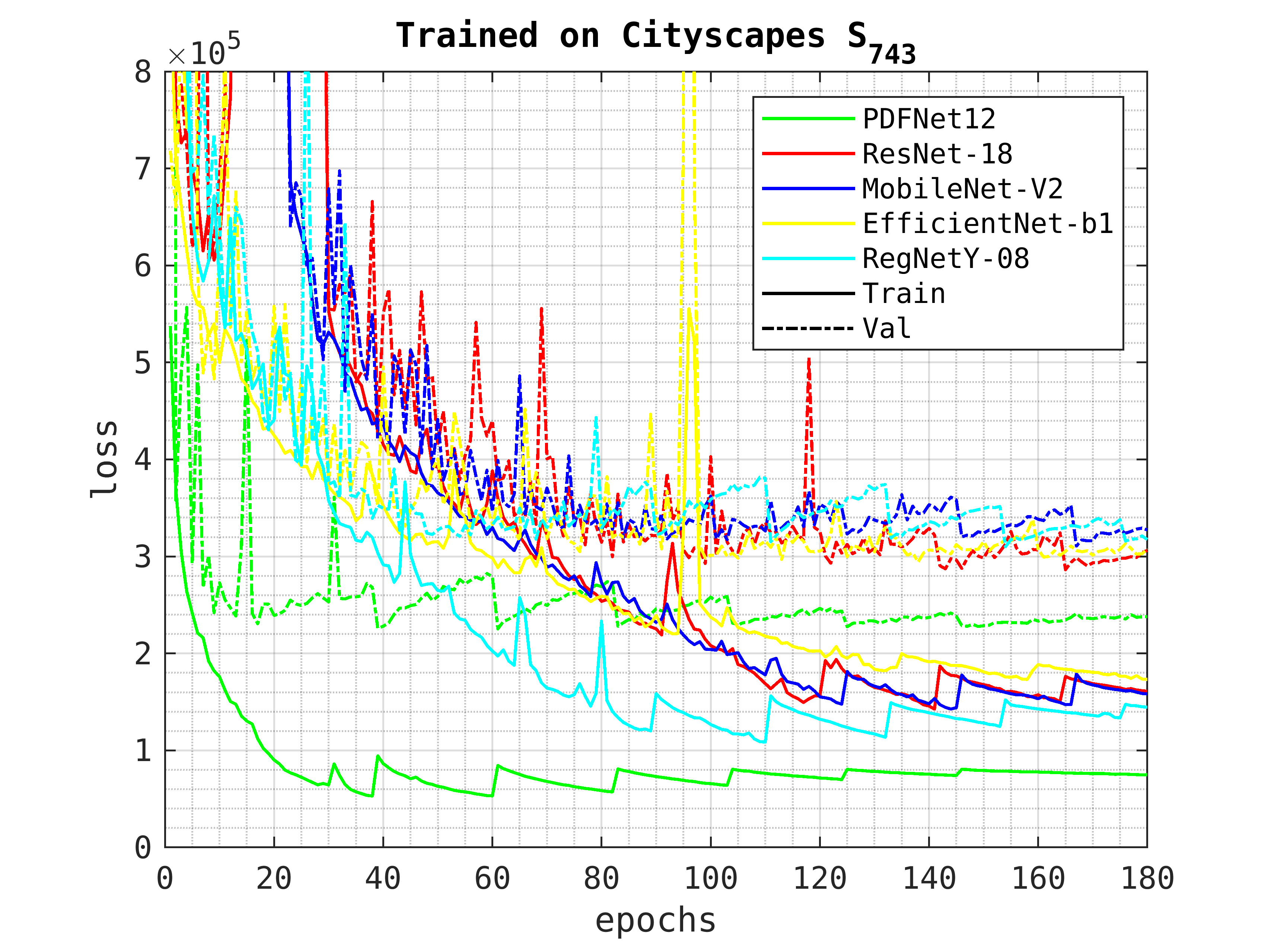}}
\subfigure{\includegraphics[width=65mm]{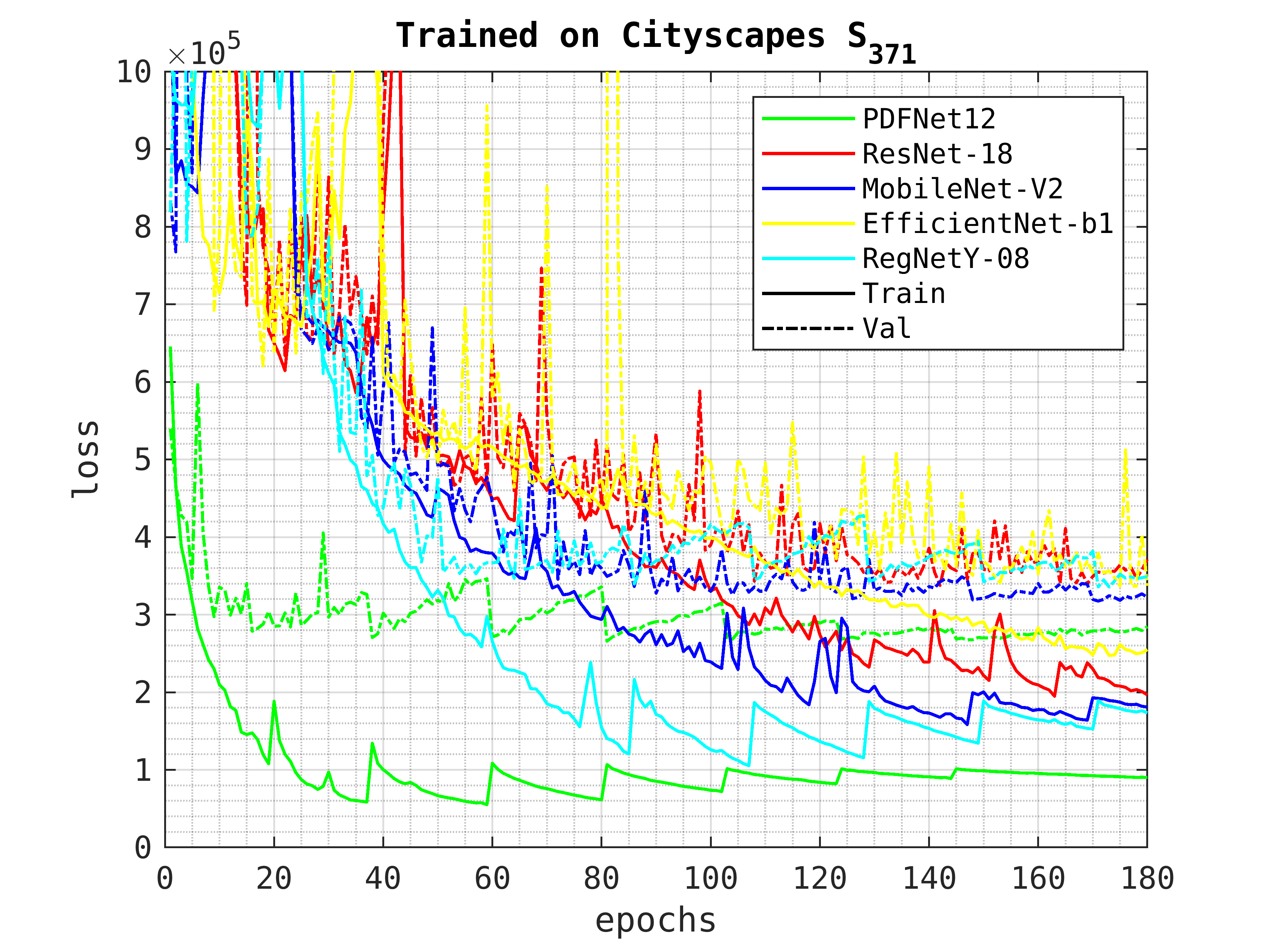}}
\subfigure{\includegraphics[width=65mm]{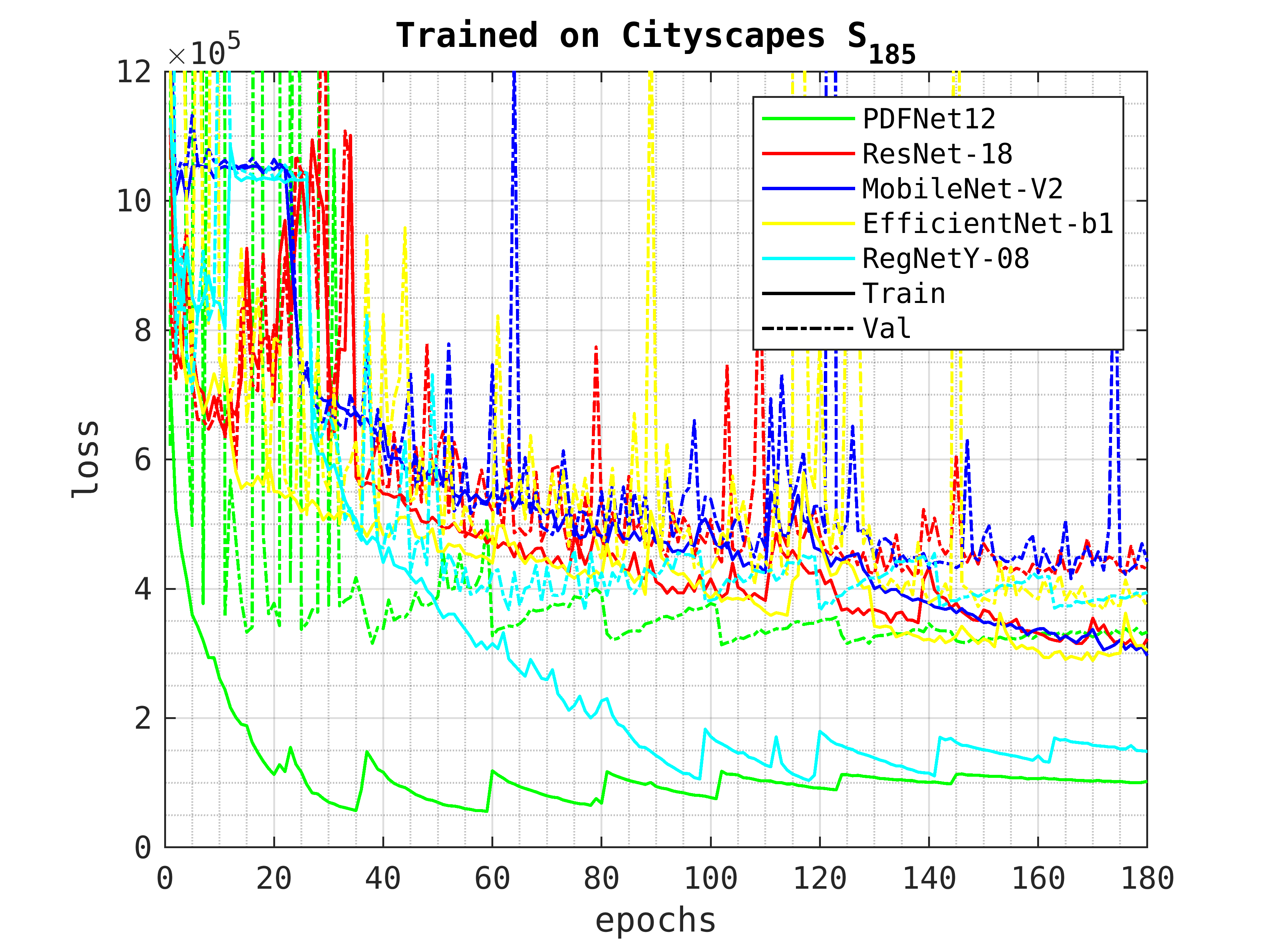}}
\subfigure{\includegraphics[width=65mm]{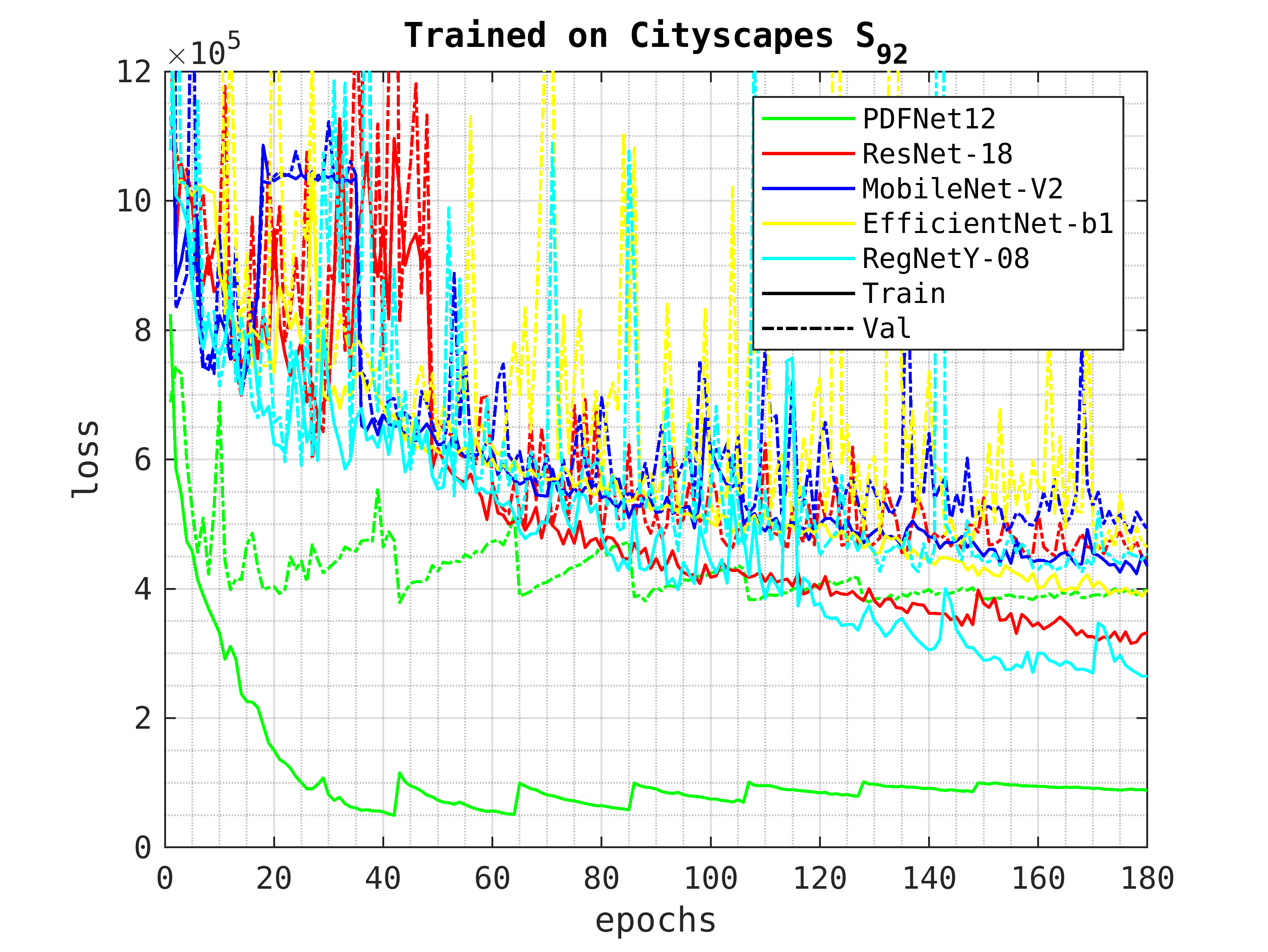}}
\caption{Cityscapes Set-2 training plots}
\label{C2}
\end{figure*}

\begin{figure*}[ht]
\centering     
\subfigure{\includegraphics[width=65mm]{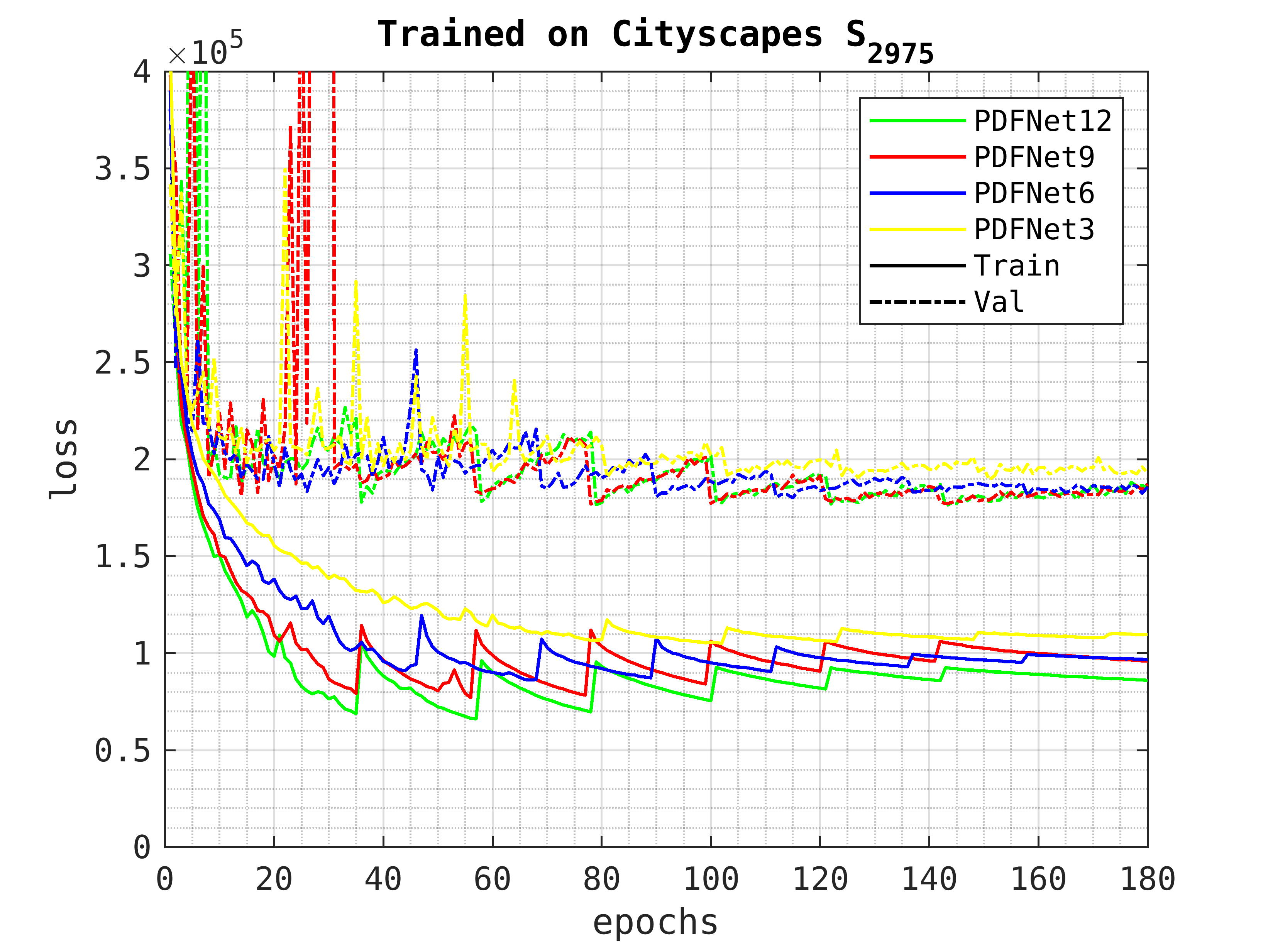}}
\subfigure{\includegraphics[width=65mm]{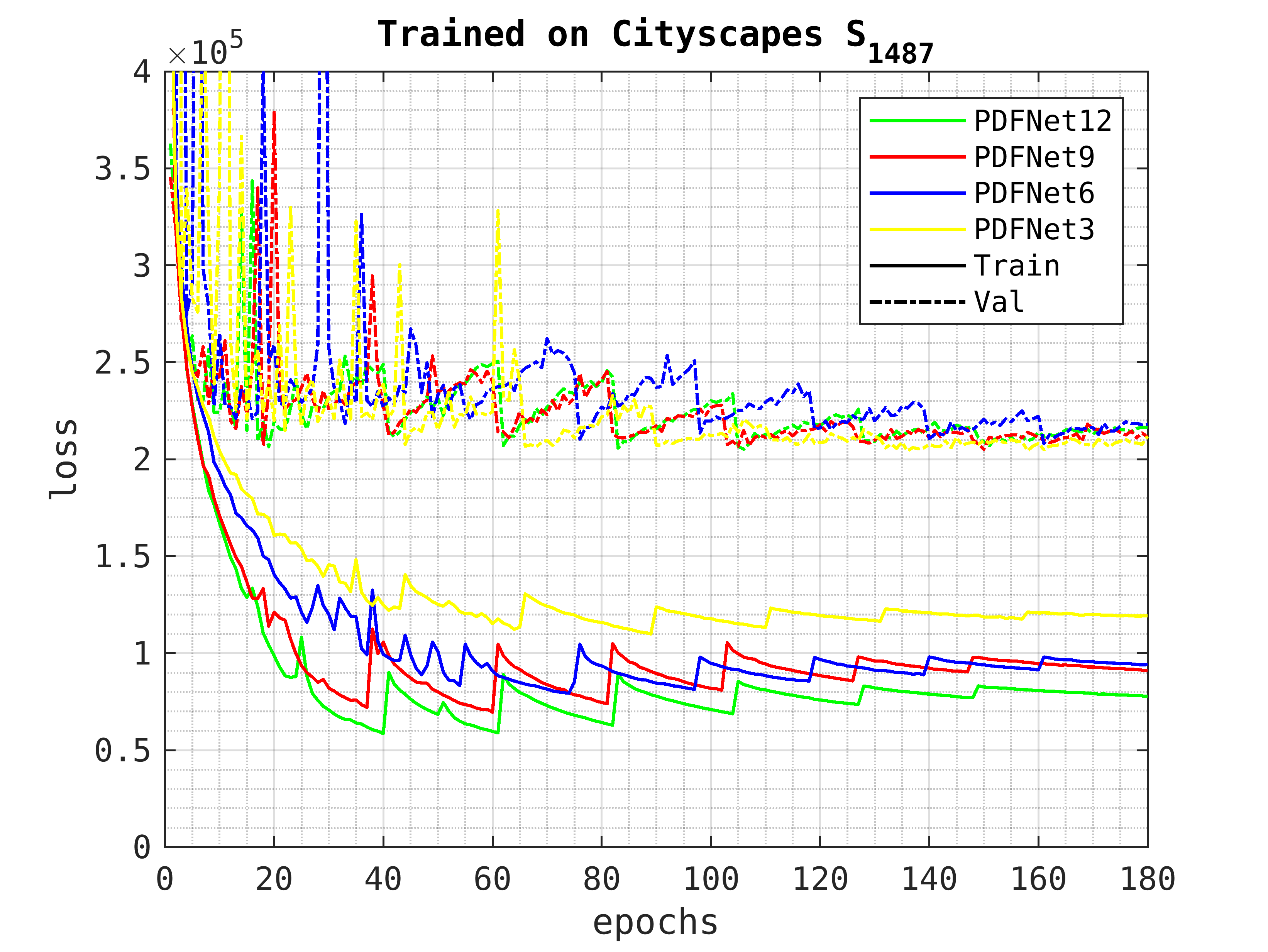}}
\subfigure{\includegraphics[width=65mm]{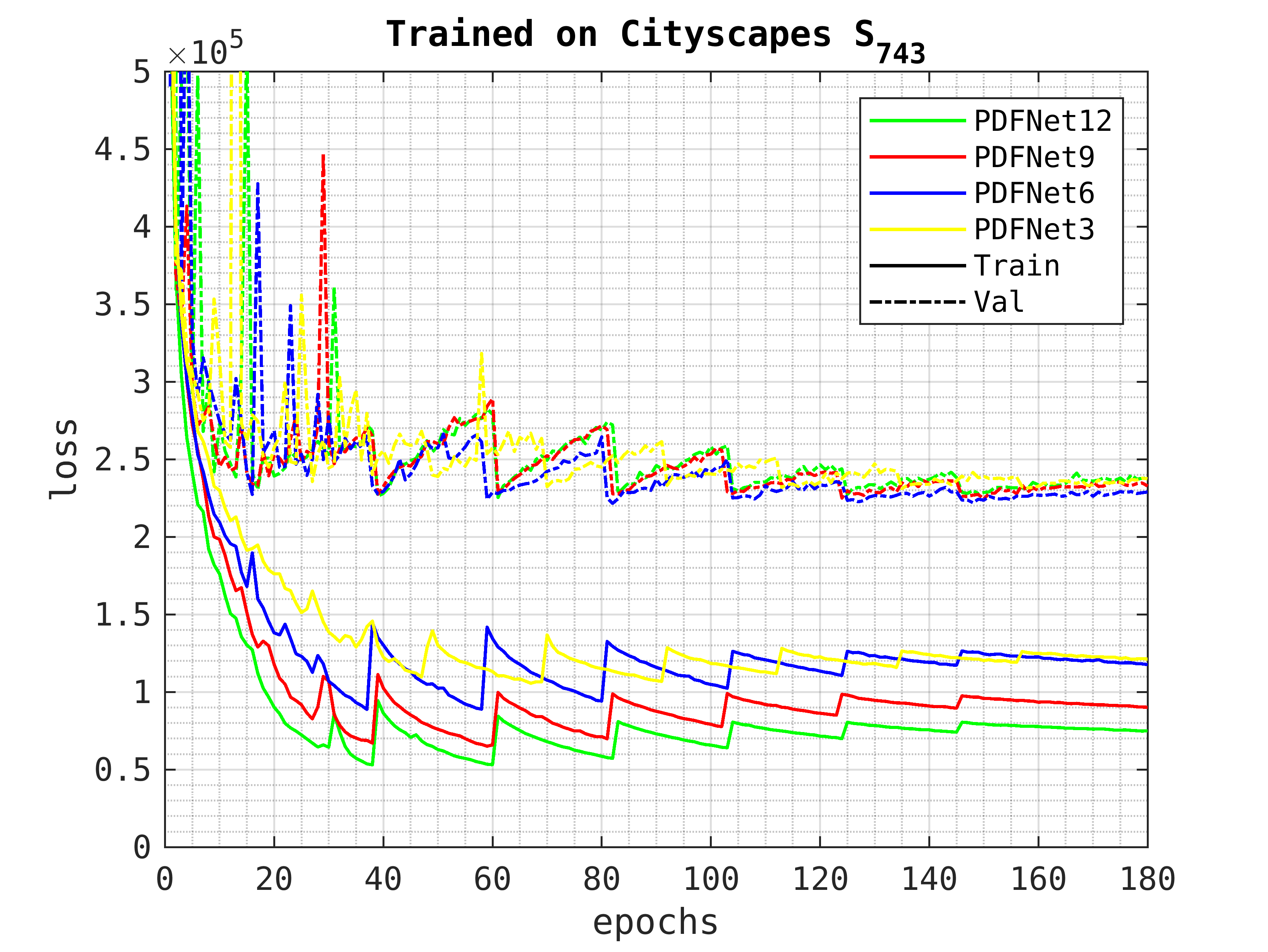}}
\subfigure{\includegraphics[width=65mm]{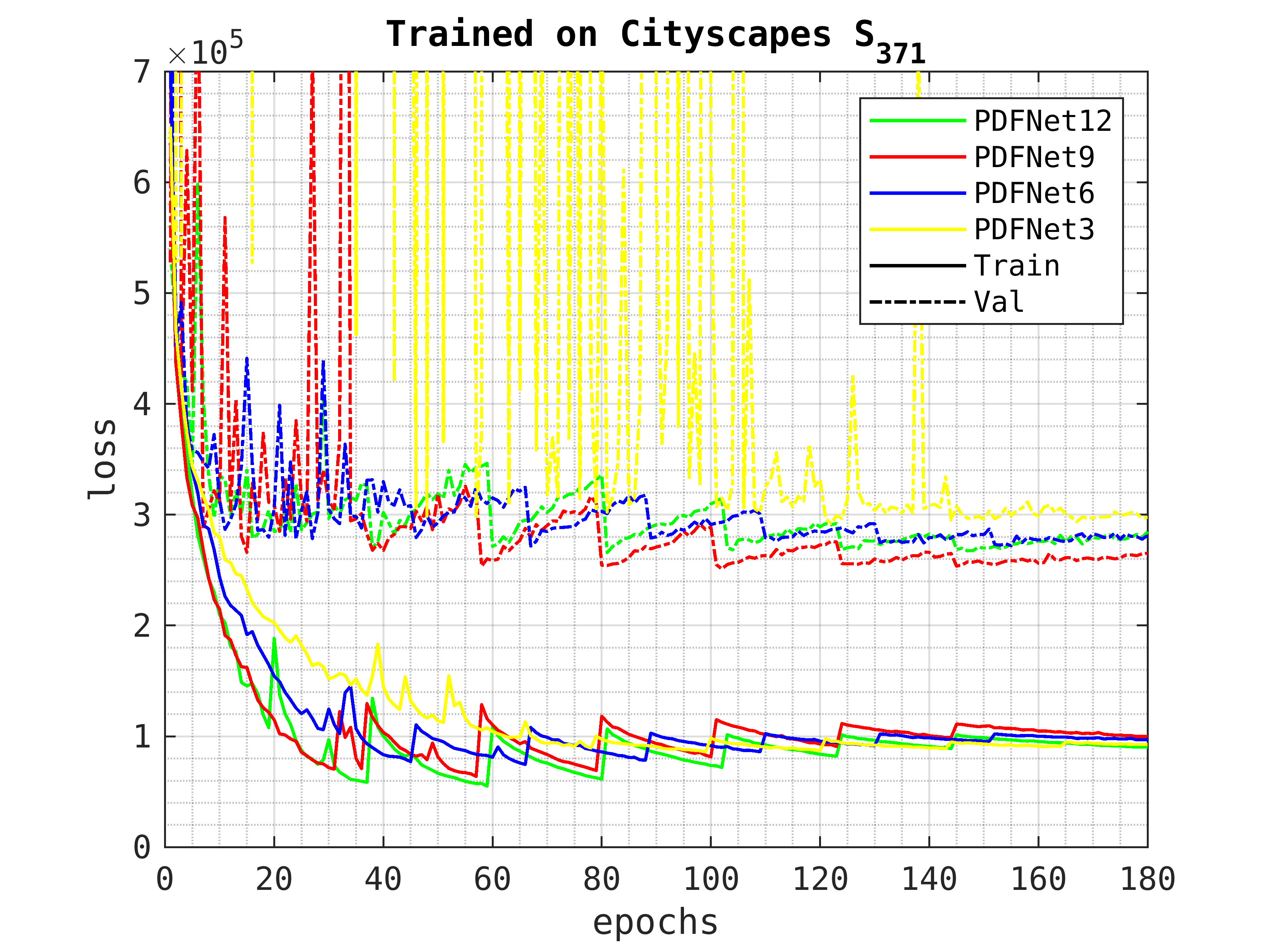}}
\subfigure{\includegraphics[width=65mm]{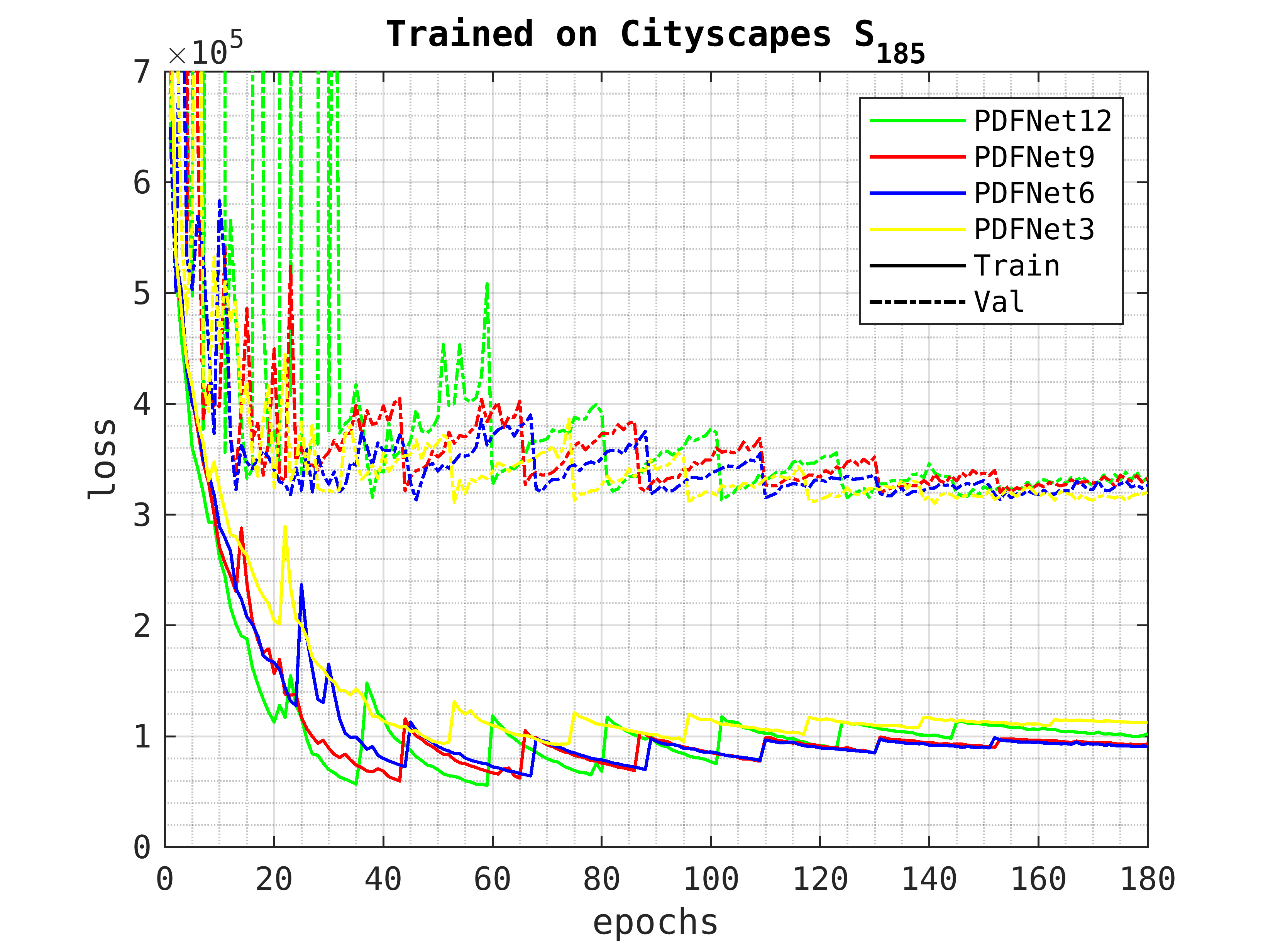}}
\subfigure{\includegraphics[width=65mm]{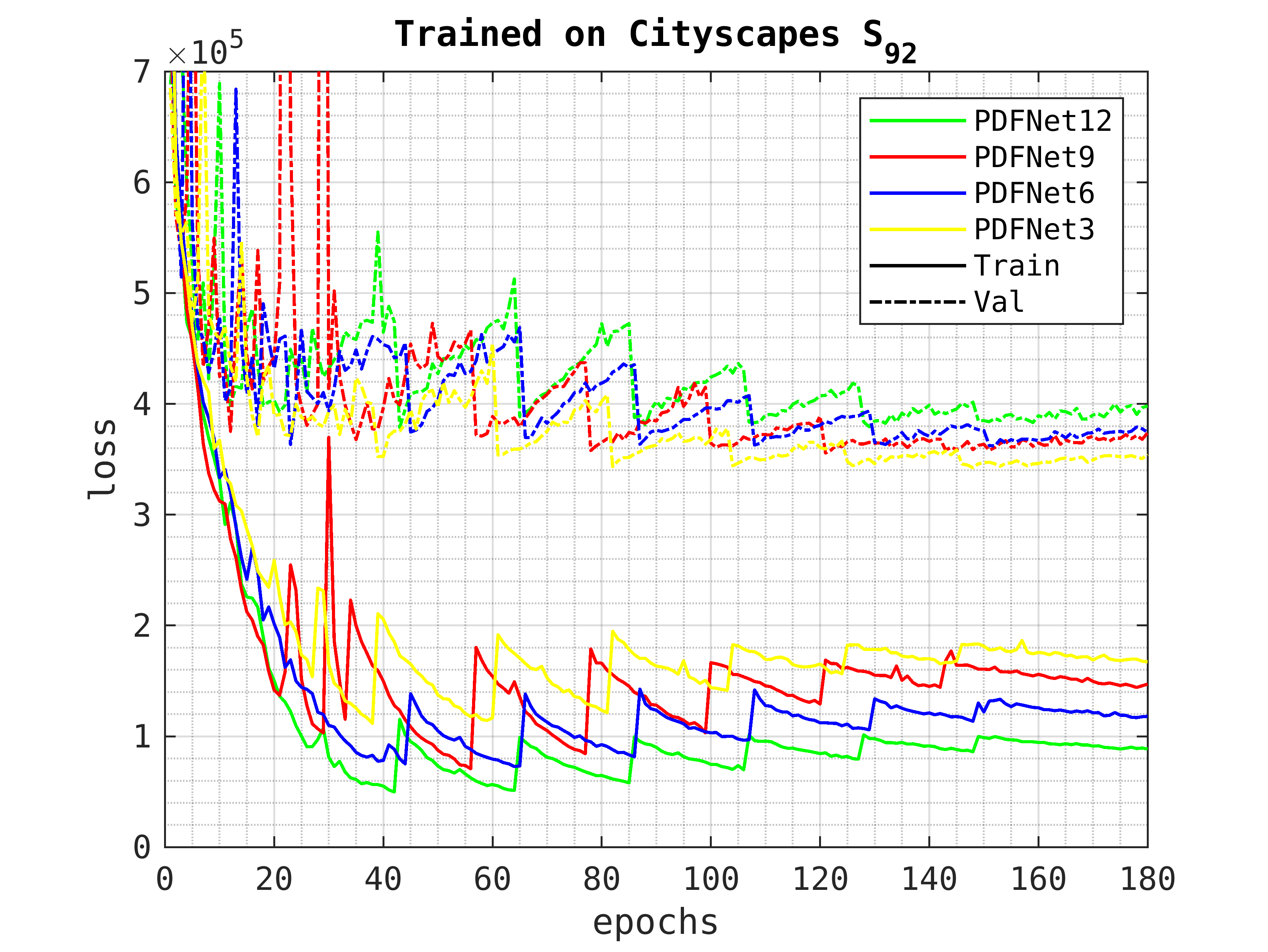}}
\caption{Cityscapes Set-3 training plots}
\label{C3}
\end{figure*}

For ease of visualization, we divide the Cityscapes experiments training plots into three sets.\newline
Set-1 consists of ResNet-101, DenseNet-161, HRNet-V2, U-Net, and PDFNet12 (shown in Figure \ref{C1}).\newline
Set-2 consists of ResNet-18, MobileNet-V2,  EfficientNet-b1, RegNetY-08, and PDFNet12 (shown in Figure \ref{C2}).\newline
Set-3 consits of PDFNet3, PDFNet6, PDFNet9, and PDFNet12 (shown in Figure \ref{C3}).

\subsection{Qualitative results}
\begin{sidewaysfigure*}[ht!]
\centering     
\subfigure{\includegraphics[width=43mm]{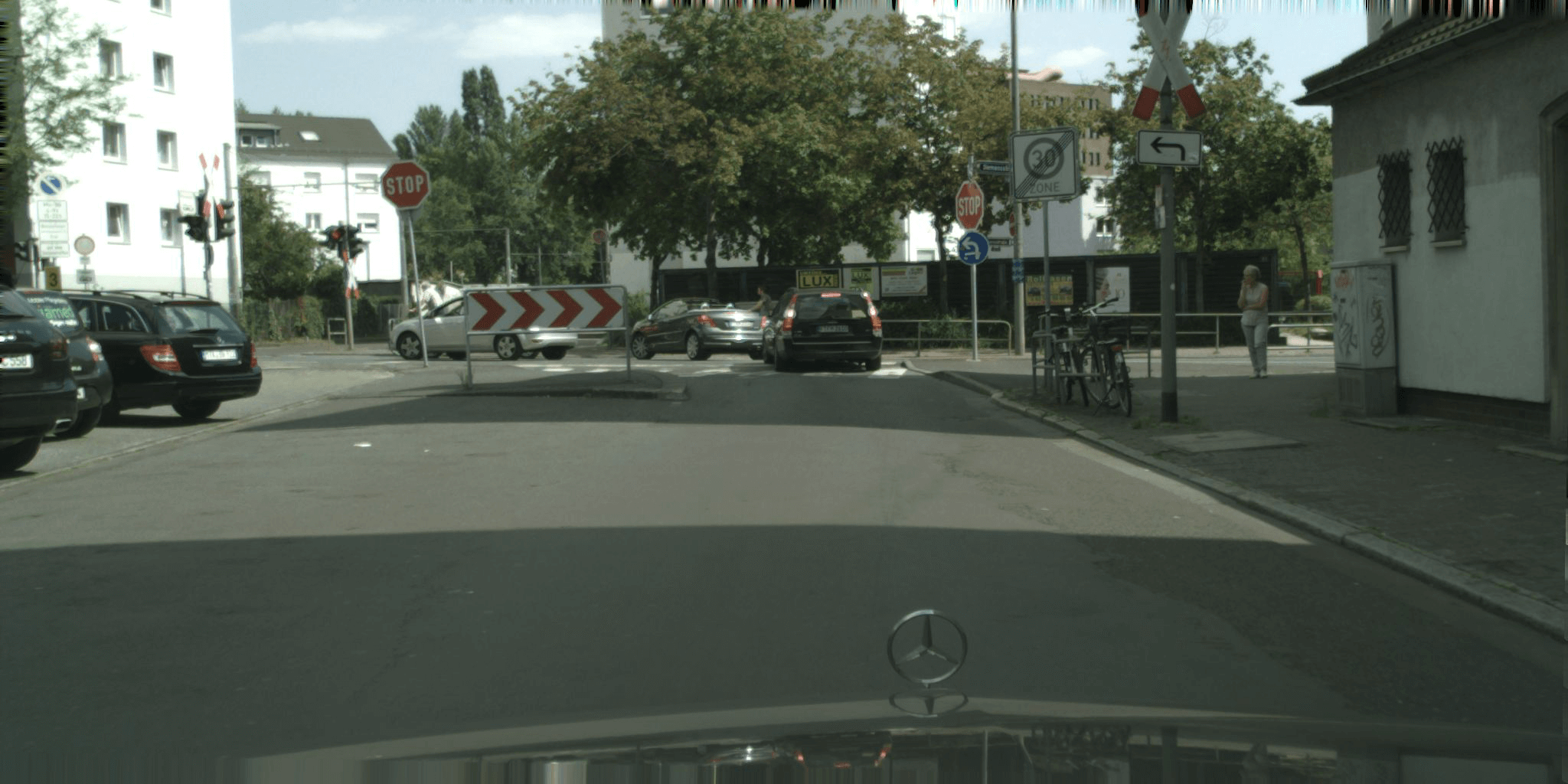}}
\subfigure{\includegraphics[width=43mm]{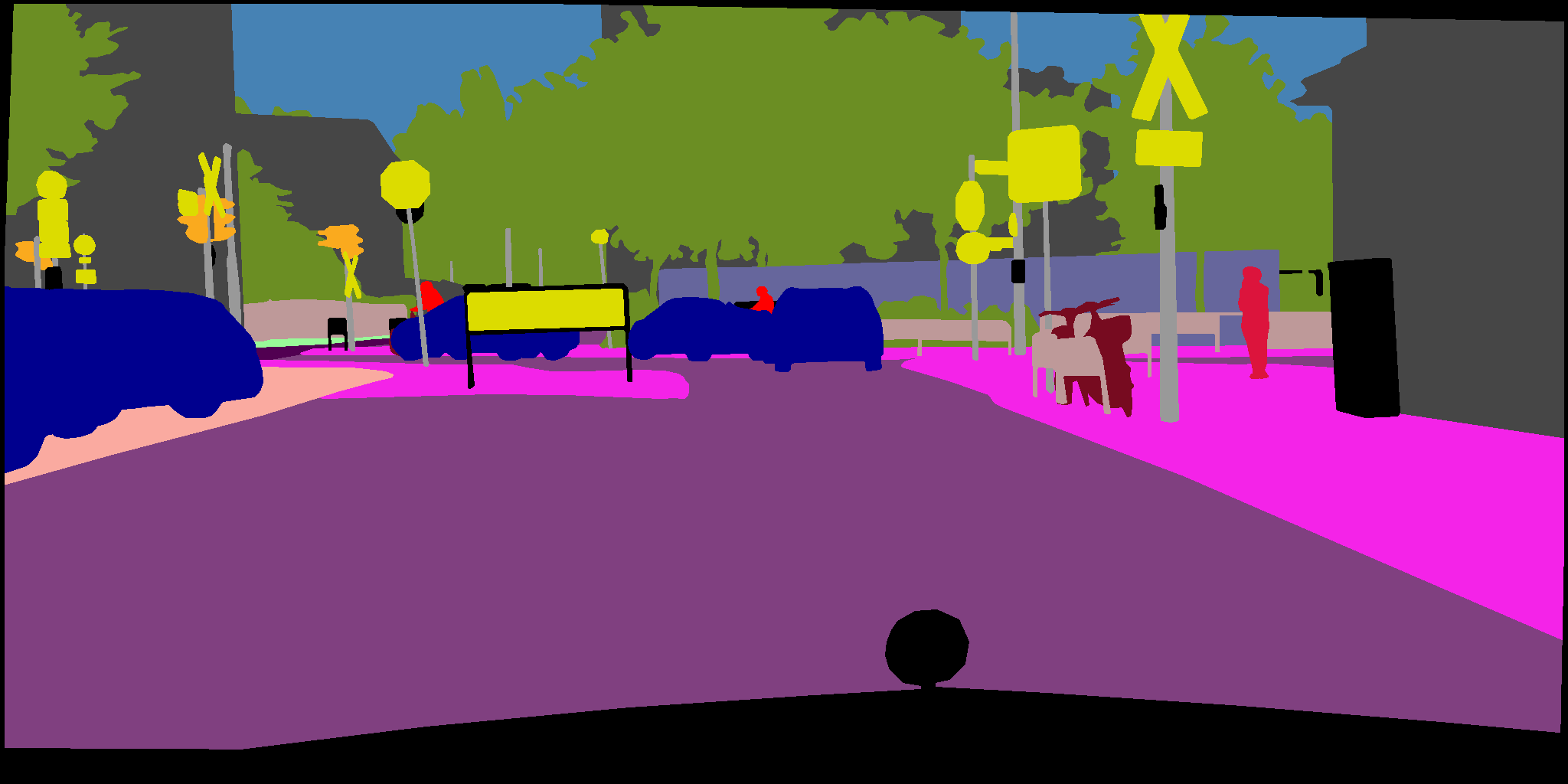}}
\subfigure{\includegraphics[width=43mm]{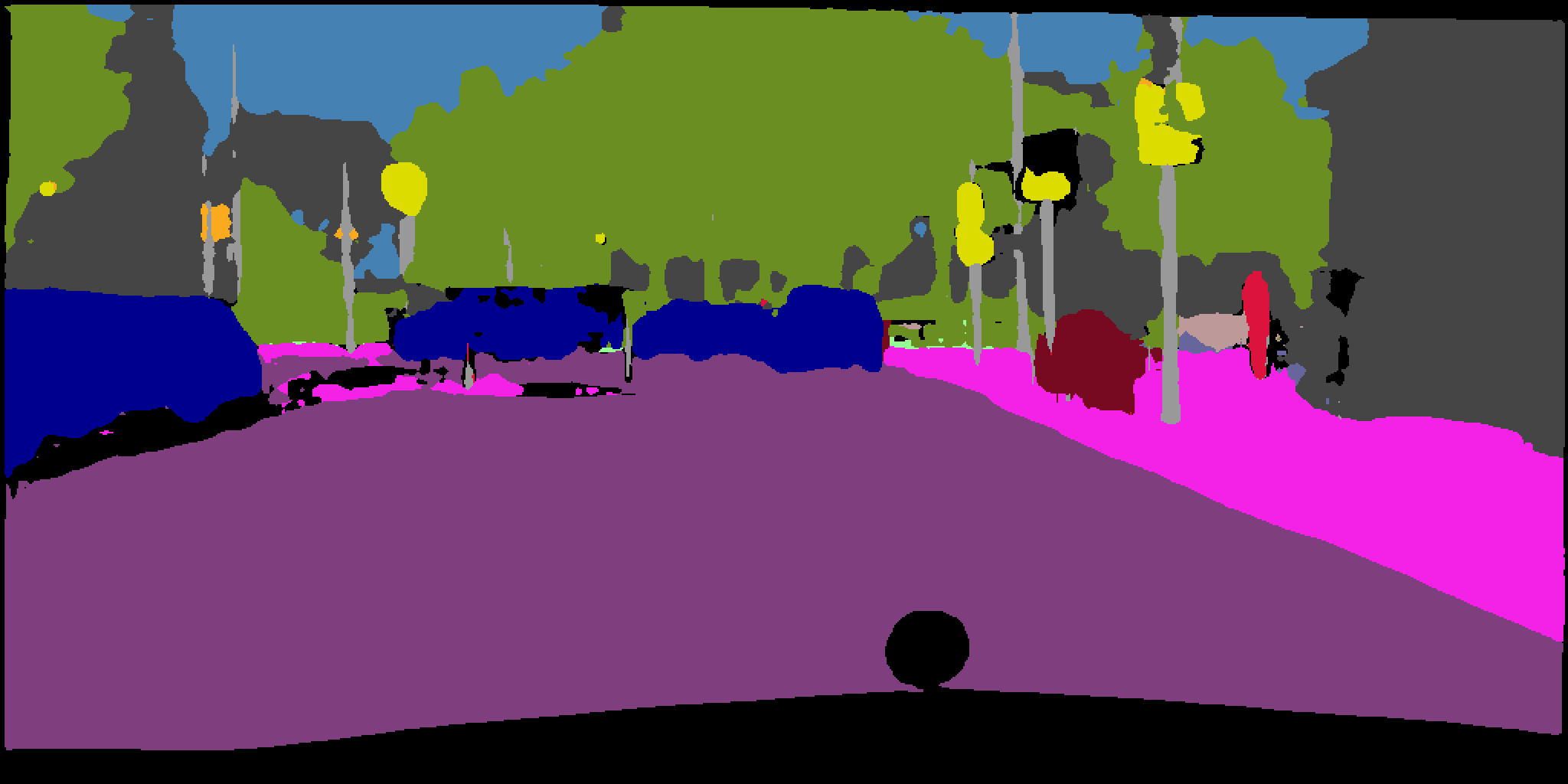}}
\subfigure{\includegraphics[width=43mm]{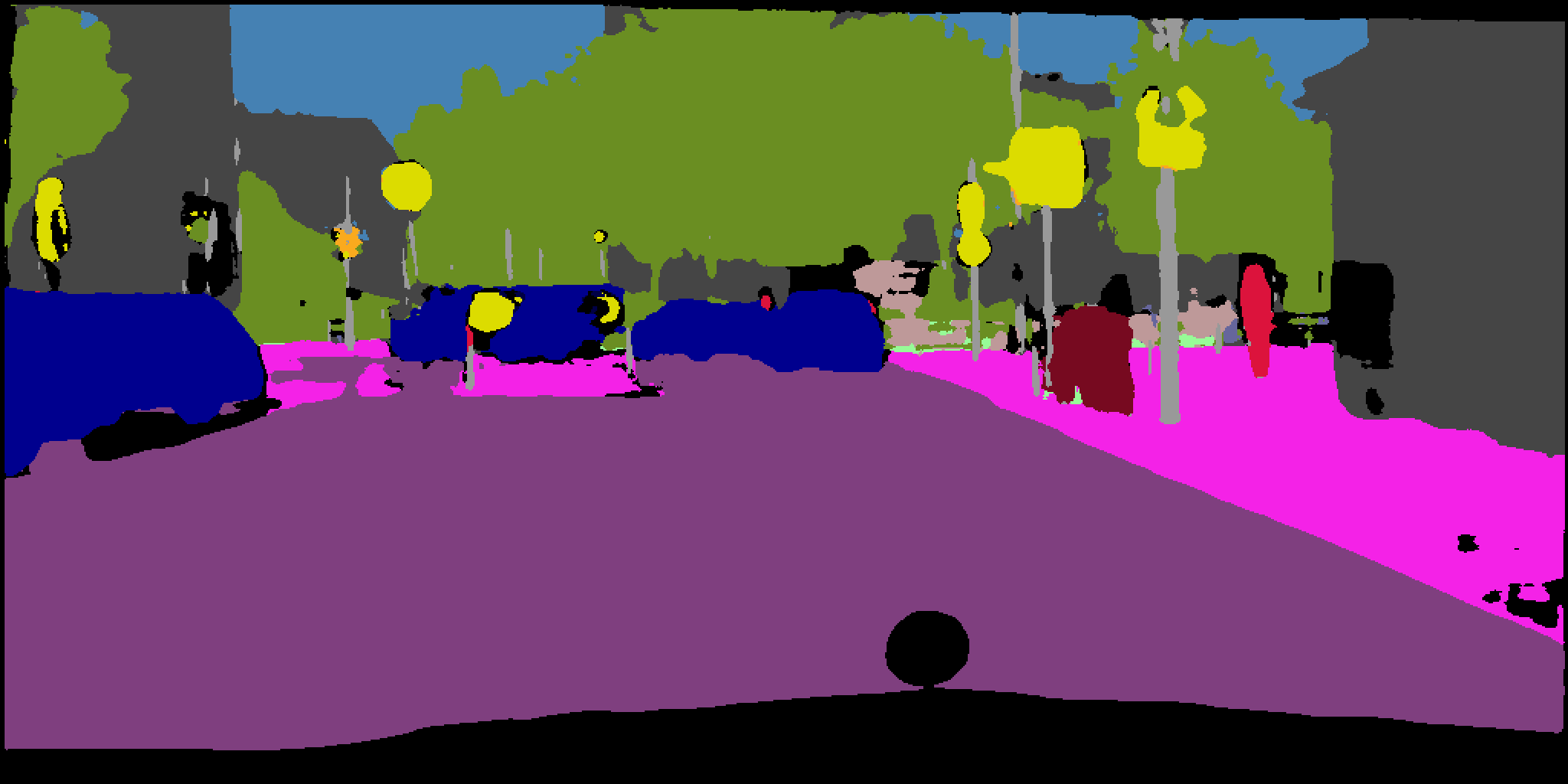}}
\subfigure{\includegraphics[width=43mm]{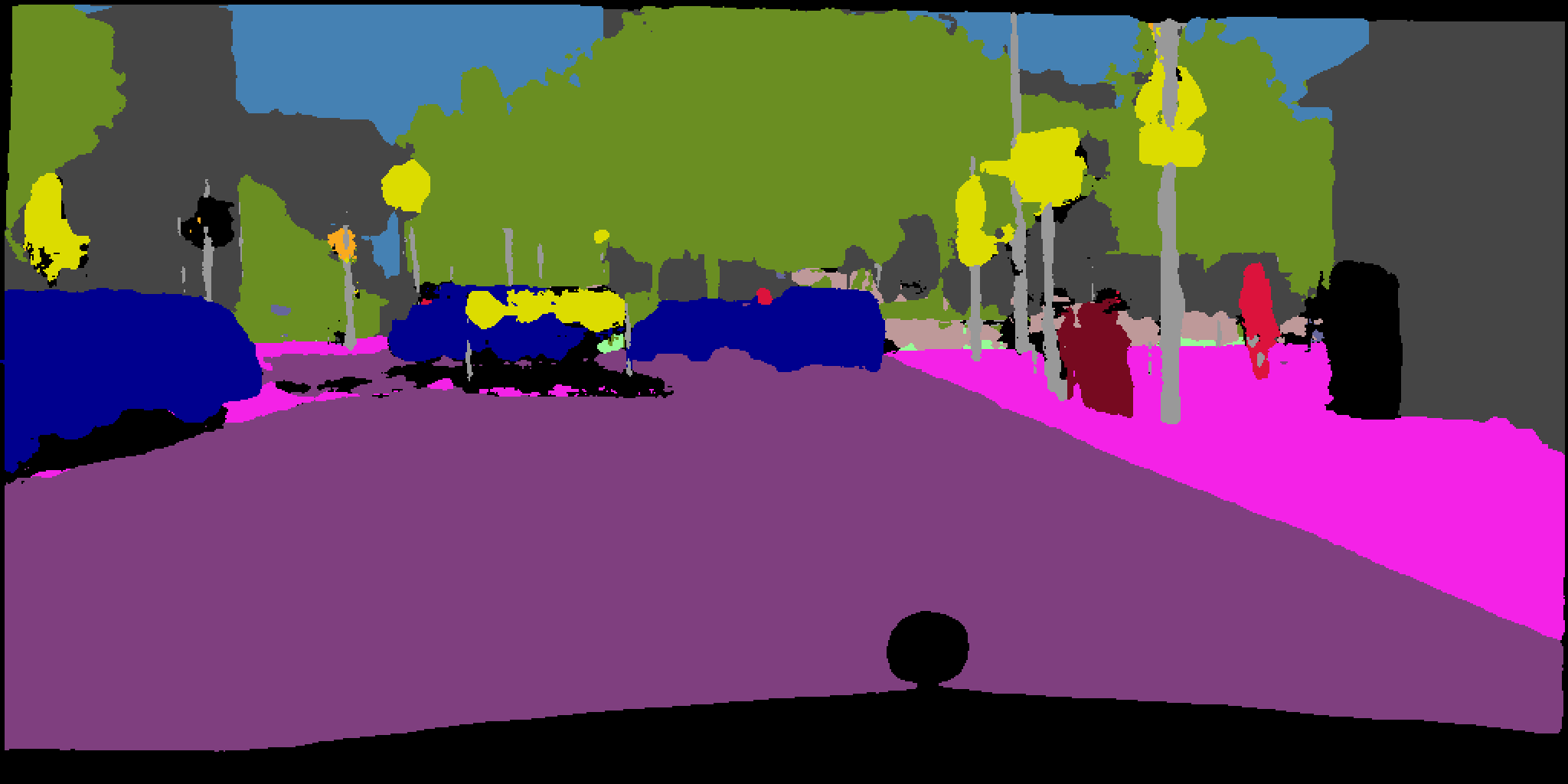}}
\subfigure{\includegraphics[width=43mm]{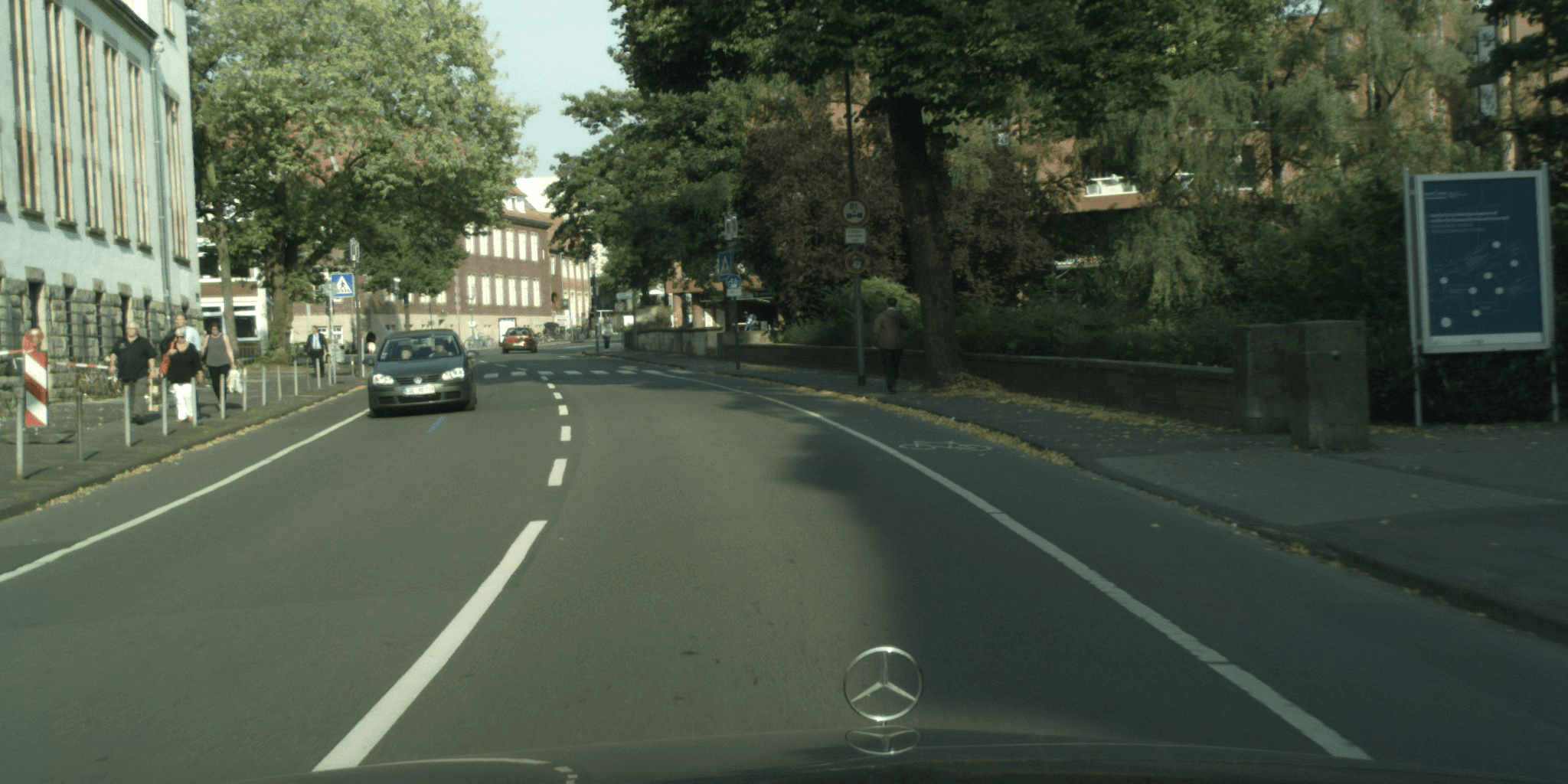}}
\subfigure{\includegraphics[width=43mm]{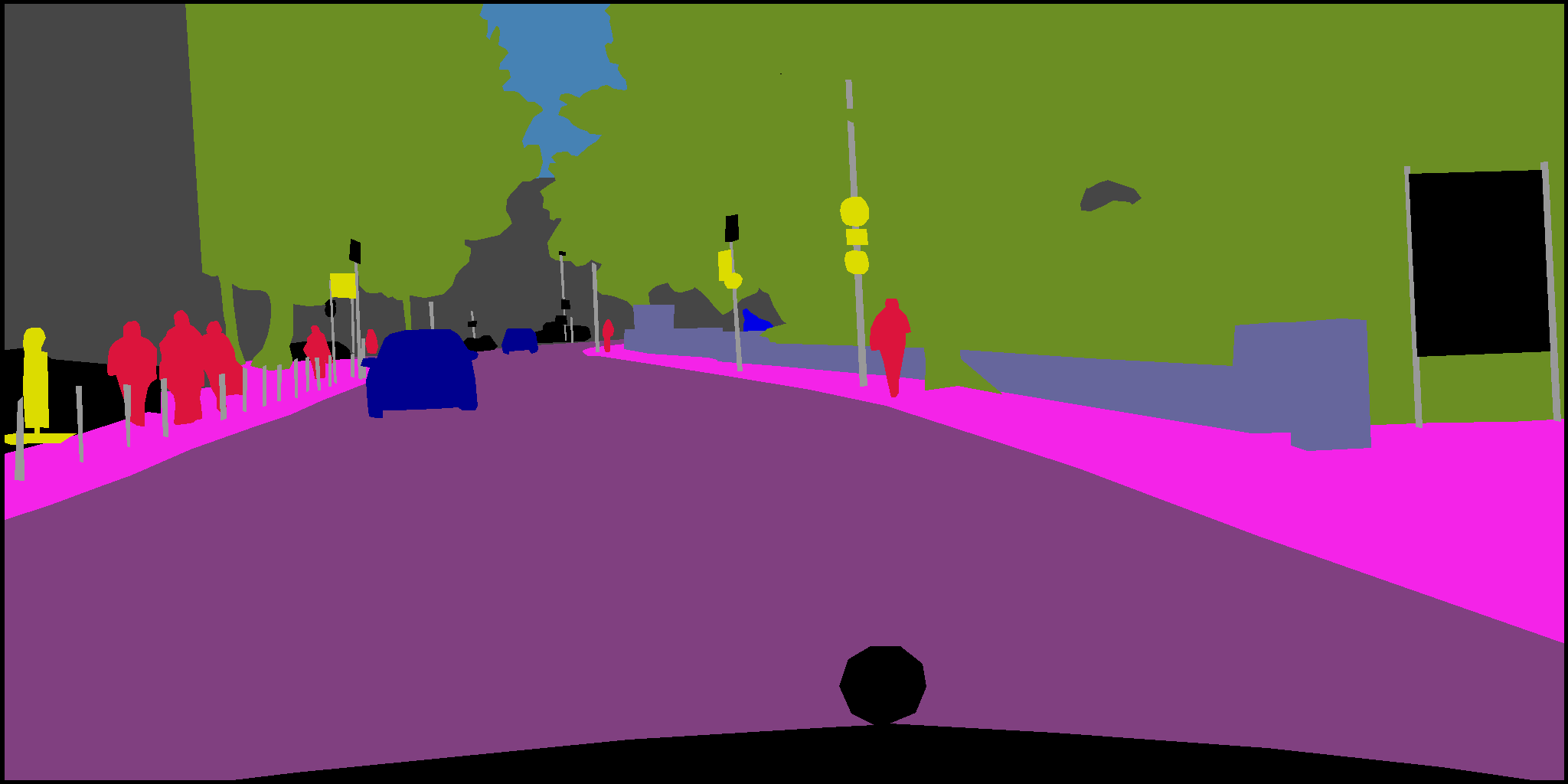}}
\subfigure{\includegraphics[width=43mm]{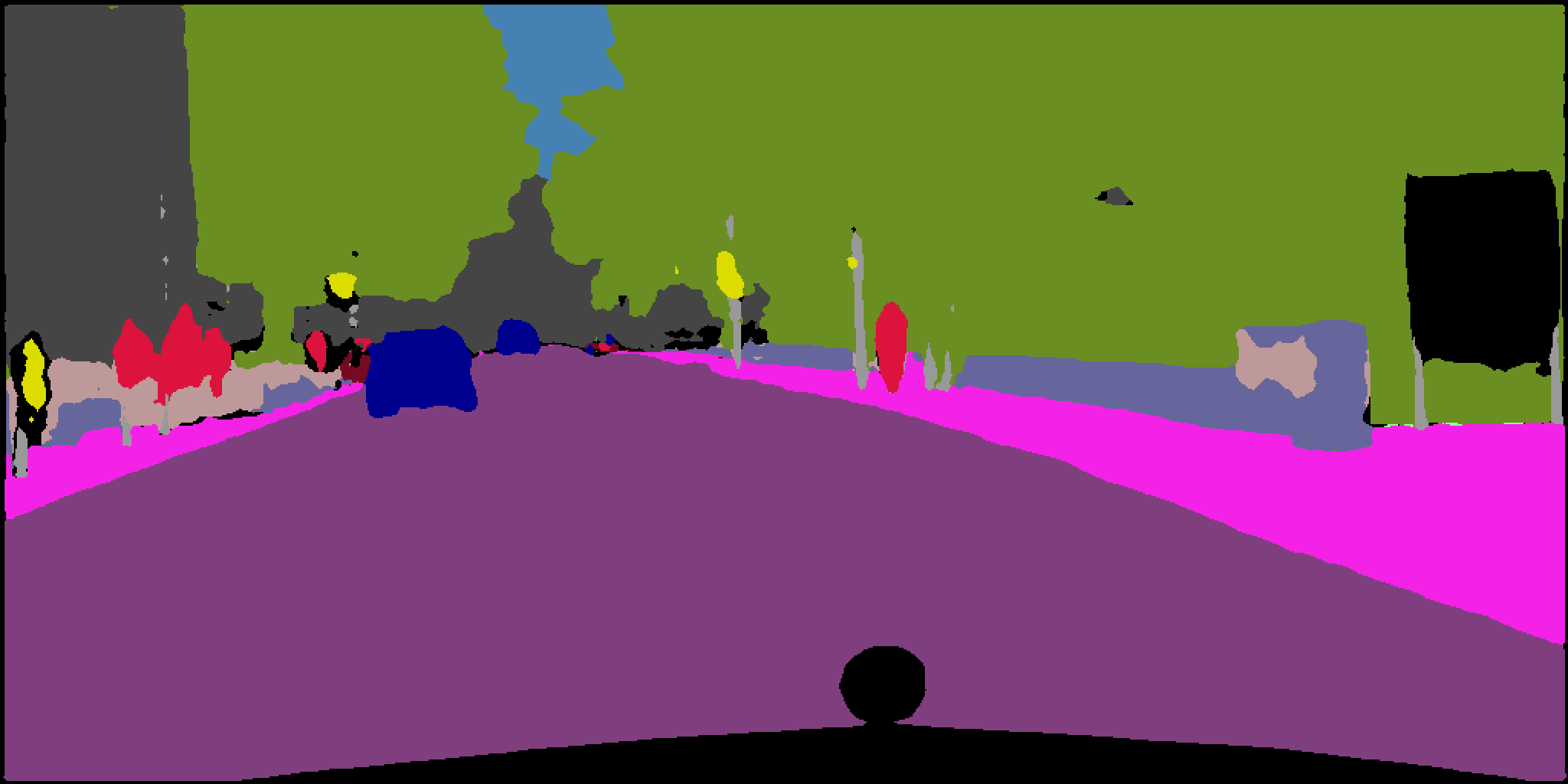}}
\subfigure{\includegraphics[width=43mm]{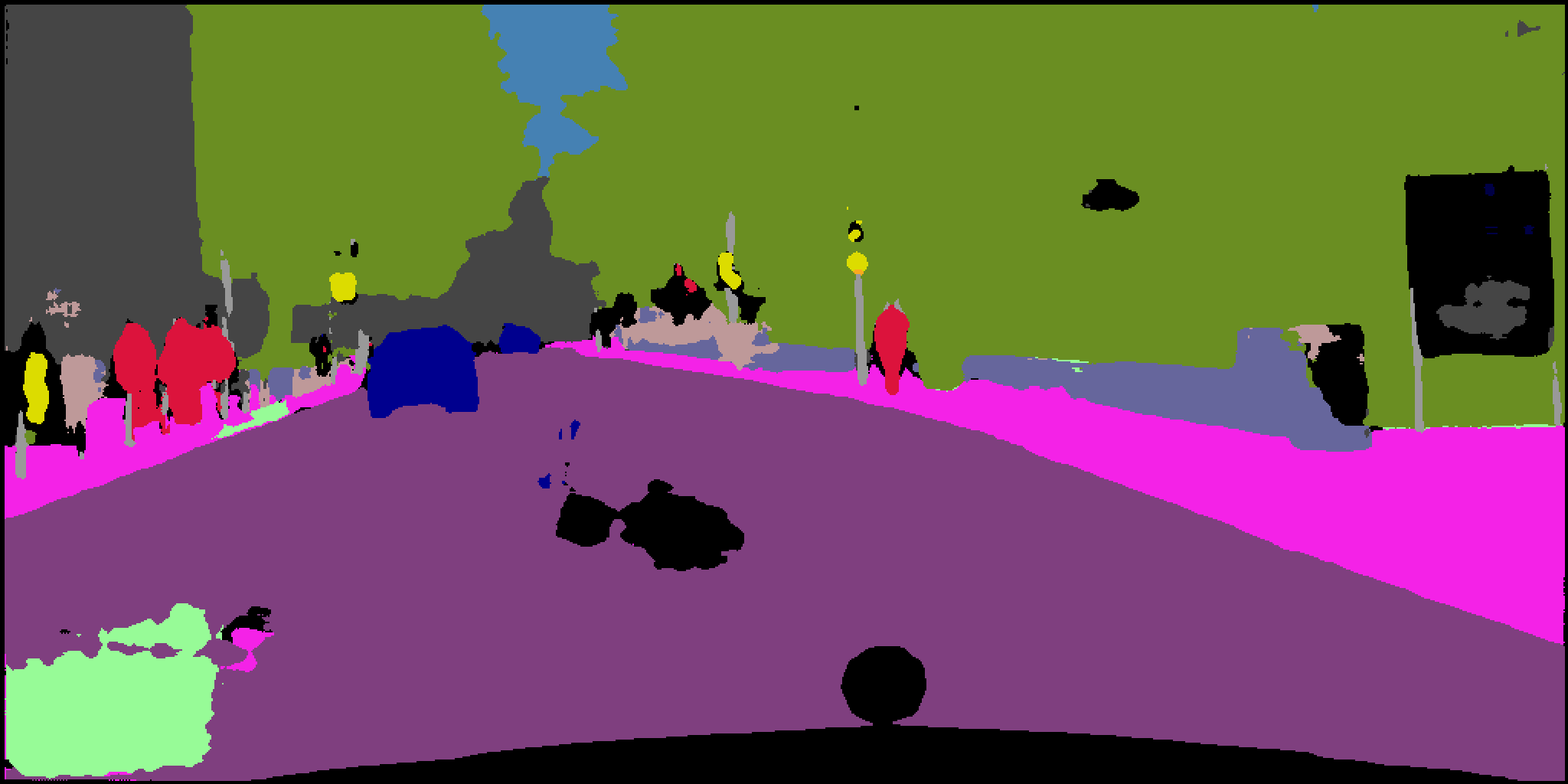}}
\subfigure{\includegraphics[width=43mm]{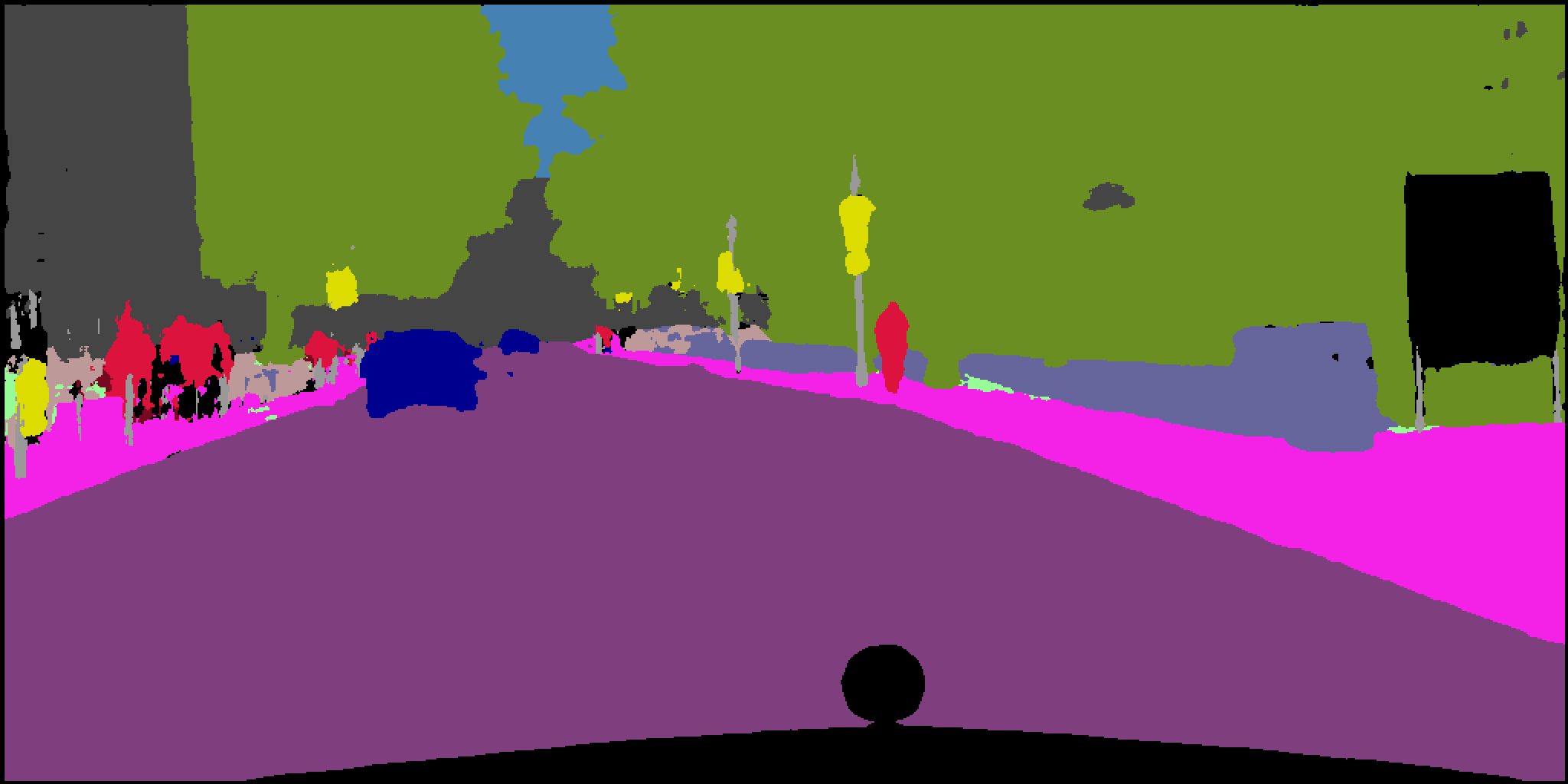}}
\subfigure{\includegraphics[width=43mm]{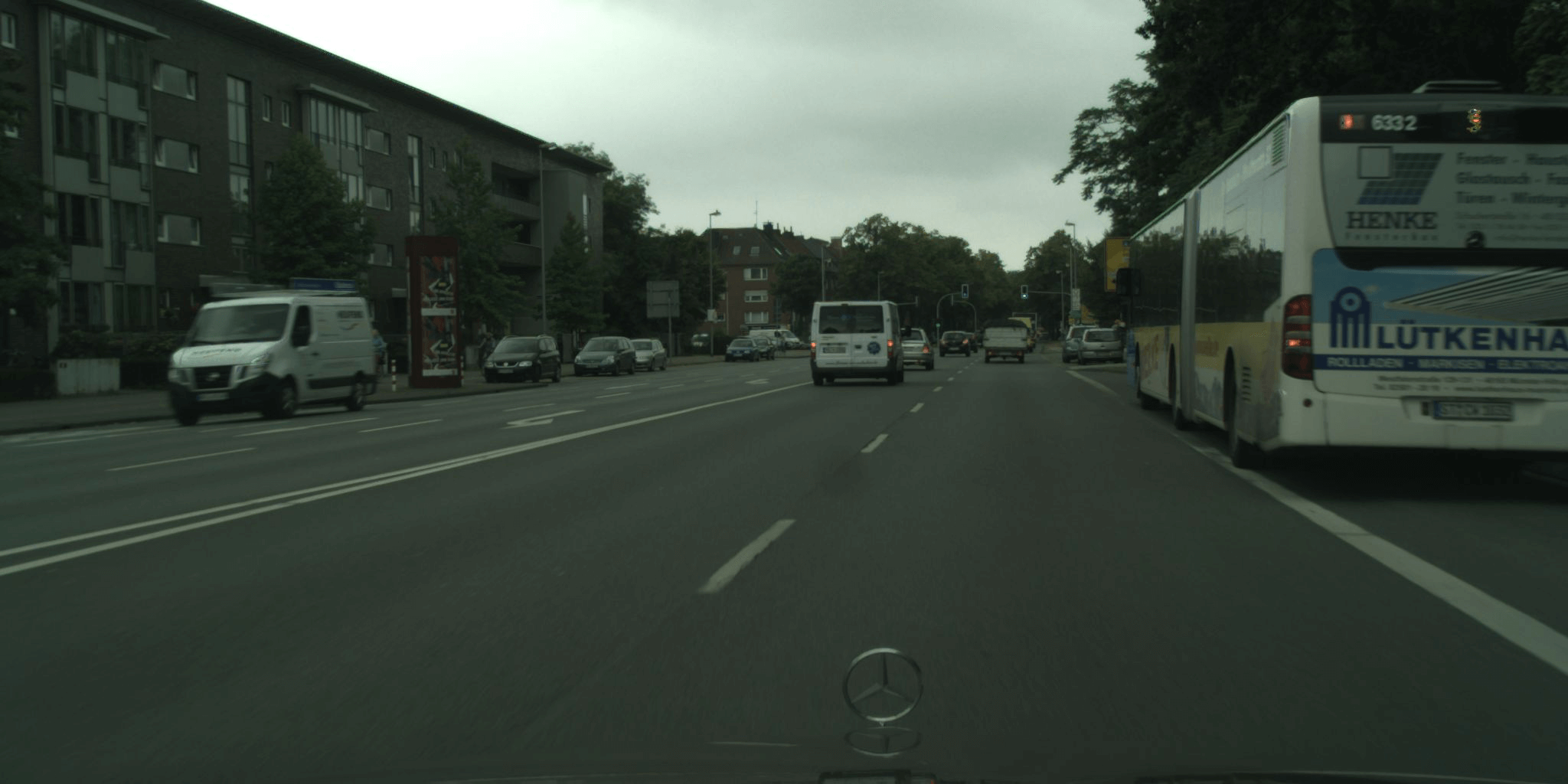}}
\subfigure{\includegraphics[width=43mm]{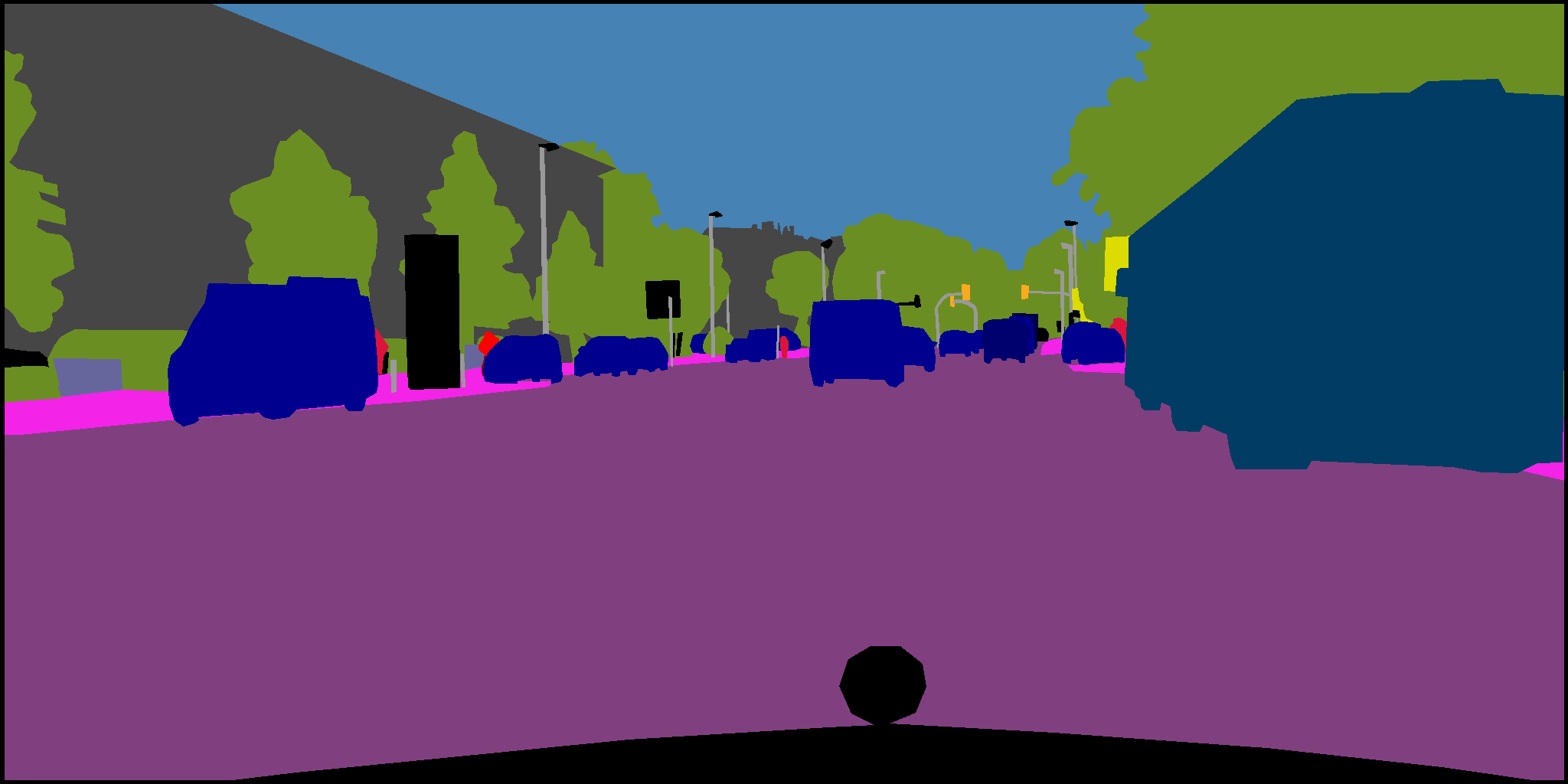}}
\subfigure{\includegraphics[width=43mm]{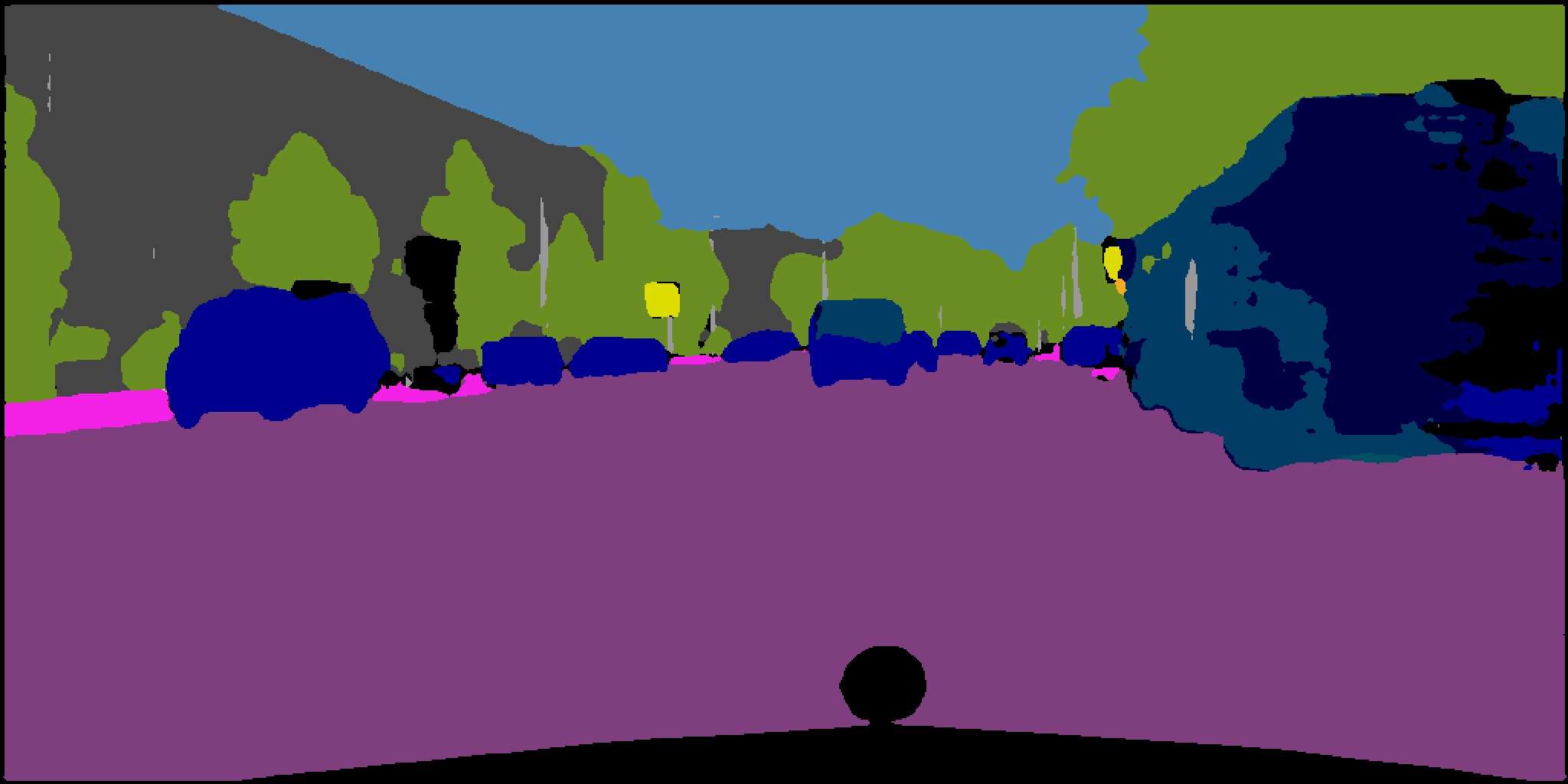}}
\subfigure{\includegraphics[width=43mm]{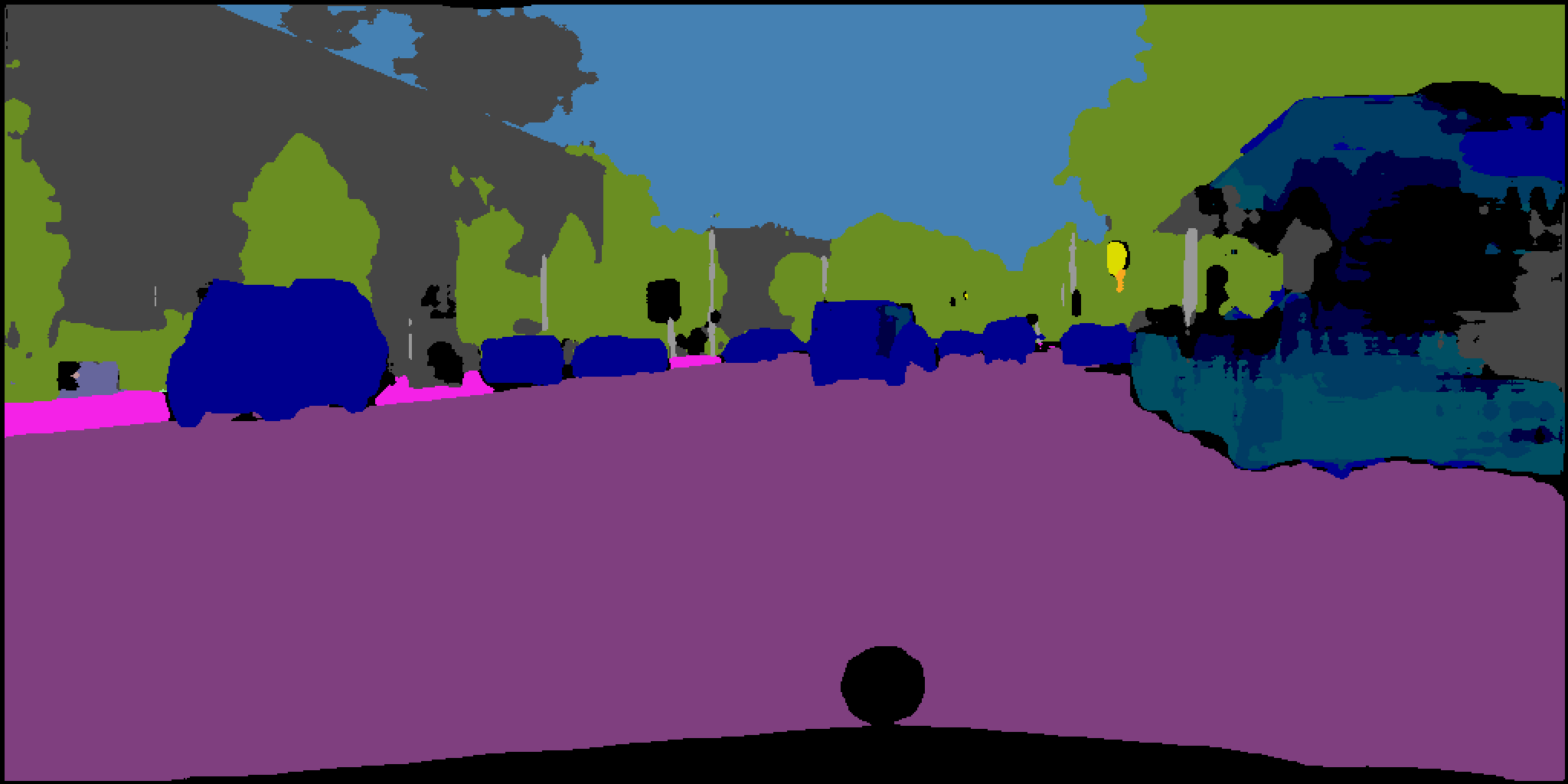}}
\subfigure{\includegraphics[width=43mm]{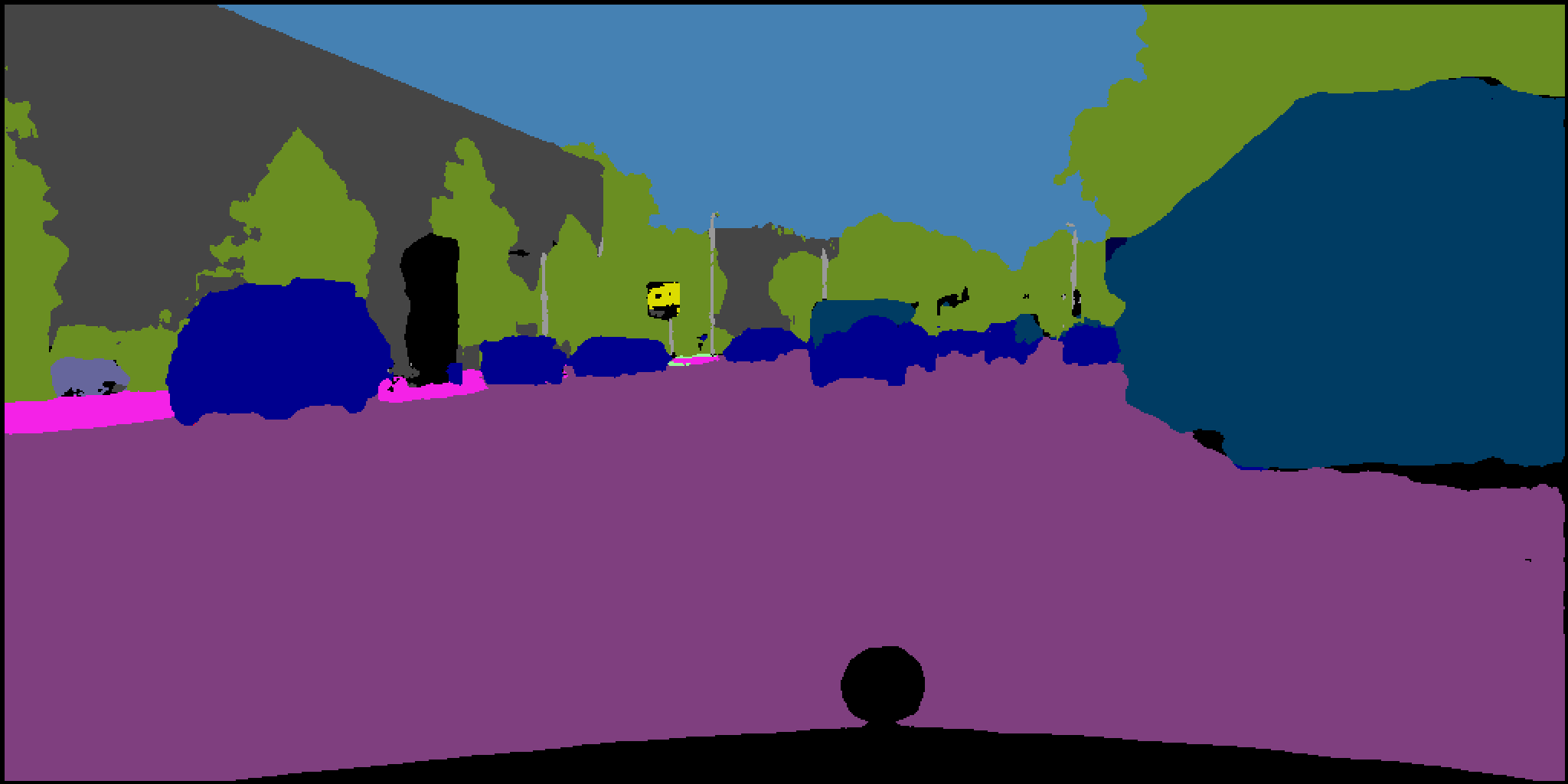}}
\subfigure{\includegraphics[width=43mm]{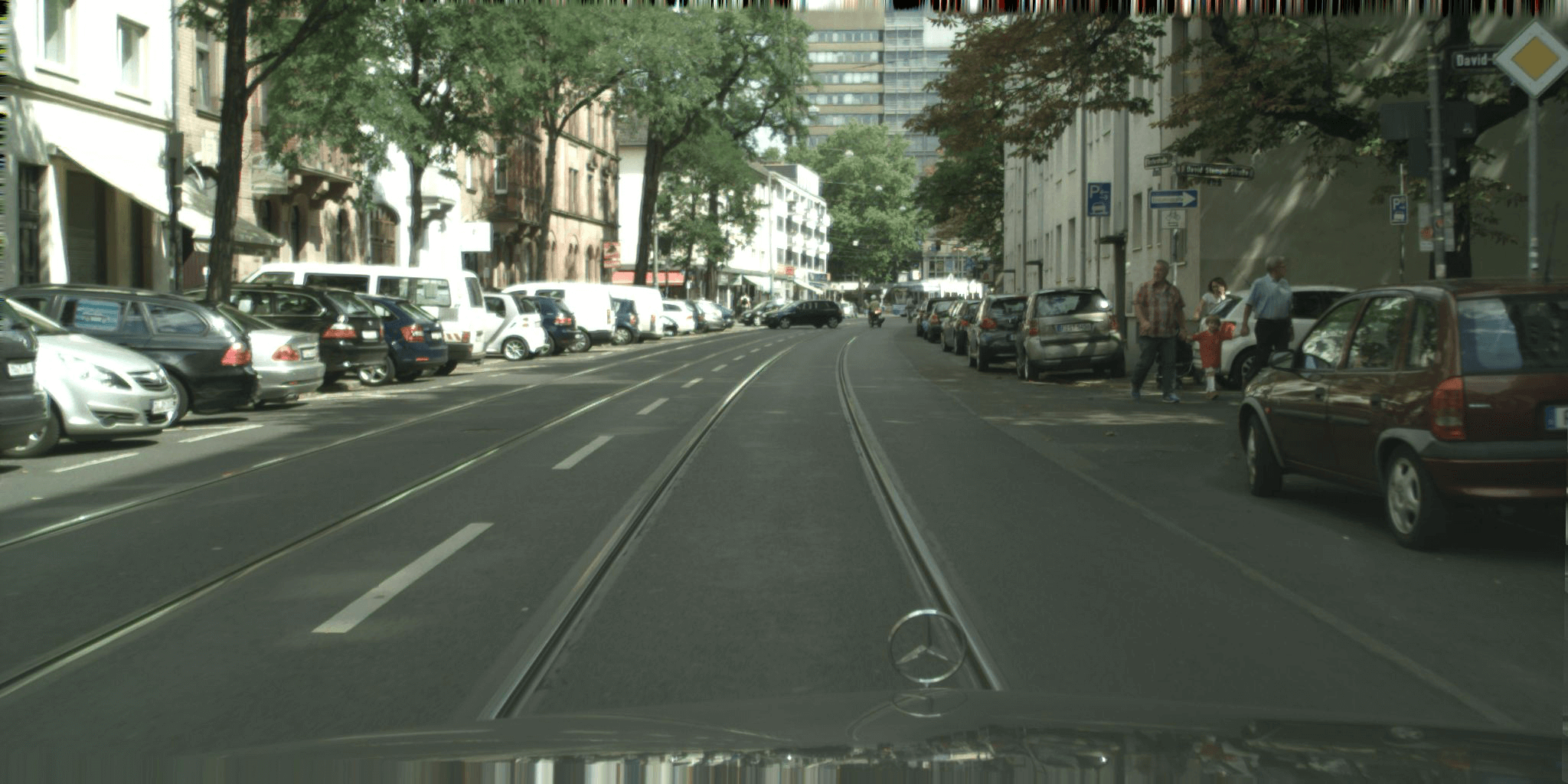}}
\subfigure{\includegraphics[width=43mm]{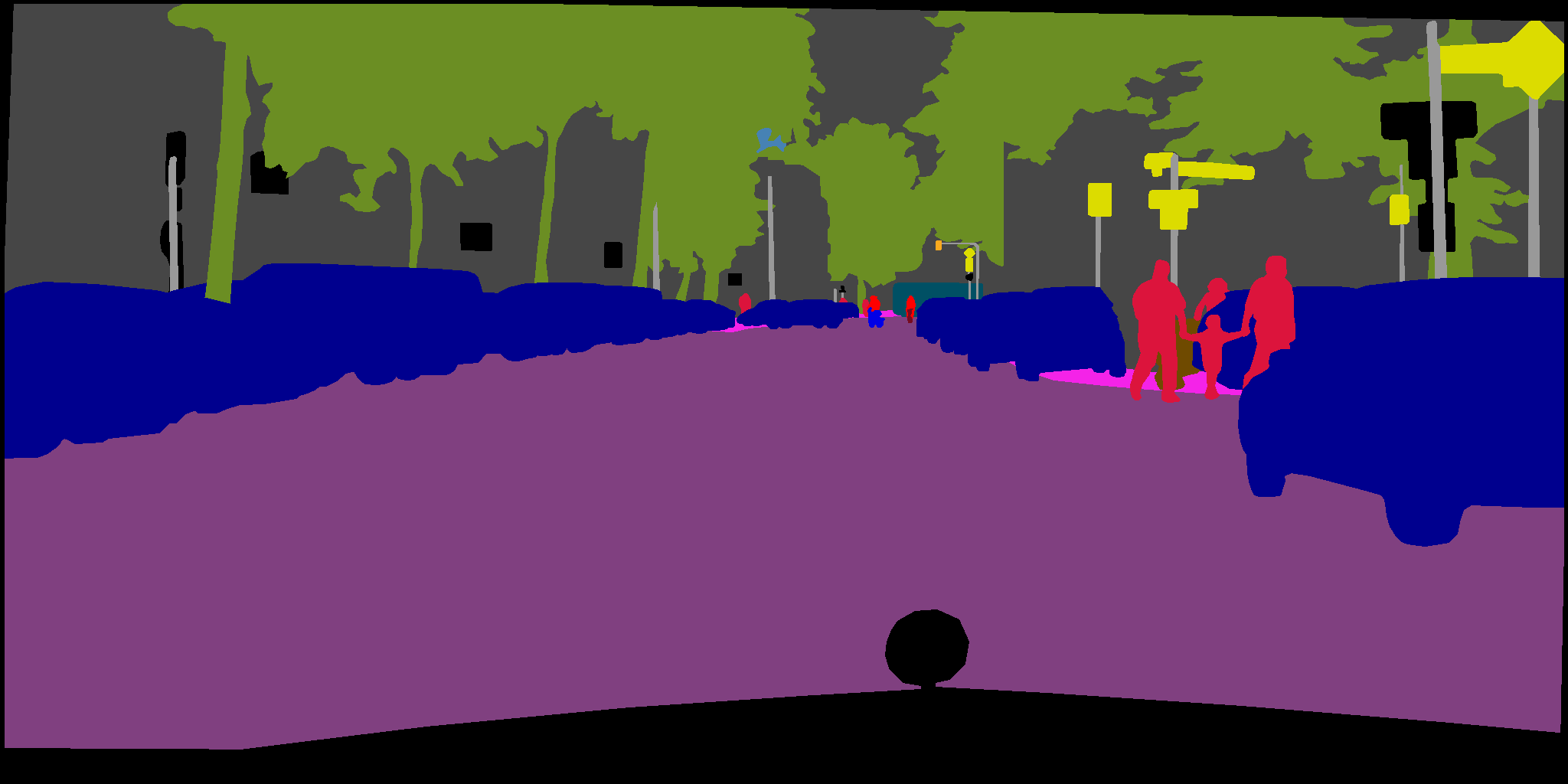}}
\subfigure{\includegraphics[width=43mm]{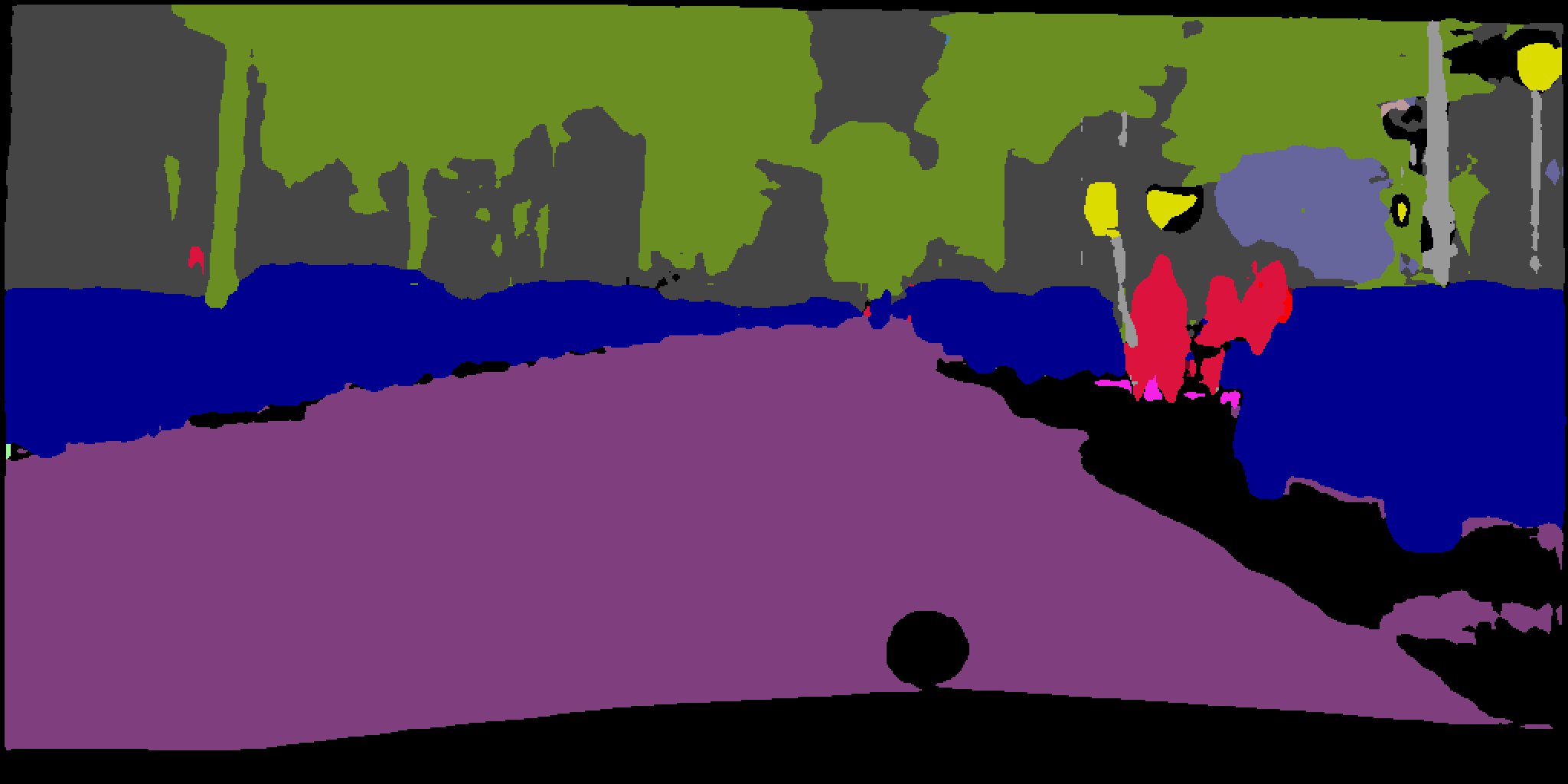}}
\subfigure{\includegraphics[width=43mm]{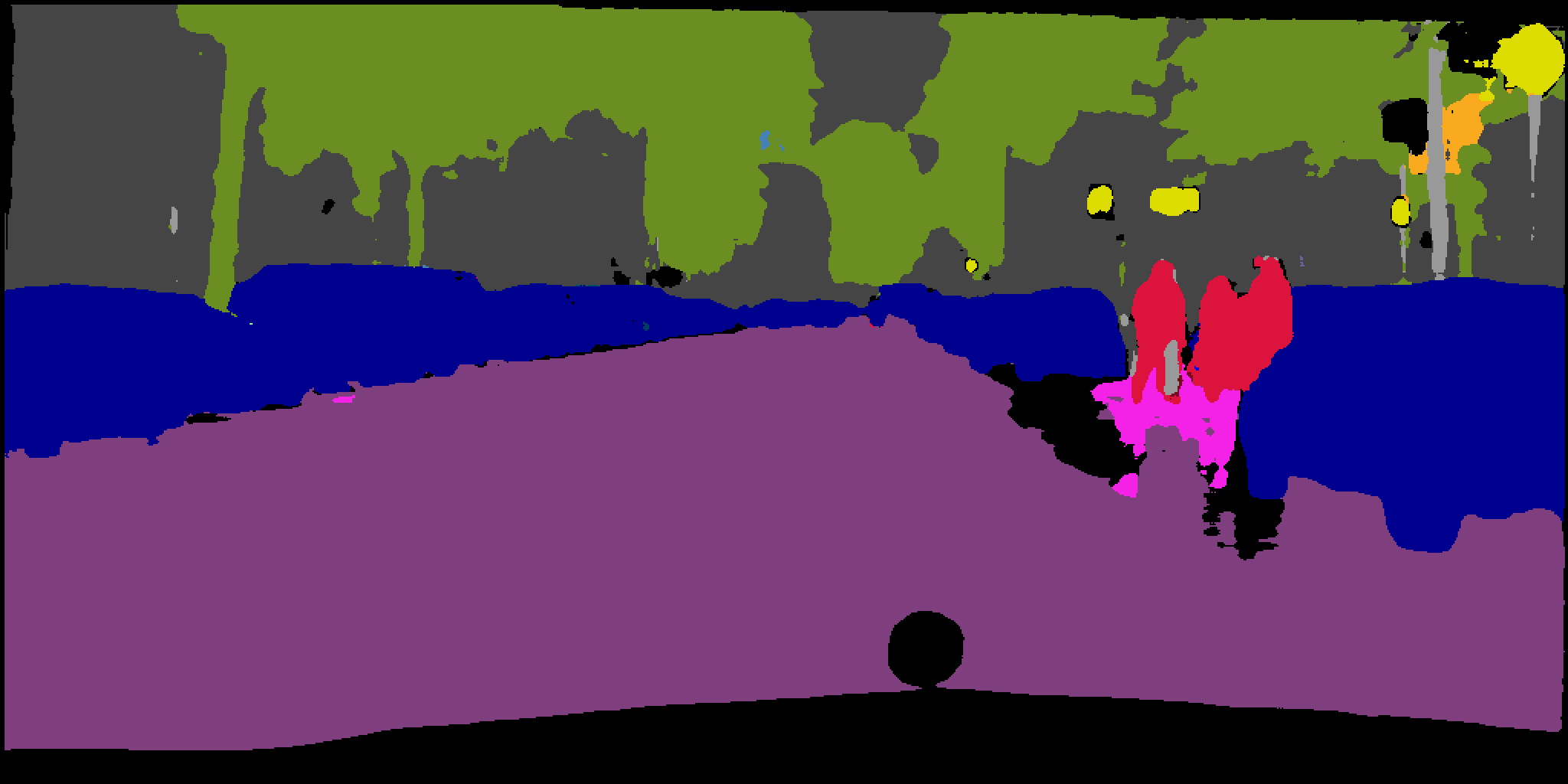}}
\subfigure{\includegraphics[width=43mm]{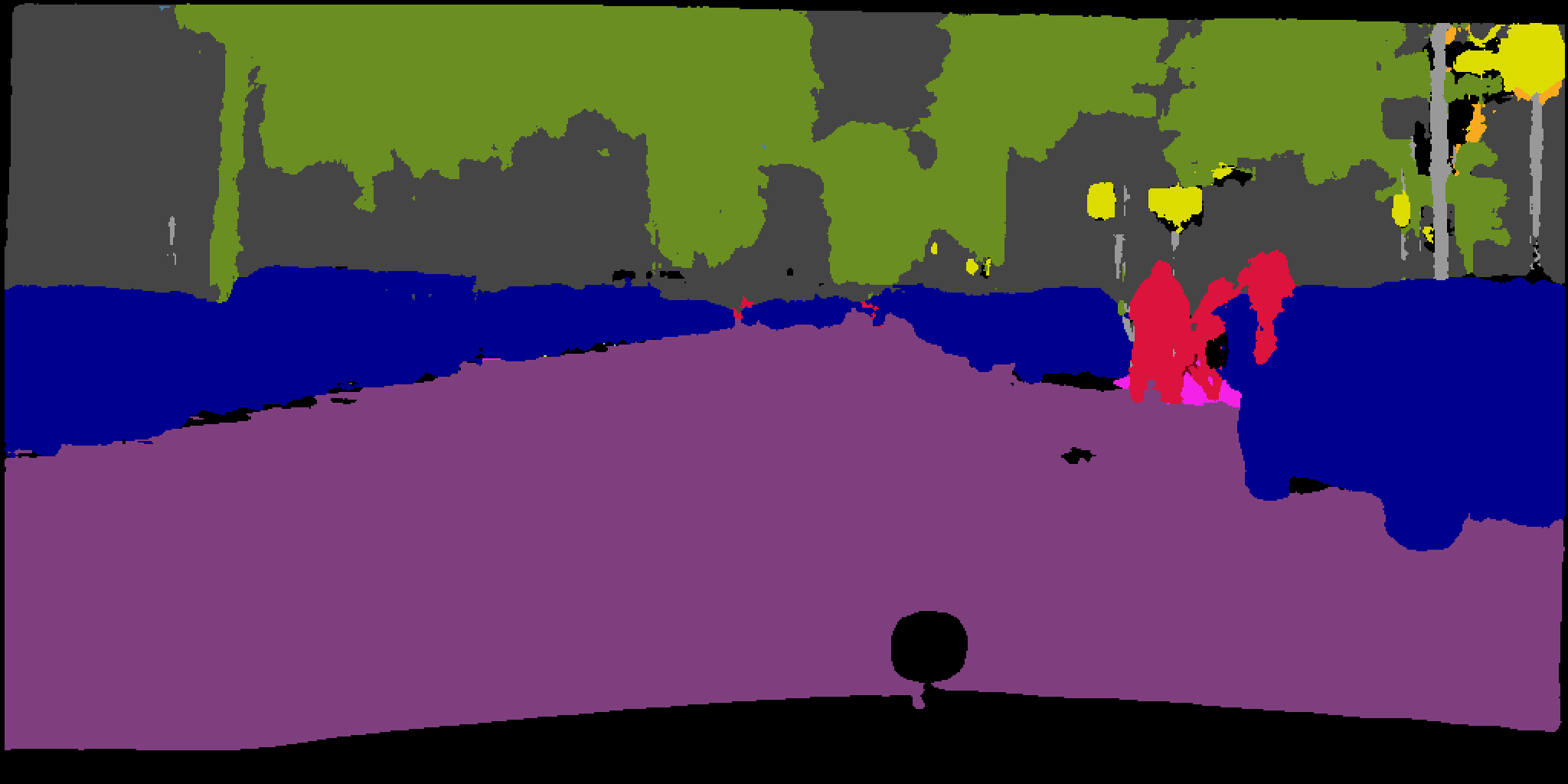}}
\subfigure{\includegraphics[width=43mm]{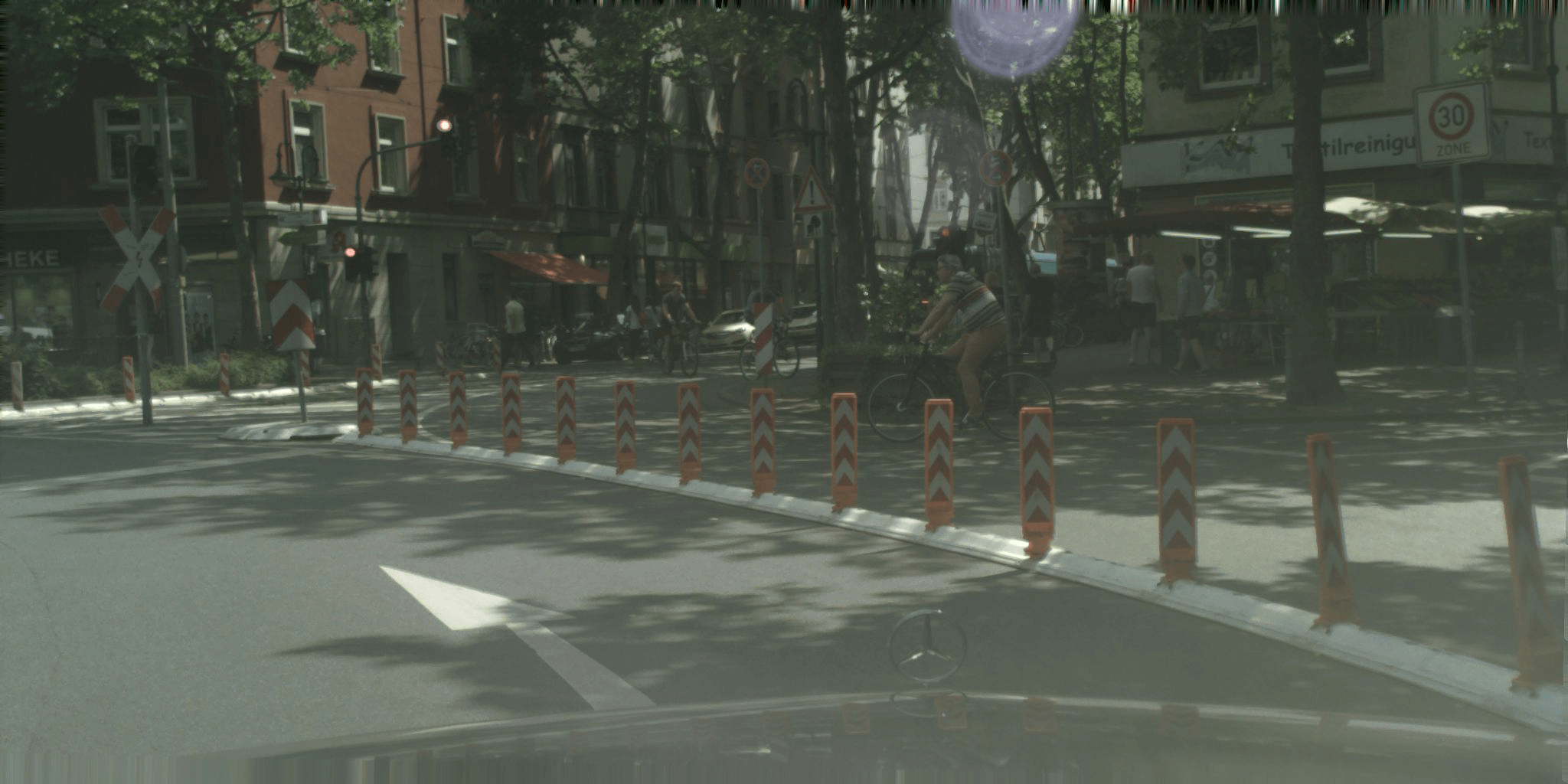}}
\subfigure{\includegraphics[width=43mm]{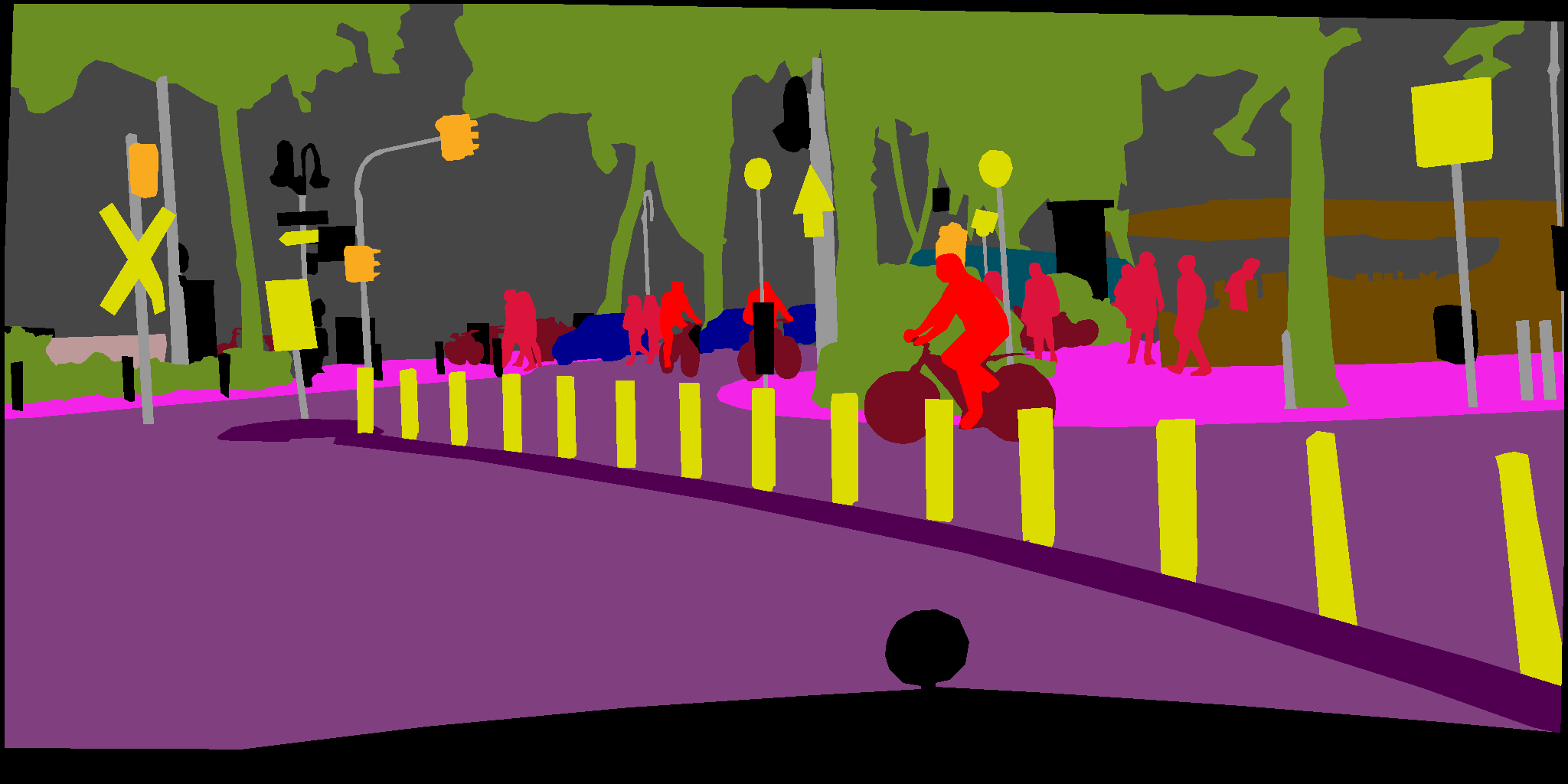}}
\subfigure{\includegraphics[width=43mm]{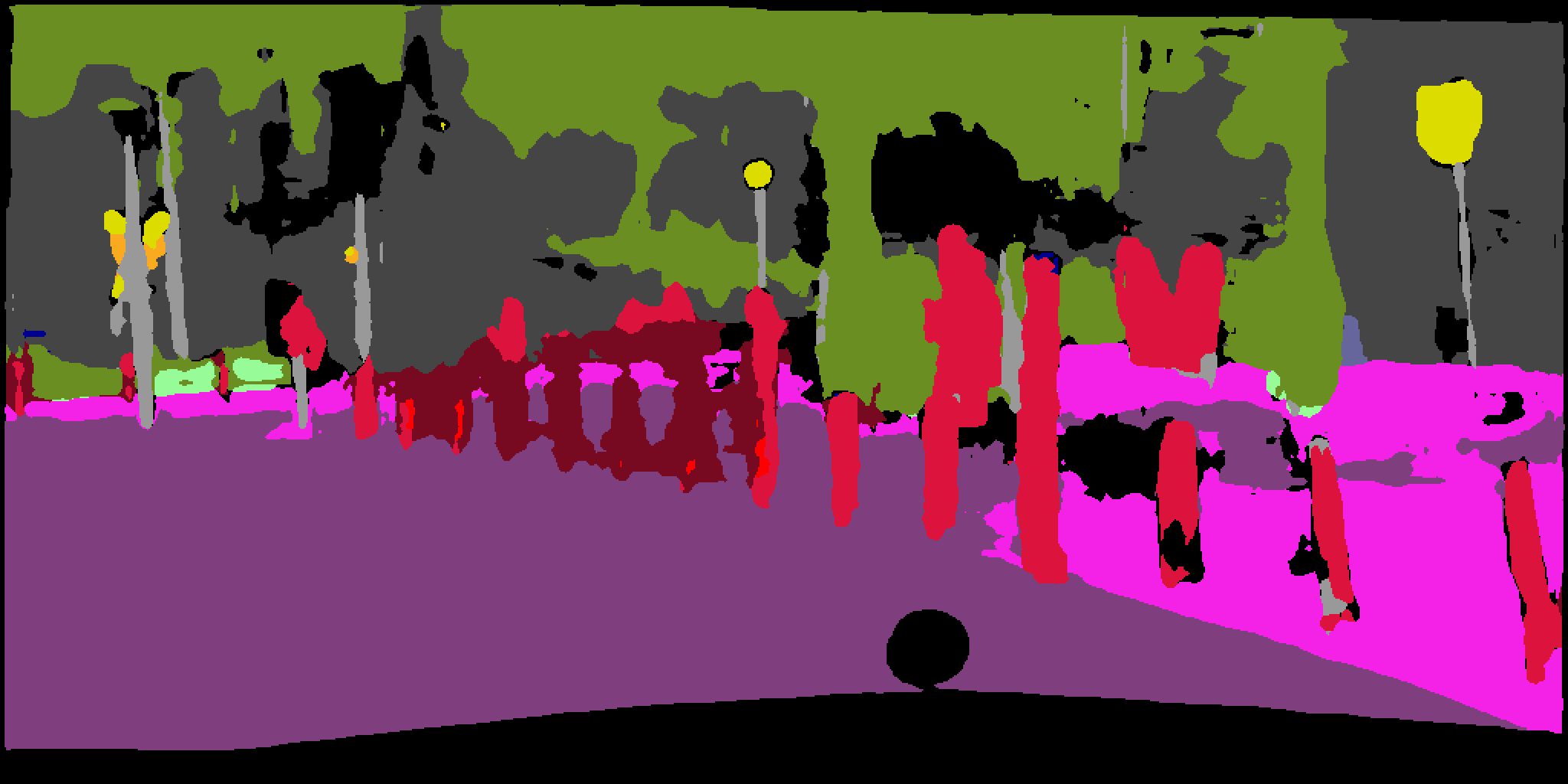}}
\subfigure{\includegraphics[width=43mm]{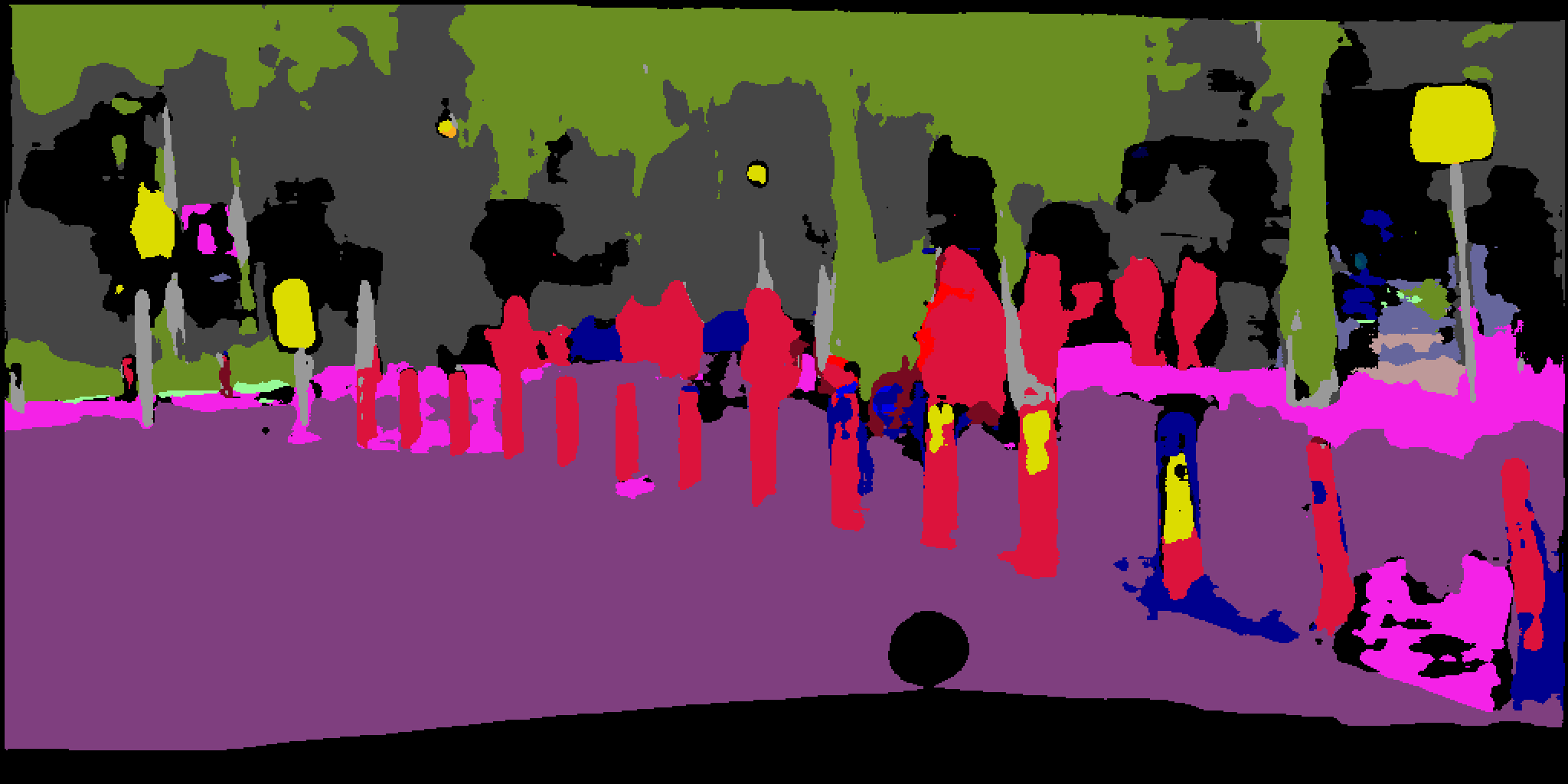}}
\subfigure{\includegraphics[width=43mm]{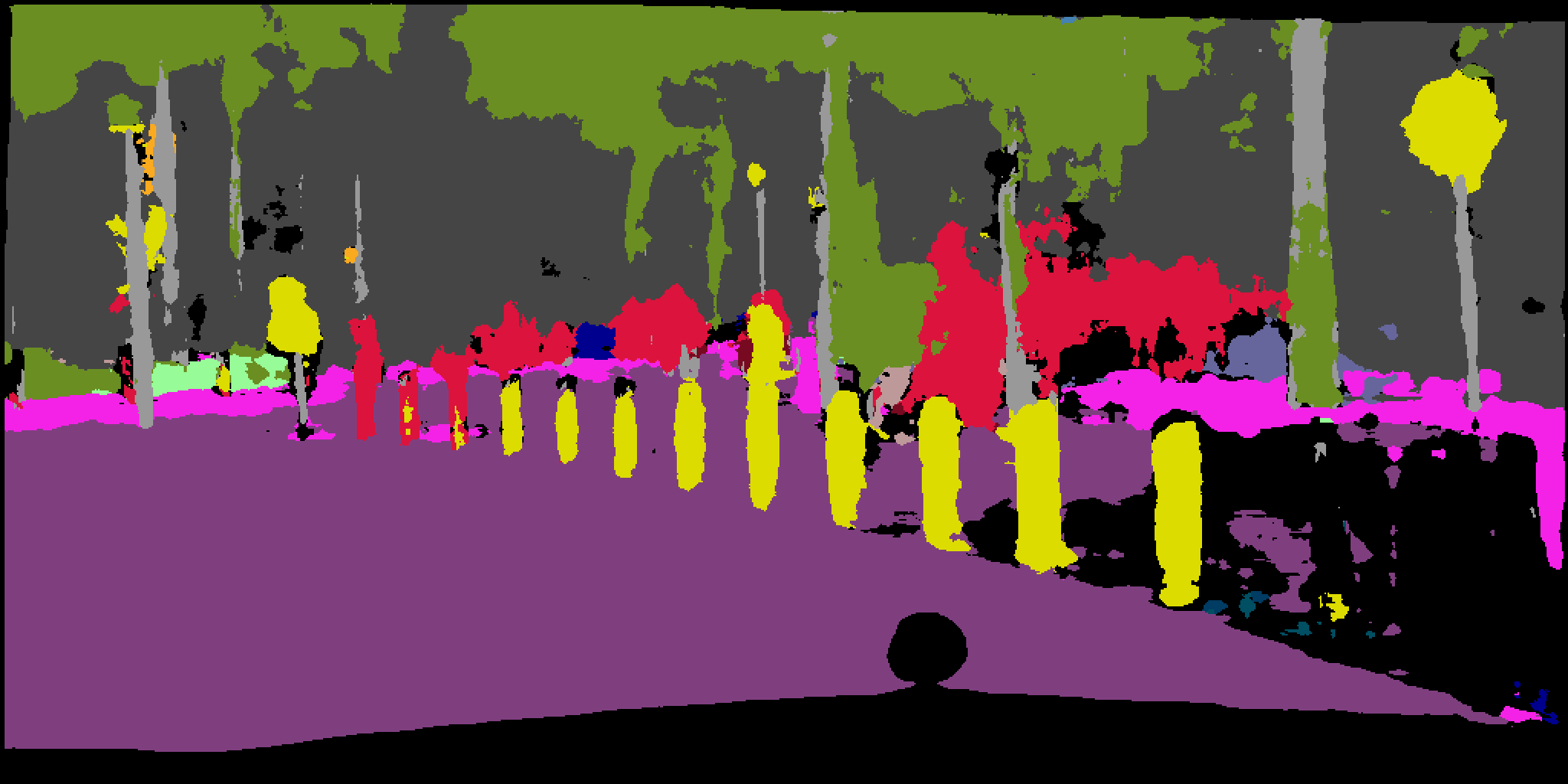}}
\subfigure{\includegraphics[width=43mm]{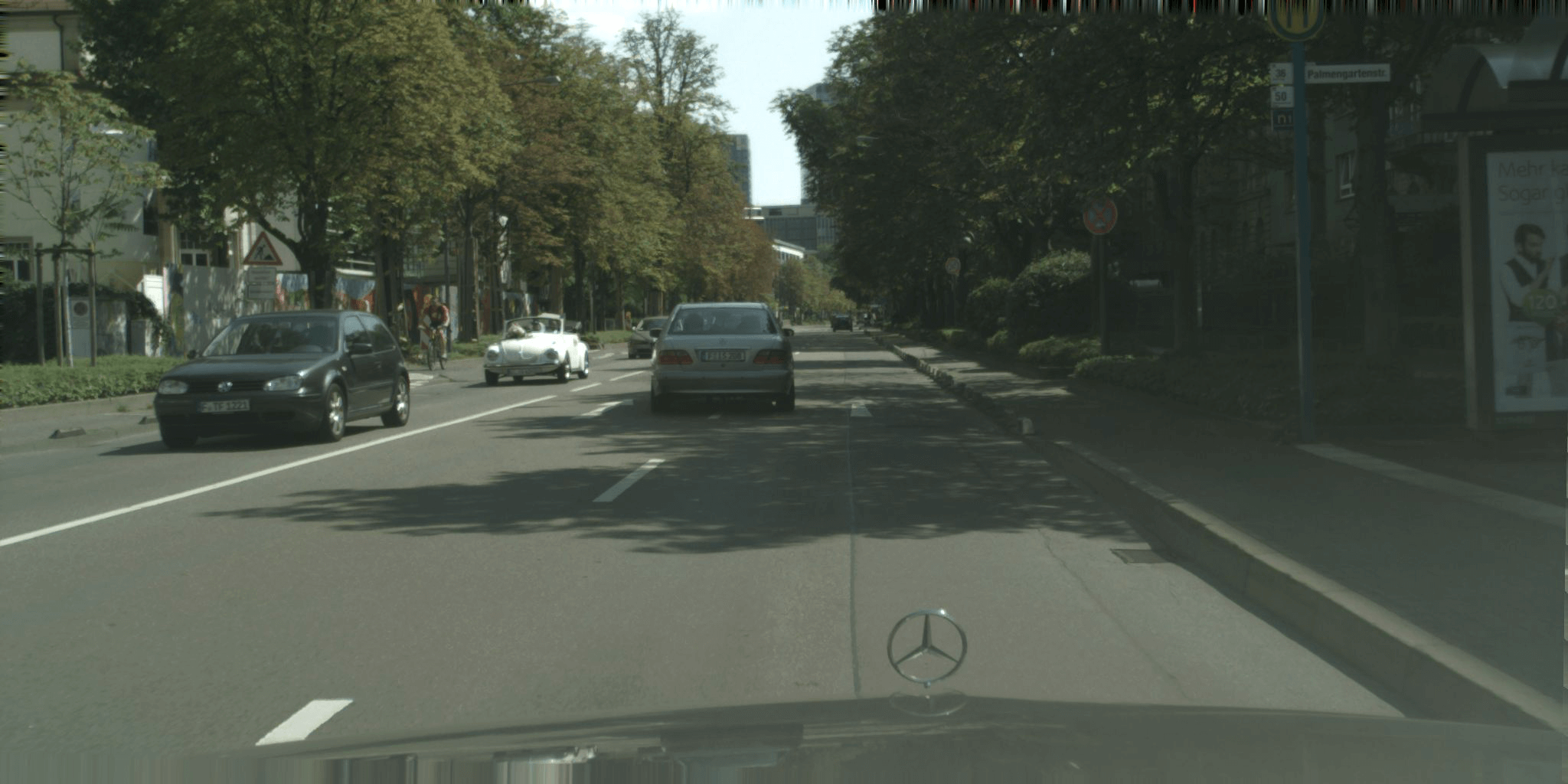}}
\subfigure{\includegraphics[width=43mm]{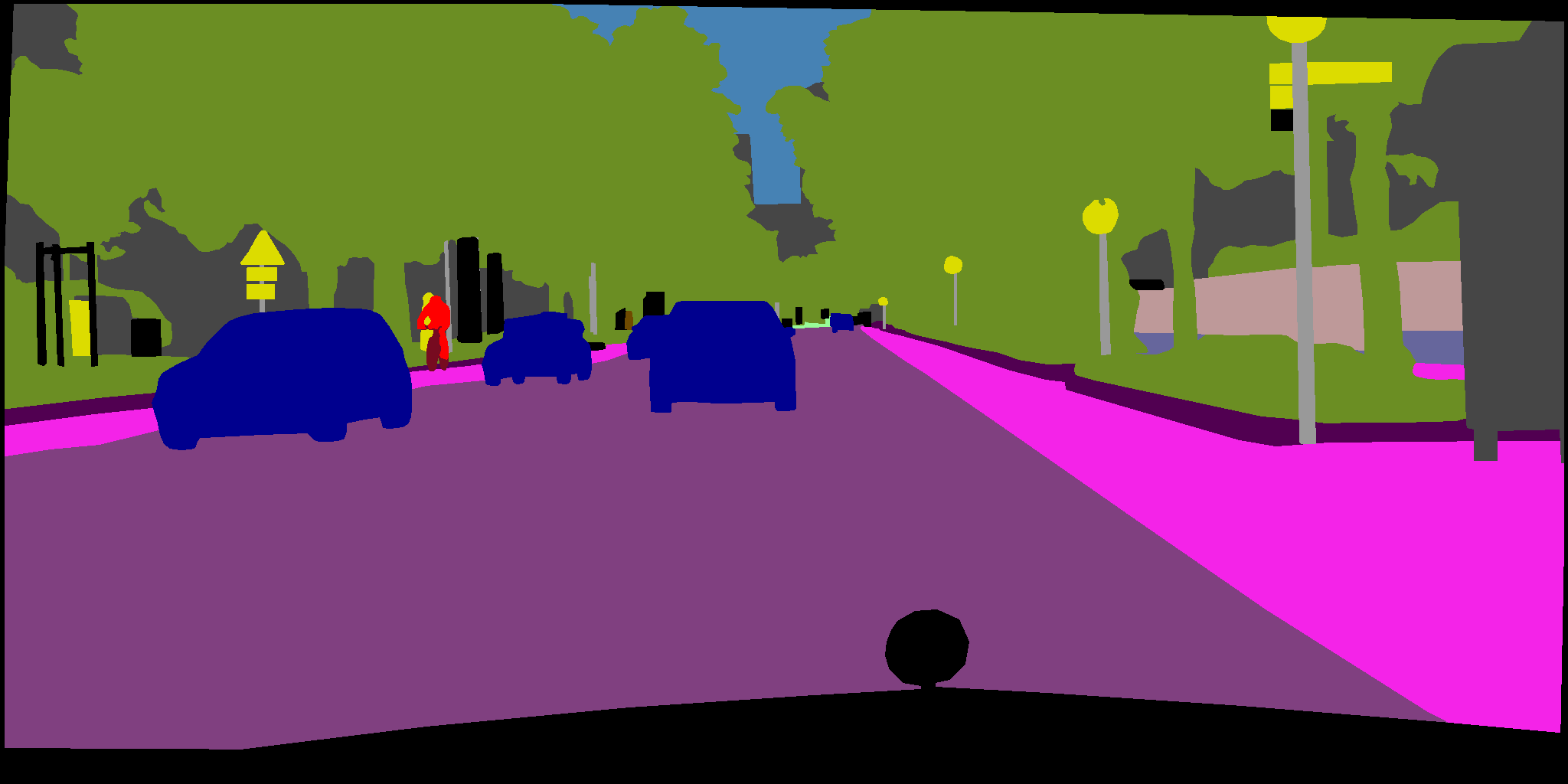}}
\subfigure{\includegraphics[width=43mm]{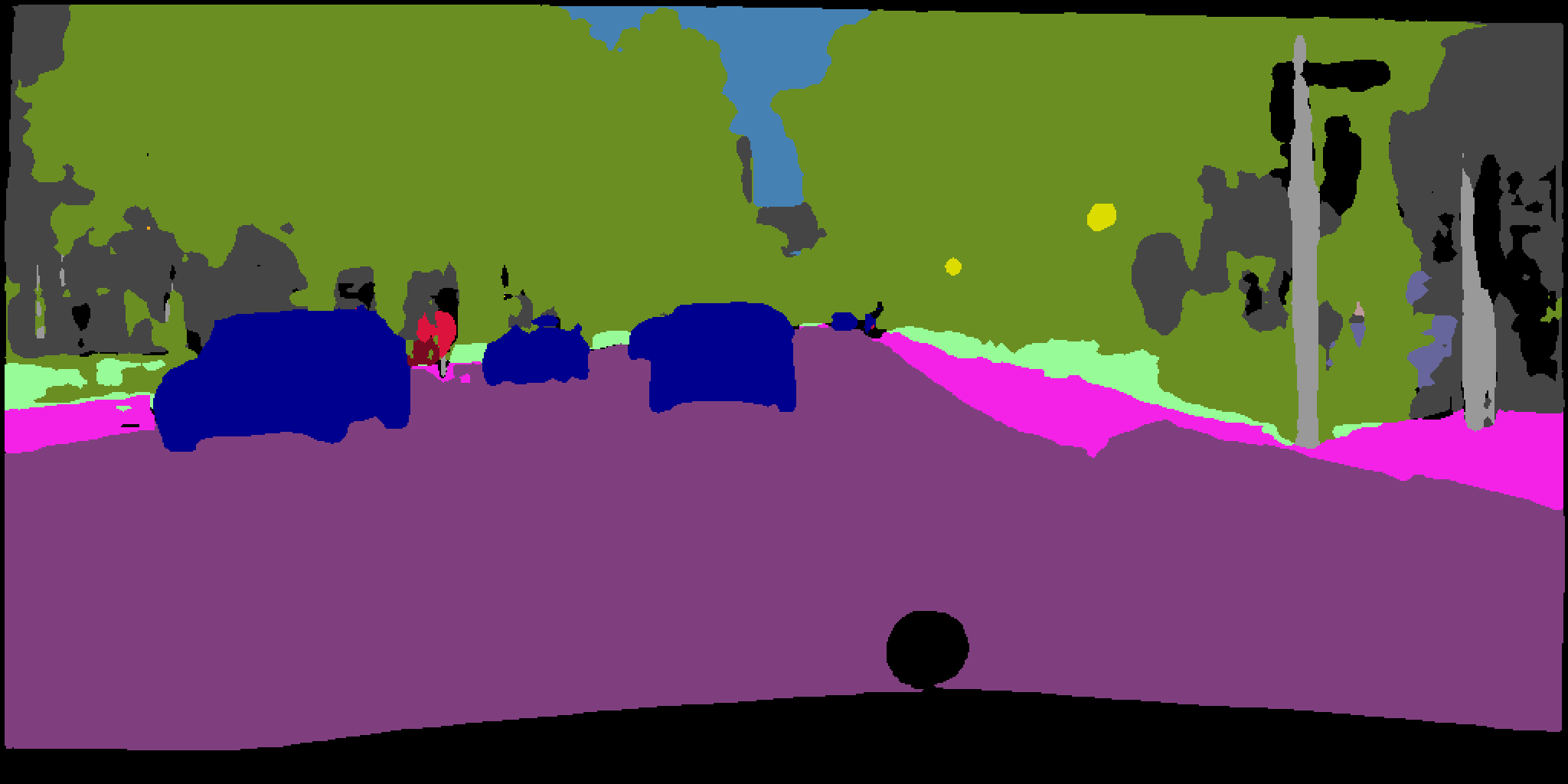}}
\subfigure{\includegraphics[width=43mm]{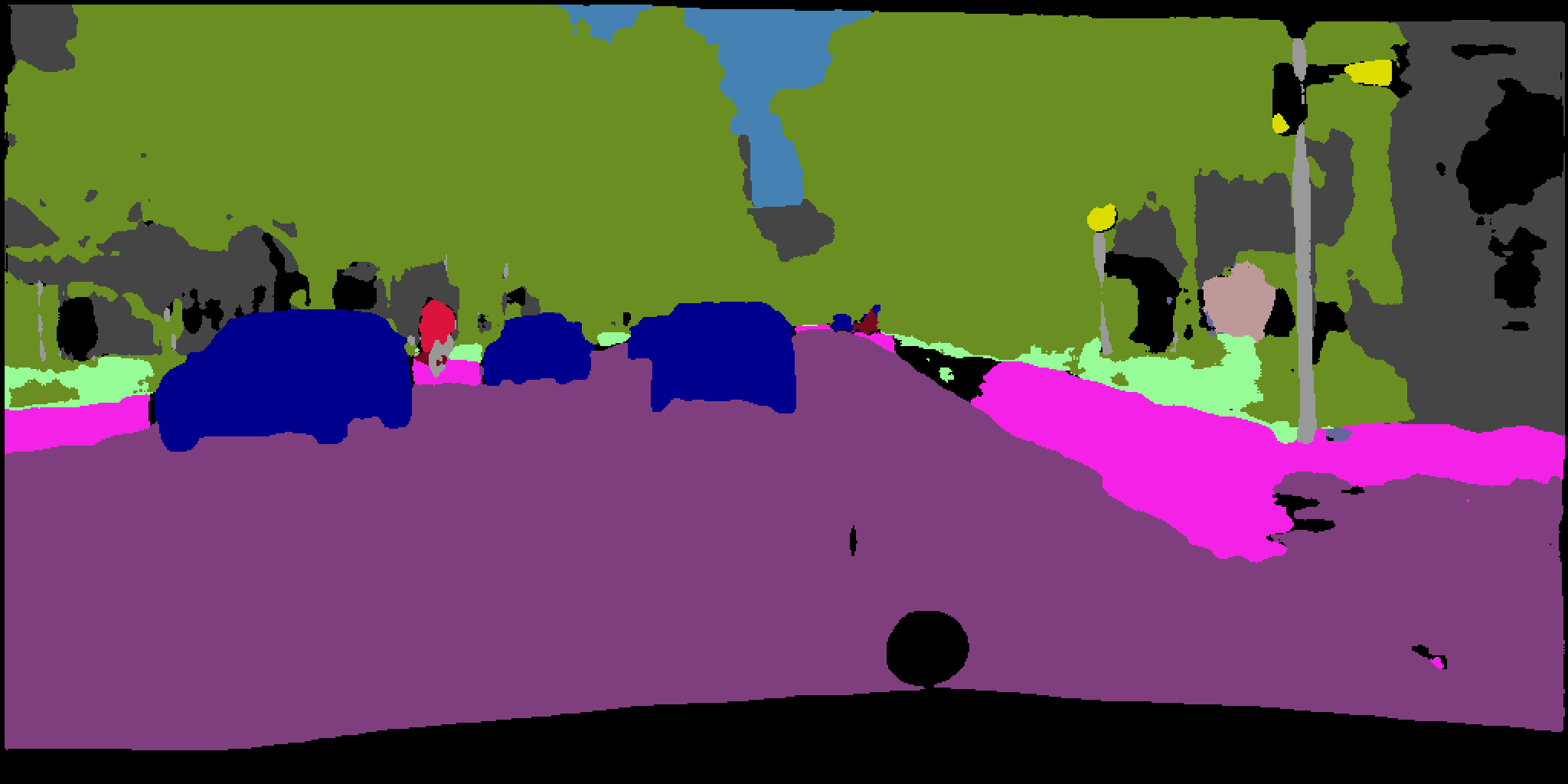}}
\subfigure{\includegraphics[width=43mm]{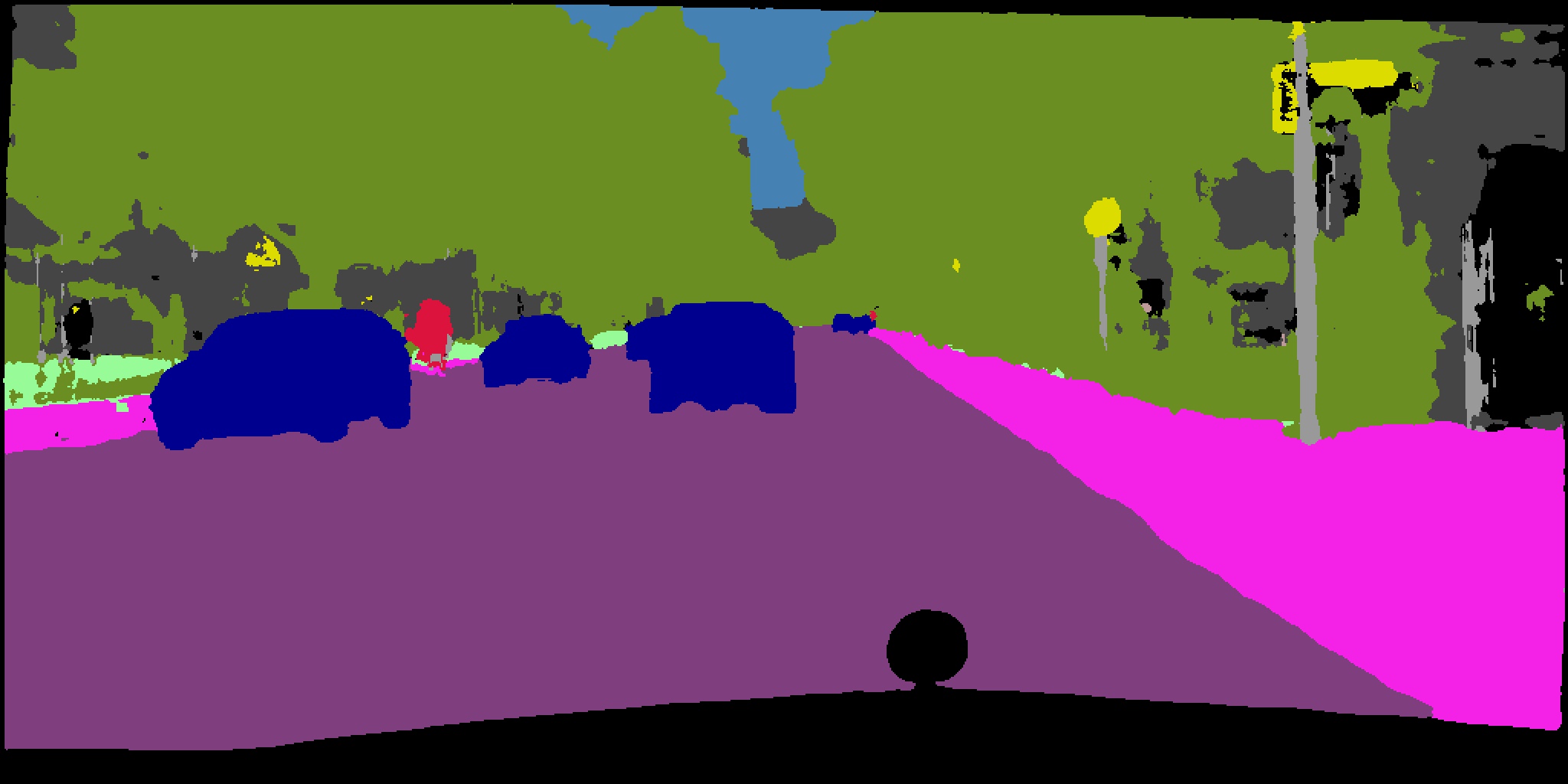}}
\caption{Qualitative results of Cityscapes baseline experiments (from the left-- input image, ground truth, HRNet-V2, U-Net, and PDFNet12 )}
\label{q1}
\end{sidewaysfigure*}

\noindent In Figure \ref{q1}, we present the qualitative results of the networks HRNet-V2 ($48.0$), U-Net ($49.3$), and PDFNet12 ($54.9$) from the Cityscapes baseline experiments.
\newpage
\section{CamVid}
\begin{figure*}
\centering     
\subfigure{\includegraphics[width=65mm]{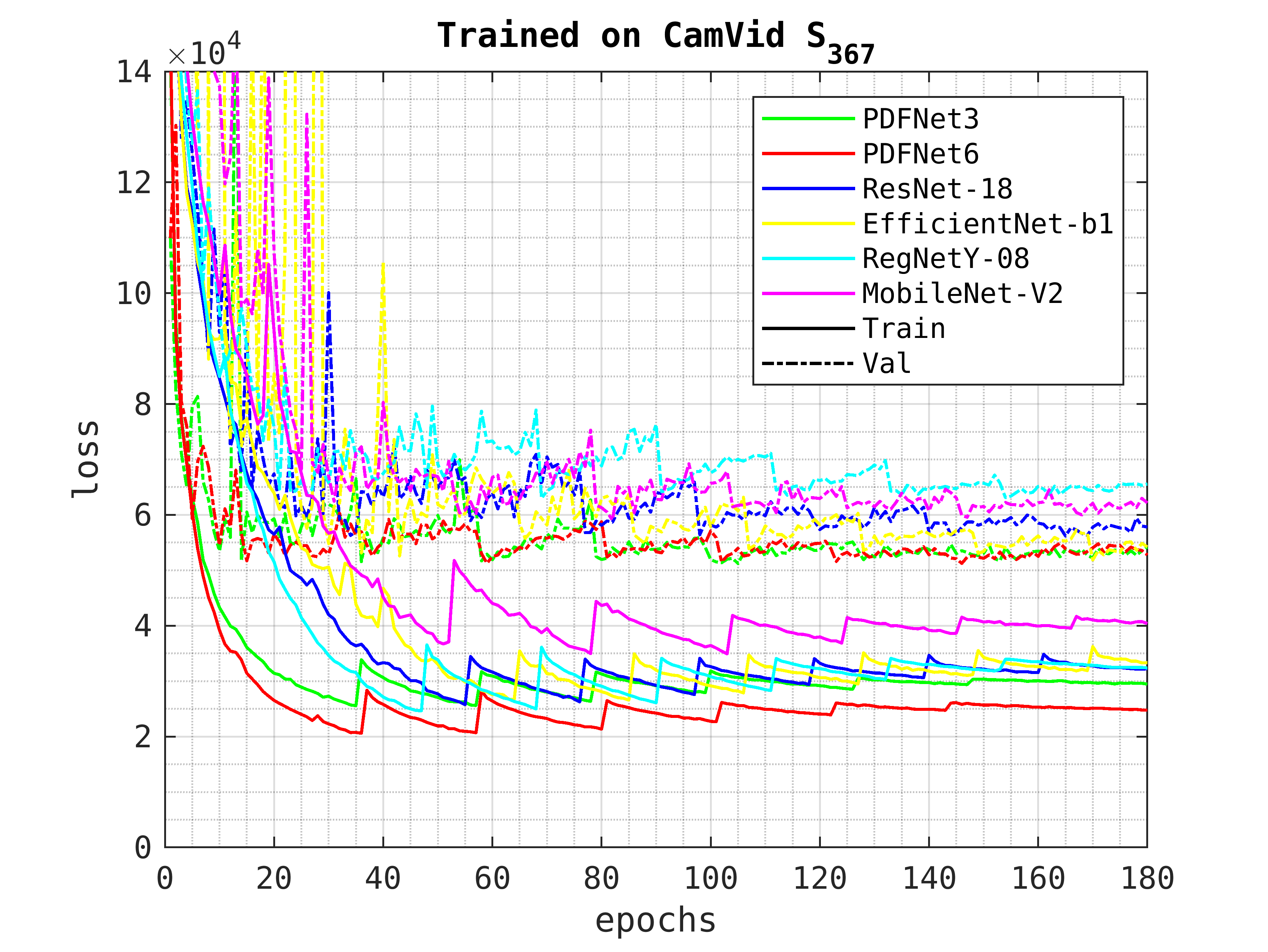}}
\subfigure{\includegraphics[width=65mm]{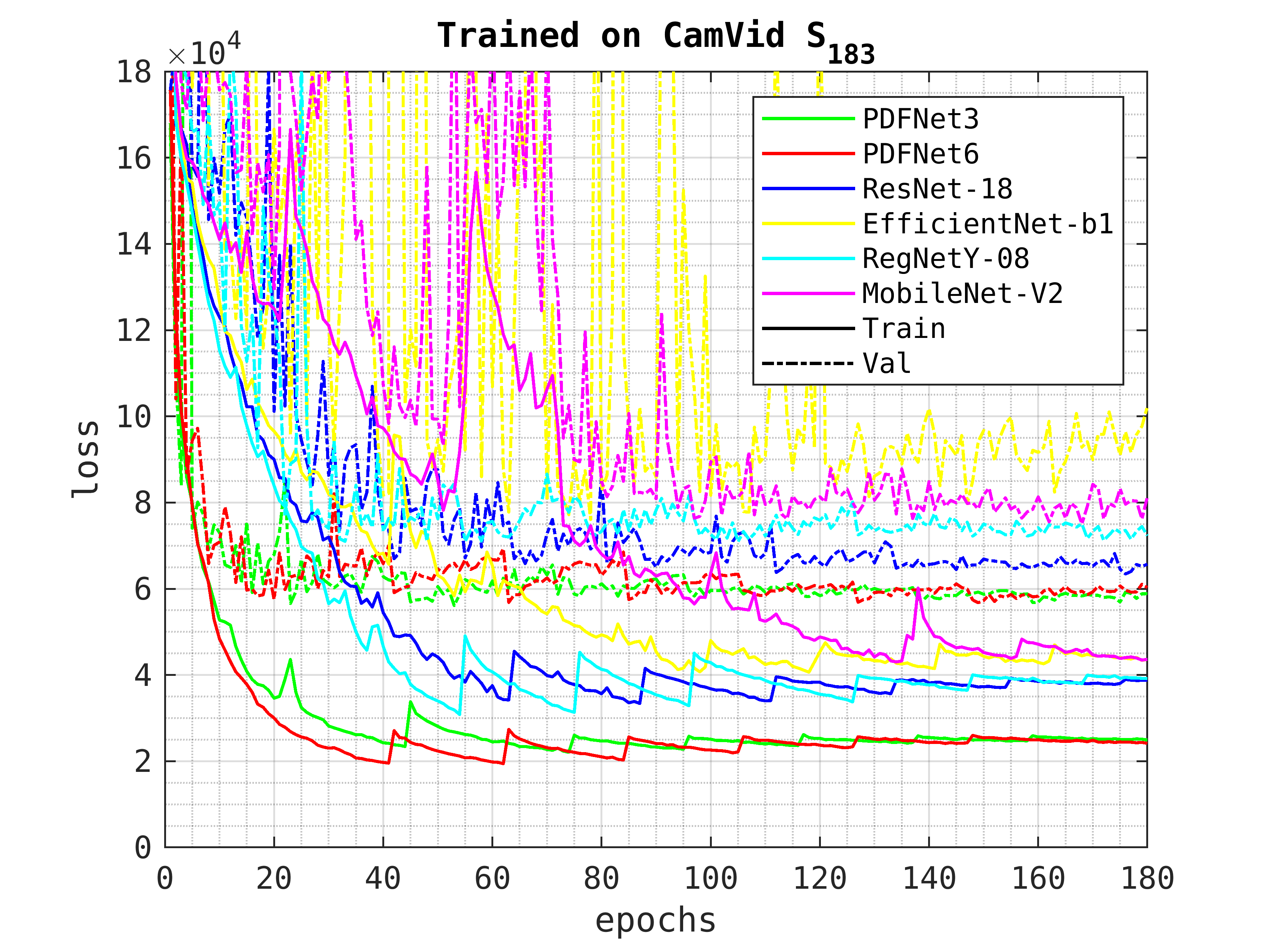}}
\subfigure{\includegraphics[width=65mm]{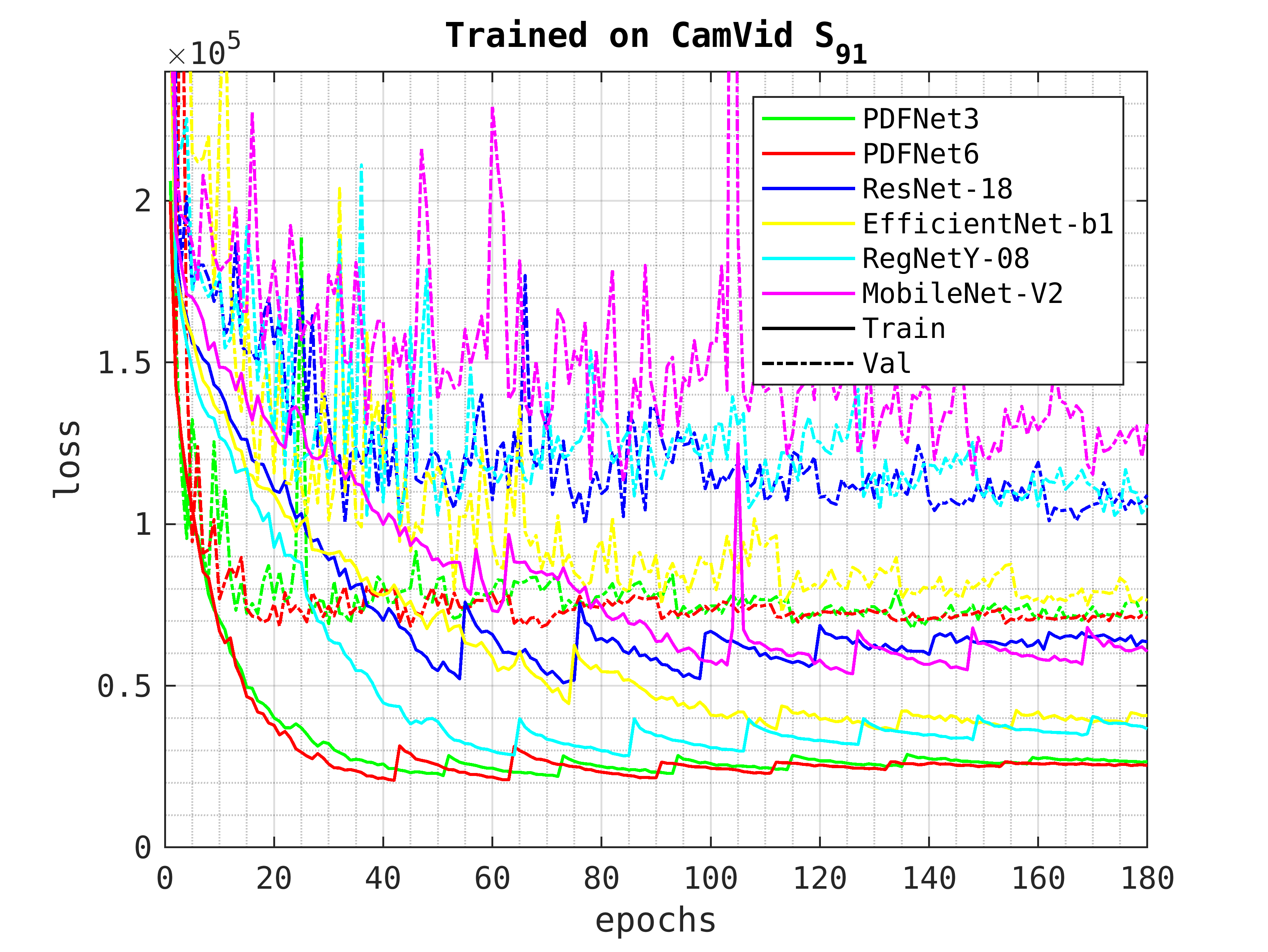}}
\caption{CamVid set-1 training plots}
\label{CV1}
\end{figure*}

\begin{figure*}
\centering     
\subfigure{\includegraphics[width=65mm]{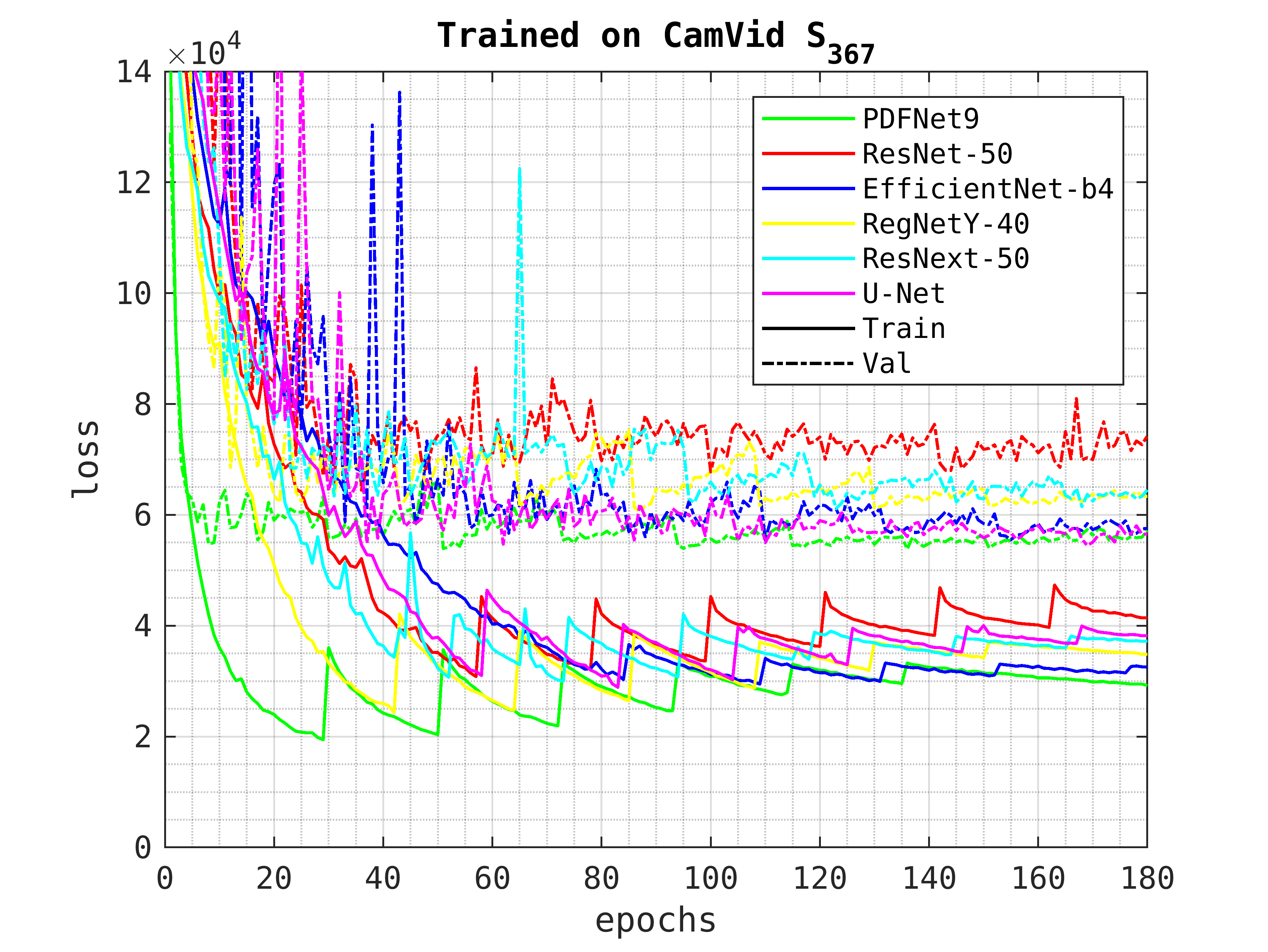}}
\subfigure{\includegraphics[width=65mm]{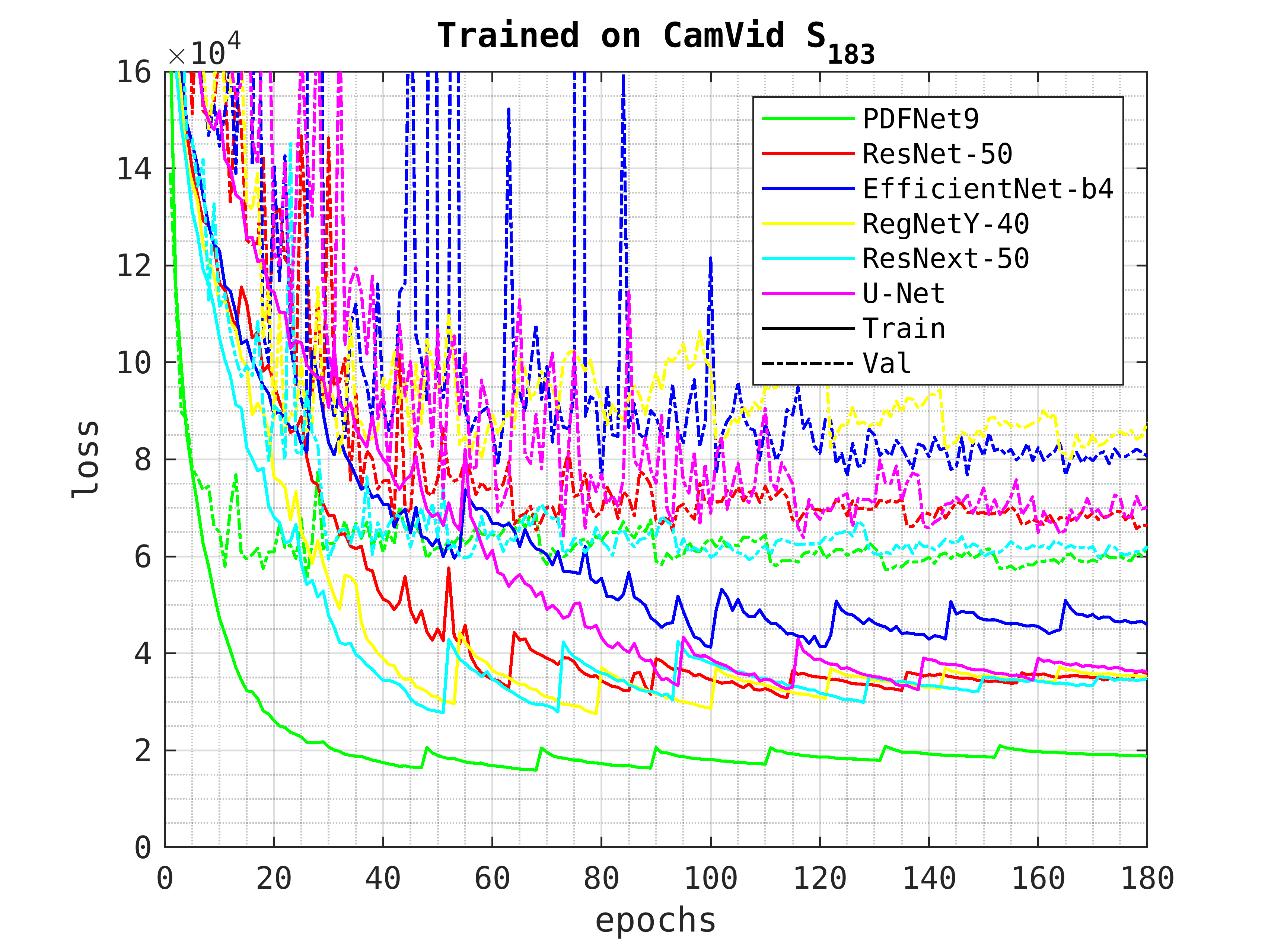}}
\subfigure{\includegraphics[width=65mm]{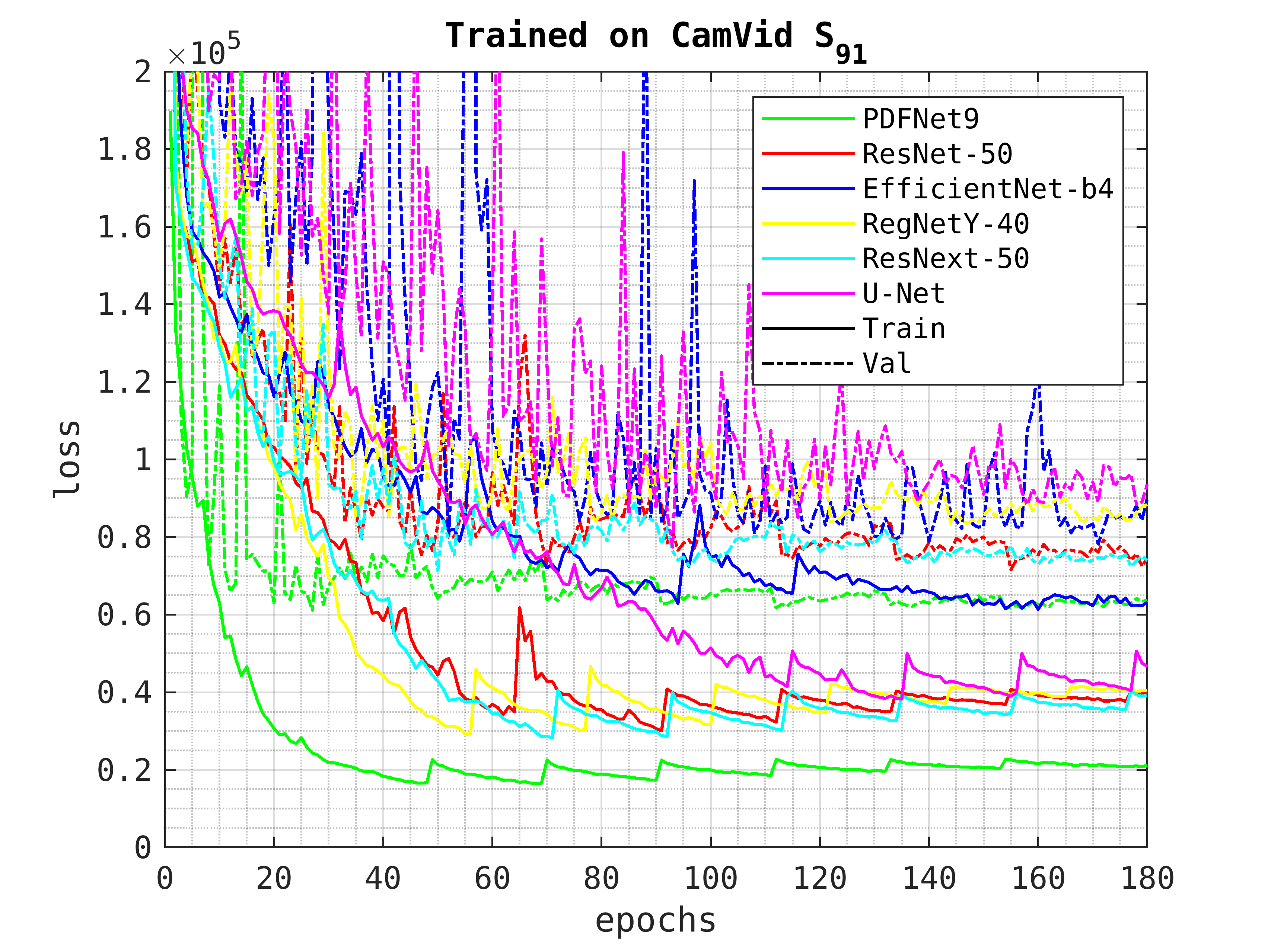}}
\caption{CamVid set-2 training plots}
\label{CV2}
\end{figure*}

\begin{figure*}
\centering     
\subfigure{\includegraphics[width=65mm]{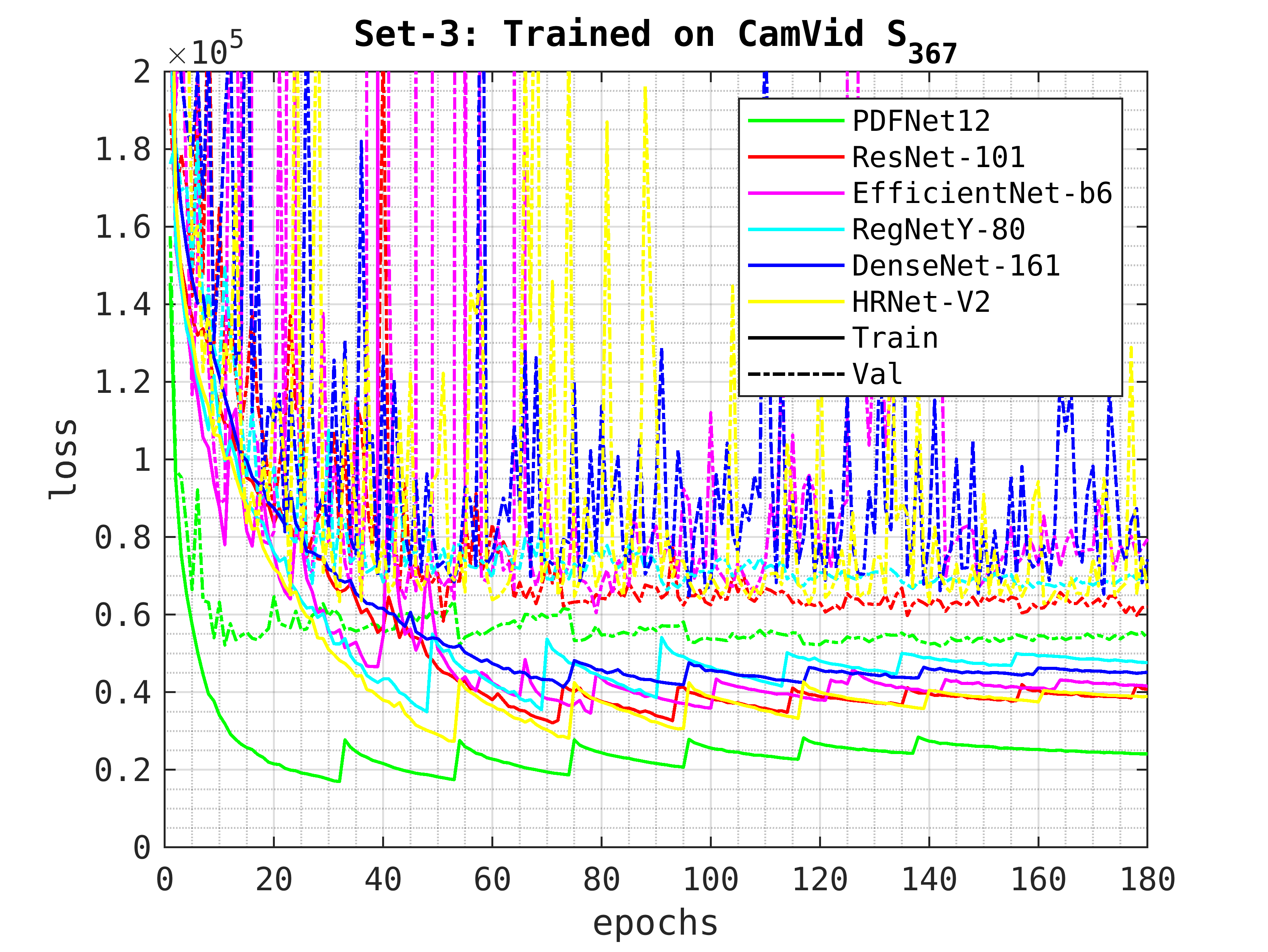}}
\subfigure{\includegraphics[width=65mm]{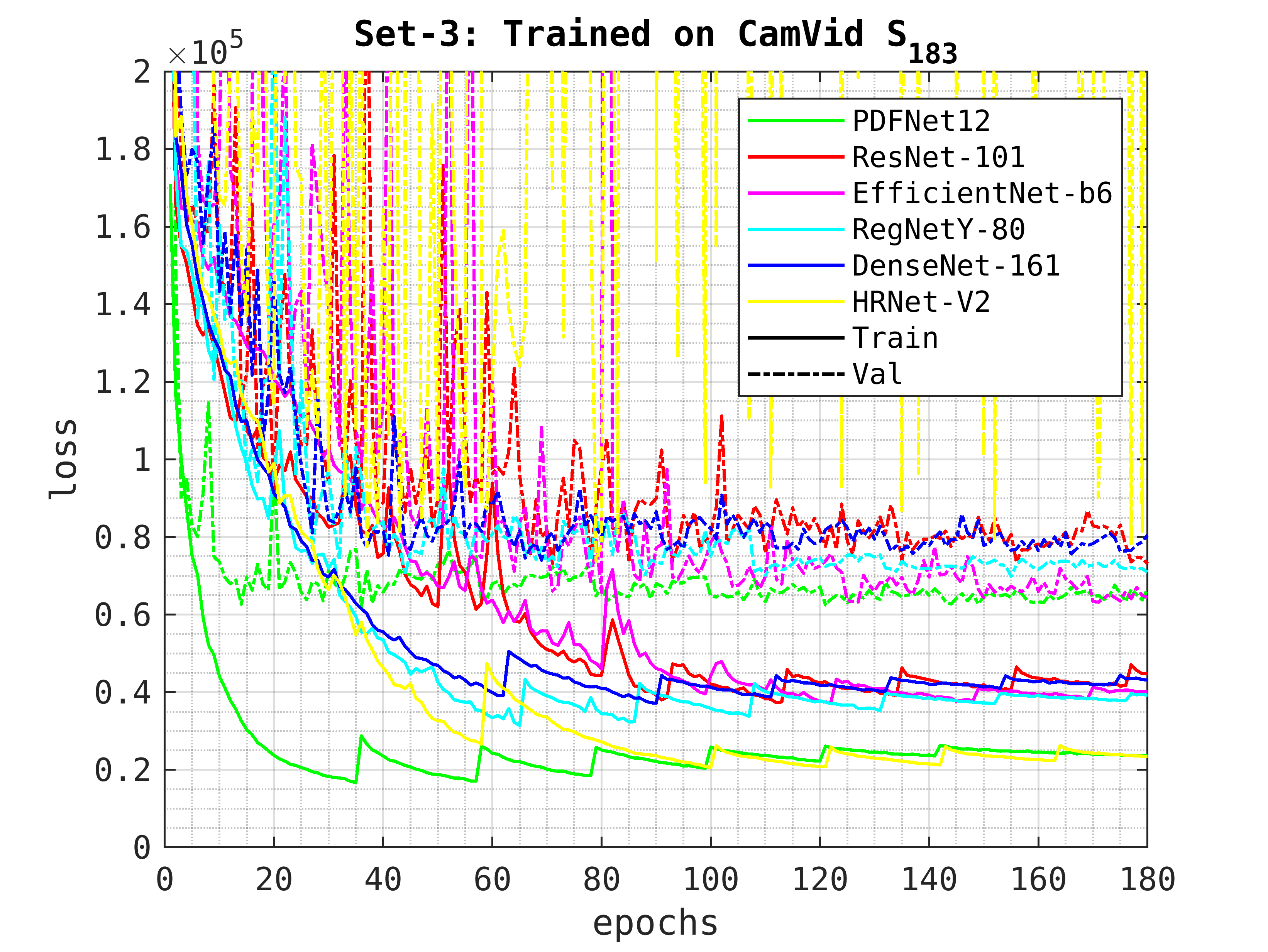}}
\subfigure{\includegraphics[width=65mm]{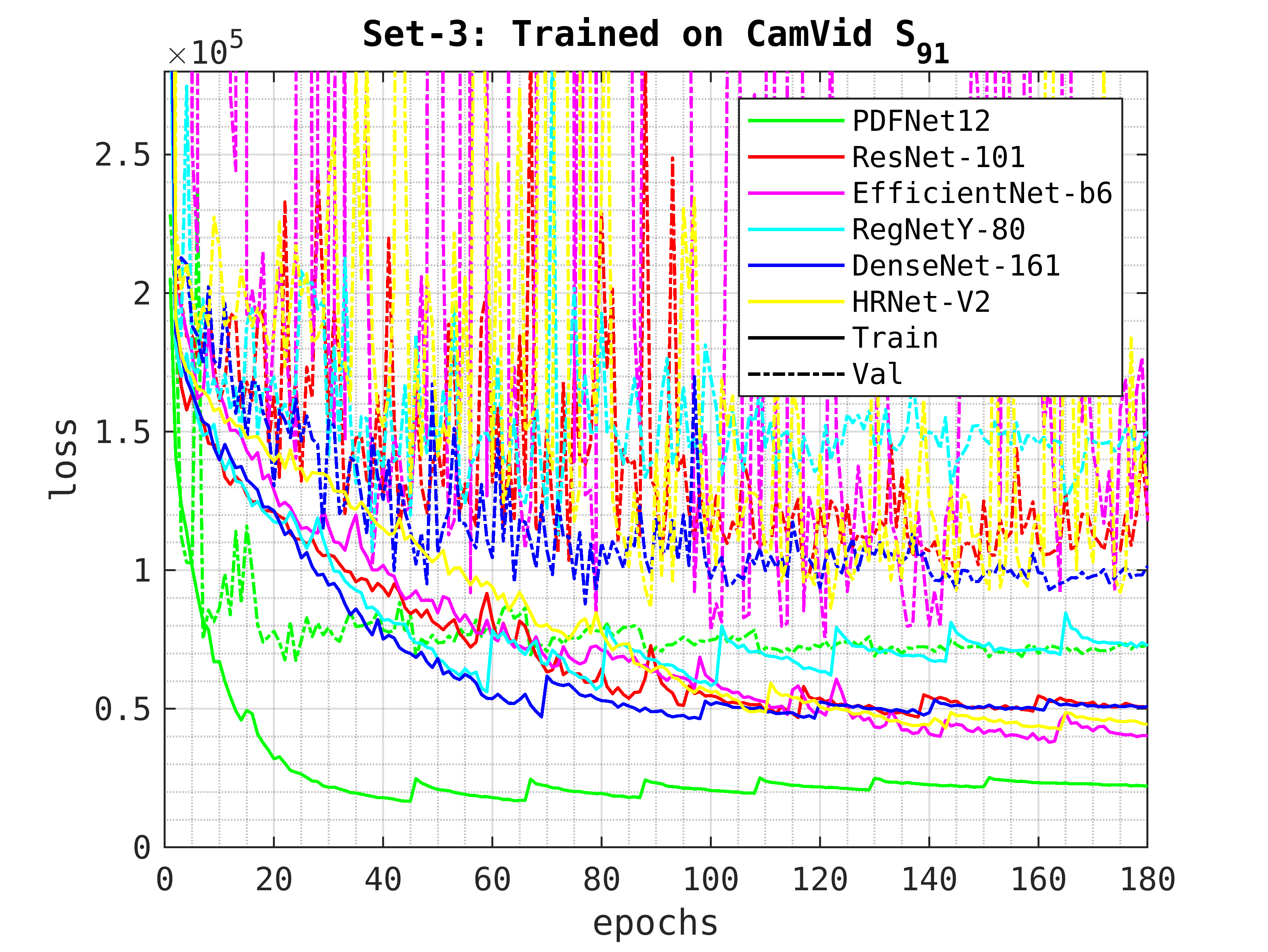}}
\caption{CamVid set-3 training plots}
\label{CV3}
\end{figure*}

\noindent In Figure \ref{CV1}, we present the training plots of the networks ResNet-18, EfficientNet-b1, RegNetY-08, MobileNet-V2, PDFNet3, and PDFNet6, trained on three subsets of training data ($T_{367}$, $T_{183}$, and $T_{91}$).\newline
In Figure \ref{CV2}, we present the training plots of the networks ResNet-50, EfficientNet-b4, RegNetY-40, ResNext-50, and PDFNet9, trained on three subsets of training data ($T_{367}$, $T_{183}$, and $T_{91}$).\newline
In Figure \ref{CV3}, we present the training plots of the networks ResNet-101, EfficientNet-b6, RegNetY-80, DenseNet-161, HRNet-V2, U-Net, and PDFNet12, trained on three subsets of training data ($T_{367}$, $T_{183}$, and $T_{91}$).


\section{KITTI}
In Table \ref{a-table-3}, we provide the class-wise results of the data ablation experiments evaluated on the KITTI training set.

\begin{table*}
\sisetup{detect-weight=true,detect-inline-weight=math}
\begin{center}

\begin{adjustbox}{width=1\textwidth}
\begin{tabular}{SlSSSSSSSSSSSSSSSSSSSS}
 \toprule
  {Subset}& {Method} & \rot{90}{road} & \rot{90}{sidewalk} &\rot{90}{building}  & \rot{90}{wall} &\rot{90}{fence}  & \rot{90}{pole} &\rot{90}{traffic light}  & \rot{90}{traffic sign} &\rot{90}{vegetation}  & \rot{90}{terrain} &\rot{90}{sky}  & \rot{90}{person} &\rot{90}{rider}  & \rot{90}{car} &\rot{90}{truck}  & \rot{90}{bus} &\rot{90}{train}  & \rot{90}{motorcycle} &\rot{90}{bicycle}  & \rot{90}{Average}      \\
  \midrule
   &ResNet18& 55.2 & 13.4 & 25.0 & 1.0 & 4.8 & 8.8 & 15.5 & 8.8 & 59.5 & 10.1 & 40.0 &  2.4 & \bfseries 0.6 &26.1 &1.5 & 4.9 &8.7 & 0.0 & 0.6 & 15.1\\
  &MobileNetV2& 61.0 & 11.2 & 44.4& 3.9 & 8.7 & 19.0 & 2.1 & \bfseries 10.3 & 73.2 & 17.2 & \bfseries 56.2 & 2.0 & 0.0 & \bfseries 52.5 & 3.2 & \bfseries 33.4 &  11.9 & 0.0 & 0.4 & \bfseries 21.6 \\
   &EfficientNetb1& 62.9 & \bfseries 18.5 & 40.3 & 0.0 & 9.8 & 16.3 & 13.2 & 7.9 & \bfseries 74.3 &  21.5 & 54.9 & 1.7 & 0.3 & 47.5 & 0.0 & 0.7 &  0.0 &0.0 &1.1 & 19.5\\
   &RegNetY08&  47.0 & 5.2 & 26.1& 0.0 & 0.2 & 0.0 & 0.0 & 0.0 & 52.9 & 6.1 & 45.0 & 0.5 & 0.0 & 22.8 &0.0 & 0.0 & 0.0 & 0.0 & 0.0 & 10.8 \\
   &ResNet101& 46.7 & 7.1 & 32.0 & 0.3 & 5.1 & 8.6 & 1.6 & 2.3 & 43.7 & 8.4 & 39.1 & 0.8 & 0.0 & 28.0 & 1.6 & 0.1 & 1.7 & 0.0 & 0.2 & 12.0\\
   &DenseNet161& 49.5 & 2.0 & 19.9& 0.5 & 1.4 & 3.1 & 0.2 & 5.2 & 47.1 & 9.7 & 26.5 & 0.9 & 0.0 & 26.8 & 0.6 & 0.0 & 5.4 & 0.0 & 0.3 & 10.5\\
   &HRNet-V2& 48.9 & 5.6 & 13.7 & 0.6 & 0.3 & 0.0 & 0.0 & 0.0 & 28.6 & 6.3 & 7.7 & 0.2 & 0.0 & 20.0 & 0.0 & 0.0 & 0.0 & 0.0 & 0.1 & 6.9 \\
 $T_{1487}$   &UNet& 50.6 & 2.5 & 33.4 & 4.5 & 5.6 & 7.9 & 11.0 & 4.2 & 57.3 & 4.9 & 43.5 & 1.3 & 0.0 & 32.7 & 1.6 & 1.7 & 3.9 & 0.0 & 0.4 & 14.1\\
   &PDFNet3& 61.5 & 3.9 & 35.0 & \bfseries 11.8 & 13.4 & 17.0 & 12.7 & 6.1 & 69.5 & 8.4 & 47.9 & 0.9 & 0.0 & 30.2 & 1.8 & 0.2 & 4.1 & 0.0 & 2.0 & 17.2 \\
   &PDFNet6& 54.5 & 5.4 & \bfseries 46.5 &  10.5 & 14.9 & 21.1 & \bfseries 18.3 & 5.1  &74.1  & 10.7 & 52.0 & 1.6 & 0.0 & 33.4 &  3.7 &7.6 & 5.2 & 0.0 & 0.8 & 19.2 \\
   &PDFNet9& 61.2 &4.5  &41.9  &3.9  & \bfseries 20.3  & \bfseries 24.4 &14.0   & 8.0  & 73.9 & 17.4 & 45.3 &1.9  & 0.2 & 37.0 & \bfseries 4.1& 16.7& \bfseries 20.2 & 0.0 & 1.4 &20.9  \\
   &PDFNet12& 60.6& 5.4 &33.2  &7.4  &12.2  &19.7  &17.3   &9.7   &69.8  & 23.3 & 47.3 & 1.2 &0.2  &29.9  &1.6  &2.4  &3.4  & 0.0 & \bfseries 3.0  &18.3  \\
   
   &PDFNet3-2S& 46.3 & 7.3 & 34.4 &  6.6 & 6.2 & 15.6 & 7.3 & 5.6 & 53.0 & 10.6 & 40.9 & 0.8 & 0.4 & 29.2 & 0.4 & 4.8 & 0.6 & 0.0 & 1.2 & 14.3\\
   &PDFNet6-2S& 64.3 & 13.0   & 35.2 & 11.3  & 13.5  & 20.7 & 11.0  & 6.3   & 58.5 & 10.6  & 44.4 & 1.1 &  0.2 & 46.1 & 1.5 & 10.7  & 16.7 & 0.0 & 2.6  & 19.4  \\
   &PDFNet9-2S& 53.4 & 15.0 & 38.5 &0.4  &2.7  &10.6  &0.1   &2.8   &66.9  &7.4  &44.1  &  1.1  &0.0& 27.9 & 0.1 &1.8  &3.0 & 0.0 &  0.9  &  14.6\\
   &PDFNet12-2s& \bfseries 66.1 & 6.2 & 40.5 & 11.5 & 12.2 & 17.0 & 4.5 & 5.3 & 69.0 & \bfseries 26.7 & 51.5 & \bfseries 2.7 & 0.0 & 24.6 & 1.3 & 3.2 & 9.4 &0.0 & 2.4 & 18.6 \\

    \midrule
   &ResNet18& 54.7 & 11.7 & 23.8 &  2.6 & 2.7 & 6.6 & 7.3 & 2.9 & 45.1 & 3.1 & 38.5 & 0.6 & 0.0 & 24.7 &  1.2 & 0.0 & 2.8 & 0.0 & 0.3 & 12.0\\
   &MobileNetV2& 61.1 & 13.1 & 39.8& 2.8 & 5.3 & 2.7 & 0.2 & 3.7 &  69.7 & 11.9 & \bfseries 62.3 &  1.3 & 0.0 & \bfseries 43.9 & 1.5 &  0.5 &  6.1 & 0.0 & 0.1 & 17.2\\
   &EfficientNetb1& 55.6 &  10.5 & 33.9 & 0.0  & 1.4 & 2.4 & 0.0 & 4.0 & 69.3 & 13.7 & 61.7 & 0.8 &  \bfseries 0.4 & 34.5 & 0.7 & 0.1 & 0.0 & 0.0 & 0.3 & 15.2\\
   &RegNetY08& 55.6 & 9.6 & 31.0 & 0.1 & 3.5 & 0.0 & 0.1 & 3.9 & 54.2 &  10.9 & 46.2 & 1.0 & 0.0 & 26.4 & 1.9 & 0.0 & 0.0 & 0.0 & 0.5 & 12.9\\
  &ResNet101& 45.8 & 0.3 & 29.5 & 0.0 & 1.2 & 4.8 & 0.0 & 1.6 & 59.1 & 14.4 & 38.3 & 0.5 & 0.0 & 24.8 & 0.0 & 0.0 & 0.0 & 0.0 & 0.5 & 11.6\\
   &DenseNet161& 48.6 & 5.1 & 17.3 & 2.1 & 0.2 & 1.7 & 0.1 & 3.0 & 22.1 & 2.3 & 28.5 & 0.5 & 0.0 & 20.0 & 0.0 & 0.4 & 0.0 & 0.0 & 0.3 &8.0\\
   &HRNet-V2& 41.1 & 6.0 & 16.8 & 0.0 & 0.0 & 0.0 & 0.0 & 0.0 & 36.6 & 9.8 & 23.6 & 0.0 & 0.0 & 19.2 & 0.0 & 0.0 & 0.0 & 0.0 & 0.0 &  8.1\\
   $T_{743}$  &UNet& 38.7 & 1.0 & 38.3 & 0.1 & 3.9 & 7.8 & 1.1 & 2.8 & 56.9 & 12.7 & 48.8 & 0.8 & 0.0 & 27.1 & 1.5 & 0.9 & 0.0 & 0.0 & 0.1 & 12.8 \\
   &PDFNet3& 47.7 & 5.8 & 33.8 &  7.2 & 6.8 & 9.6 & 3.4 & \bfseries 7.3 & 49.1 & 7.3 & 44.1 & 0.7 & 0.0 & 20.0 & 1.0 & 1.7 & 1.1 & 0.0 & 0.1 & 13.0\\
   &PDFNet6& 54.5 & 5.4 & \bfseries 46.5 &  \bfseries 10.5 & 14.9 &\bfseries 21.1 & \bfseries 18.3 & 5.1  & \bfseries 74.1  & 10.7 & 52.0 & \bfseries 1.6 & 0.0 & 33.4 & \bfseries 3.7 & \bfseries 7.6 & 5.2 & 0.0 & 0.8 & \bfseries 19.2 \\
   &PDFNet9& 63.1 &7.3  &35.5  & 3.7 & 14.0 & 20.9 &17.3   &5.3   & 65.6 & 10.5 & 42.5 & 0.9 &0.1  & 31.1 & 1.7 &3.3 & 4.6 &0.0  &0.7  &17.3  \\
   &PDFNet12& \bfseries 68.0& \bfseries 12.5 & 32.3 & 1.4 & 8.8 & 15.8 & 8.1  &7.2   & 70.3 & \bfseries 17.6 & 37.7 & 0.8 & 0.0 &33.1  &0.3  & 9.7& 0.0 & 0.0 & \bfseries2.1  &17.1  \\
   
   &PDFNet3-2S& 47.0 & 5.6 & 31.6 &  4.8 & 4.8 & 9.6 & 5.8 & 3.0 & 53.8 & 3.1 & 43.3 & 0.6 & 0.1 & 14.3 & 0.5 & 1.0 & 1.8 & 0.0 & 0.2 & 12.1\\
   &PDFNet6-2S& 61.1 & 5.5   & 34.6 & 5.0  & 10.9  & 12.5 & 3.6  & 4.6   & 59.8 & 12.6  & 44.8 & 0.9 &  0.1 & 22.3 & 1.2 & 0.6  &  1.5 & 0.0 &  0.0  & 14.8  \\
   &PDFNet9-2S& 62.0 & 4.4 & 38.2 & 9.7  & 9.2  & 14.2  & 12.3   & 5.8   & 67.1  & 4.8  & 33.8  &  1.8  & 0.0 & 35.0 & 0.8 & 3.3  & \bfseries 14.1 & \bfseries 0.1 &  0.3  &  16.7 \\   
   &PDFNet12-2s& 55.2& 8.8 & 37.2 & 5.8 & \bfseries 15.0  &11.8  & 12.4  &6.5   &64.2  &3.1  &45.7  &0.8  & 0.1 & 29.5 & 2.3 & 6.1& 3.8 &0.0  & \bfseries 2.1  & 16.3 \\
   \midrule
  &ResNet18& 48.8 & 1.4 & 22.0 & 0.0 & 0.3 & 2.4 & 0.0 & 1.9 & 43.5 & 16.7 & 33.4 & 0.7 & 0.0 & 27.4 & 0.0 & 0.0 & 0.0 &  0.0 & 0.1 & 10.5\\
   &MobileNetV2& 57.8 & 5.6 & 34.5 & 0.0 &  7.7 & 5.0 & 1.3 & 2.9 & 55.8 & 10.7 & \bfseries 76.8 & 0.6 & 0.0 & 38.4 &2.1 &  0.3 & 0.0 & 0.0 & 0.1 & 15.8 \\
   &EfficientNetb1& 50.5 & 8.2 & 31.5& 0.6 & 5.3 & 3.1 & 3.9 & 2.7 & 54.6 & 5.1 & 74.6 & 0.9 & 0.0 & 28.8 & 0.0 & 0.0 & 0.0 & 0.0 & 0.3 & 14.2\\
   &RegNetY08&  51.8 & 0.7 & 32.5 &  0.0 & 1.9 & 0.1 & 1.8 & 3.2 & 57.3 &   19.9 &  42.5 & 0.8 & 0.0 & 26.4 & 1.0 & 0.0 & 0.0 & 0.0 & 0.1& 12.6\\
  &ResNet101& 33.1 & 3.7 & 21.9 & 0.0 & 0.5 & 1.6 & 0.0 & 3.8 & 37.8 & 3.8 & 34.1 & 0.5 & 0.0 & 26.0 & 0.7 & 0.0 & 0.0 & 0.0 & 0.0 & 8.8\\
   &DenseNet161& 52.1 & 6.5 & 22.1 & 0.3 & 1.0 & 0.0 & 0.0 & 0.0 & 49.5 & 9.6 & 19.9 & 0.3 & 0.0 & 23.7 & 0.0& 0.0 & 0.0 & 0.0 & 0.1 & 9.7\\
   &HRNet-V2& 47.4 & 1.1 & 18.0 & 0.0 & 0.0 & 0.0 & 0.0 & 0.0 & 29.8 & 5.7 & 34.3 & 0.6 & 0.0 & 13.7 & 0.0 & 0.0 & 0.0 & 0.0 & 0.0 & 7.9 \\
  $T_{371}$   &UNet& 36.6 & 0.1 & 20.2 & 0.0 & 0.6 & 2.2 & 0.3 & 2.4 & 51.3 & 1.9 & 39.8 & 0.4 & 0.0 & 19.2 & 0.8 & 0.0 & 0.3 & 0.0 & 0.0 & 9.3 \\
   &PDFNet3& 55.5 & 6.7 & 33.7 &  2.2 & \bfseries 14.5 &  14.0 & 7.9 &\bfseries 8.6 & 64.1 & 11.0 & 44.2 & 1.3 &  0.1 & 27.7 & 1.9 & 5.8 & \bfseries 3.2 &  0.0 & 0.8 & 16.0\\
   &PDFNet6& 50.8 & 8.4 & \bfseries 40.8 &  0.4 & 8.3 & 11.4 & 4.5 & 5.6  & 63.2  & 7.0 & 47.5 & 0.7 & 0.0 & 27.6 & 1.3 & 1.9 & 2.8 & 0.0 & 0.3 & 15.2 \\
   &PDFNet9& 59.1 &  9.8 &  34.2 &  7.3 & 11.5 & 10.8  & 3.8  & 6.4   & 64.2  & 6.0 & 39.3 & 0.8 &  0.3 & 32.3 & 0.2 & 3.0 & 1.9 & 0.3 & \bfseries 2.2 & 15.4 \\
   &PDFNet12&\bfseries 65.9&9.2  &37.2  & \bfseries 6.6  & 9.0 & 10.8 & \bfseries 13.5  &5.7   & \bfseries 65.2  & \bfseries 20.4  &46.0  & 0.7 & \bfseries 0.3  & \bfseries 40.1 & \bfseries 2.8 &3.5  & 0.0 &  0.2  & 0.5 & \bfseries 17.8 \\
   
   &PDFNet3-2S& 59.8 & 2.5 & 37.7 &  1.6 & 8.3 & 7.0 & 0.0 & 2.7 & 55.2 & 6.8 & 44.9 & 0.7 & 0.1 & 19.0 &  0.1 & 3.5 & 2.7 & 0.0 & 0.1 & 13.3\\
   &PDFNet6-2S& 48.6 & \bfseries 12.1   & 36.7 & 0.8  & 6.4  & 8.2 & 3.6  &4.1   & 64.7 & 3.4  & 44.6 & \bfseries 1.6 & 0.1 & 28.3 & 0.8 & \bfseries 10.3  & 1.4 & 0.0 & 0.6  & 14.5  \\
   &PDFNet9-2S& 59.1 & 9.8 & 34.2 & 7.3  & 11.5  & 10.8  & 3.8   & 6.4   & 64.2  & 6.0  & 39.3  & 0.8  & \bfseries 0.3 & 32.3 & 0.2 & 3.0 & 1.9 & \bfseries 0.3 & \bfseries 2.2  &  15.4 \\   
   &PDFNet12-2s&63.1 &7.8  &41.6  &4.8  & 8.0 & \bfseries 16.2  &9.4   &5.2   &63.9  &8.9  &55.1  &\bfseries 1.6  & 0.0 & 25.8 & 0.5 &  7.0 &1.4  & 0.1 & 0.6 & 16.9 \\
   \midrule
   &ResNet18& 44.5 & 1.1 & 19.8 & 0.0 & 0.0 & 0.0 & 0.0 & 0.0 & 47.1 & 3.3 & 17.0 & 0.0 & 0.0 & 17.2 & 0.0 & 0.0 & 0.0 & 0.0 & 0.0 & 7.9\\
  &MobileNetV2& 57.1 & 12.3 & 28.6 & 0.0 & 0.0 & 0.0 & 0.0 & 0.0 & 55.3 & \bfseries 19.0 & 44.0 & 0.0 & 0.0 & \bfseries 33.7 & 0.0 & 0.0 & 0.0 & 0.0 & 0.0 & 13.2\\
   &EfficientNetb1& 38.9 & 3.5 & 20.3 & 0.0 & 0.2 & 0.0 & 0.0 & 0.1 & 28.8 & 0.9 &39.2 & 0.4 & 0.0 & 28.1 &0.0 & 0.0 & 0.0 & 0.0 & 0.0 & 8.1\\
   &RegNetY08& 54.9 & 5.5 & 30.8 &  0.0 & 1.3 & 0.0 & 0.8 & 1.4 & 40.6 &  12.0 & 43.0 & 0.5 & 0.0 &  27.1 & 0.2 & 0.0 & 0.0 & 0.0 &0.0& 11.5 \\
   &ResNet101& 51.2 & 0.8 & 18.6 & 0.0 & 0.0 & 0.0 & 0.0 & 0.0 & 40.0 & 14.0 & 22.9 & 0.0 & 0.0 & 20.4 & 0.0 & 0.0 & 0.0 & 0.0 & 0.0 & 8.8\\
   &DenseNet161& 53.6 & 5.5 & 24.3 & \bfseries 2.8 & 1.9 & 0.0 & 0.0 & 0.0 & 46.7 & 12.9 & 35.8 & 0.2 & 0.0 & 27.8 & 0.3 & 0.0 & 0.0 & 0.0 & 0.1 & 11.2\\
   &HRNet-V2& 54.8 & 5.2 & 25.3 & 0.0 & 0.0 & 0.0 & 0.0 & 0.0 & 42.9 & 0.0 & 32.3 & 0.0 & 0.0 & 17.2 & 0.0 & 0.0 & 0.0 & 0.0 & 0.0 & 9.3 \\
  $T_{185}$   &UNet& 22.3 & 1.6 & 29.8 & 0.2 & 0.2 & 0.0 & 0.0 & 3.2 & 49.9 & 4.0 & 42.9 & 0.5 & 0.0 & 15.4 & 0.0 & 0.0 & 0.6 & 0.0 & 0.1 & 9.0 \\
   &PDFNet3& 53.1 & 4.6 & 24.9 &  0.8 & 2.9 & 7.0 & 6.3 & 4.1 & 58.5 & 3.3 & 43.3 & 0.5 &\bfseries  0.3 & 18.7 & 0.3 & 4.0 & 0.0 & 0.0 & 0.0 & 12.3\\
   &PDFNet6& 49.6 & \bfseries 19.4 & 29.1 &  0.0 & 3.4 & 12.2 & 5.8 & 3.7  & 59.8  & 4.4 & 46.7 & 0.8 & 0.1 & 29.9 & 1.2 & \bfseries 6.8 & 0.1 & 0.0 & 0.3 & 14.4 \\
   &PDFNet9& 58.3 & 15.6 & 31.9 & 0.0 &2.3  &12.1  &2.1   &  5.3 & 58.0 &7.1  &46.6  &0.8  & 0.1 & 27.5 & 1.4 &2.5 & 1.7 & 0.0 & 0.2 & 14.4 \\
   &PDFNet12& 59.6& 8.1 & \bfseries 40.3 & 0.6 &\bfseries 5.6 & \bfseries 15.3  & \bfseries 13.5   & \bfseries 6.0   & \bfseries 67.0 & 3.8 &53.4  &  1.0  &0.0  &27.3  &0.8  &0.4  &0.0  &0.0  & \bfseries 0.6   & \bfseries 16.0 \\
 
   &PDFNet3-2S& 54.9 & 5.2 & 27.5 &  0.3 & 3.4 & 9.8 & 2.8 & 3.8 & 57.1 & 6.3 & 44.6 & 0.6 &  0.0 & 18.3 & 1.6 & 4.5 & 1.4 & 0.0 & 0.1 & 12.7\\
   &PDFNet6-2S& 58.8 & 14.4   & 25.4 & 0.1 & 4.2  & 10.9 & 0.3  & 3.2   & 53.4  & 3.3  &\bfseries 46.9 & 0.8 & 0.0 & 29.2 & \bfseries 2.2 & 6.4  & \bfseries 2.8 & 0.0 & 0.2  & 13.9  \\
   &PDFNet9-2S& 59.4 & 6.0 & 30.9 & 0.6  & 3.1  & 12.3  & 6.1   & 3.2  & 60.6  & 2.5  & 46.1  & \bfseries 1.2  & 0.1 & 32.7 & 1.1 & 8.8  & 0.0 & 0.0 & \bfseries 0.6  &  14.5\\
   &PDFNet12-2s& \bfseries 63.6 &  9.0& 30.7 &1.9  &1.4  &11.9  & 0.1  &4.4   &66.4  & 6.8 & 44.7 & 1.0  & 0.0 & 30.0 & 1.7 & 0.3 &  2.2 & \bfseries 0.1  &0.4  &14.6  \\
   \midrule
   
   &ResNet18& 52.8 & 4.4 & 22.4& 0.0 & 0.0 & 0.0 & 0.0 & 0.0 & 45.3 & 7.8 & 35.5 & 0.0 & 0.0 & 17.8 & 0.0 & 0.0 & 0.0 & 0.0 & 0.0 & 9.8\\
   &MobileNetV2& 54.8 &\bfseries 7.8 & 32.5 & 0.0 & 0.0 & 0.0 & 0.0 & 0.0 & 56.3 & \bfseries 14.4 & \bfseries 74.7 & 0.0 & 0.0 & 17.5 & 0.0 & 0.0 & 0.0 & 0.0 & 0.0 & 13.6 \\
   &EfficientNetb1& 31.9 & 2.8 & 13.0 & 0.1 & 0.0 & 0.0 & 0.0 & 0.1 & 4.2 & 8.9 & 38.4 &0.2 & 0.0 & 16.0 & 0.0 & 0.0 & 0.0 & 0.0 & 0.0 & 6.1\\
   &RegNetY08& 54.4 & 9.5 & 20.0 & 0.0 & 0.0 & 0.0 & 0.0 & 0.0 & 30.2 & 12.0 & 45.5 & 0.0 & 0.0 & 20.4 & 0.0 & 0.0 & 0.0 & 0.0 & 0.0 & 10.1\\
  &ResNet101& 38.2 & 0.2 & 17.6 & 0.0 & 0.0 & 0.0 & 0.0 & 0.0 & 29.8 & 9.1 & 30.1 & 0.0 & 0.0 & 17.1 & 0.0 & 0.0 & 0.0 & 0.0 & 0.0 & 7.5\\
  &DenseNet161& 50.9 & 2.1 & 17.7 & 0.0 & 0.0 & 0.0 & 0.0 & 0.0 & 38.9 & 3.7 & 24.7 & 0.1 & 0.0 & 21.3 & 0.0 & 0.0 & 0.0 & 0.0 & 0.0 & 8.4\\
   &HRNet-V2& 52.2 & 2.2 & 19.7 & 0.0 & 0.0 & 0.0 & 0.0 & 0.0 & 47.0 & 0.0 & 69.3 & 0.0 & 0.0 & 22.3 & 0.0 & 0.0 & 0.0 & 0.0 & 0.0 & 11.2\\
  $T_{92}$  &UNet& 12.2 & 0.5 & 20.0 & 0.1 & 0.0 & 0.0 & 0.0 & 2.4 & 54.9 & 3.7 & 39.8 & 0.3 & 0.0 & 15.4 & 0.0 & 0.0 & 0.8 & 0.0 & 0.0 & 7.9 \\
   &PDFNet3& 50.1 & \bfseries 16.9 & 39.3 &  0.3 & 7.0 & 14.5 & 0.3 & 2.3 & 58.6 & 7.6 & 49.0 & 0.9 & 0.0 & 30.1 & \bfseries 1.0 & 3.2 & 3.5 & 0.0 & 0.3 & 15.0\\
   &PDFNet6& 58.9 &13.4   & \bfseries 40.0 &1.0  &3.2  & 14.3 &\bfseries 6.3  &4.1   & 64.6 &5.8  & 54.2 & 0.8 &\bfseries  0.4 & 31.5 & 0.0 & \bfseries 7.4  & \bfseries 5.7 & 0.0 & \bfseries 0.9  &16.4  \\
   &PDFNet9& 53.4 & 15.0 & 38.5 &0.4  &2.7  &10.6  &0.1   &2.8   &66.9  &7.4  &44.1  & \bfseries 1.1  &0.0& 27.9 & 0.1 &1.8  &3.0 & 0.0 & \bfseries 0.9  &  14.6\\
   &PDFNet12& \bfseries 66.2& 9.7 & 38.0 & 0.4 & \bfseries 8.4 & \bfseries 16.5 & 6.2  &\bfseries 5.7   & \bfseries 69.3  &6.6  &64.3  & \bfseries 1.1  & 0.0 &\bfseries 38.0  & 0.9 &0.0 & 0.0 & 0.0 & 0.0 &\bfseries 17.4 \\
   &PDFNet3-2S& 58.5 & 9.3 & 32.2 &  0.8 & 2.9 & 7.8 & 0.9 & 2.8 & 61.2 & 11.0 & 49.5 & 0.8 & 0.1 & 24.6 &  0.0 & 6.7 & 1.2 &\bfseries 0.4 & 0.4 & 14.3\\
   &PDFNet6-2S& 51.4 & 16.5   & 39.5 & 0.0  & 0.1  & 0.0 & 0.0  & 0.0   & 65.5 & 4.2  & 49.6 & 1.0 & 0.0 & 29.1 & 0.0 & 0.0  & 4.8 & 0.0 &  0.1  & 13.8  \\
   &PDFNet9-2S& 44.0 & 8.6 & 35.4  & 0.2  & 6.0  & 7.5  & 0.9   & 1.9   & 64.4  & 5.9  & 56.1  & \bfseries 1.1  & 0.0 & 35.0 & \bfseries 1.0 & 0.0  & 0.0 & 0.1 & 0.1  &  14.1 \\
   &PDFNet12-2s&59.9 & 3.6 &35.1  & \bfseries 1.5  & 0.1 & 9.6 & 0.0  &0.6   &57.6  &4.8  &53.2  & 0.8 &0.0  &24.1  &0.3  &1.8 & 1.6 &0.0  &0.0  &13.4  \\
   \bottomrule
\end{tabular}
\end{adjustbox}
\end{center}
 \caption{Class-wise results of the data ablation experiments evaluated on the KITTI training set}
 \label{a-table-3}
\end{table*}

\end{document}